\documentclass[10pt,journal,compsoc]{IEEEtran}

\usepackage[T1]{fontenc}

\ifCLASSOPTIONcompsoc
    \usepackage[nocompress]{cite}
\else
    \usepackage{cite}
\fi
\usepackage{amsmath,amsfonts}
\usepackage{array}
\usepackage{booktabs}
\usepackage{graphicx}
\usepackage{tabularx}
\usepackage[table]{xcolor}
\usepackage[hidelinks]{hyperref}
\usepackage{tikz}
\usepackage[skins,breakable]{tcolorbox}
\usepackage{cleveref}

\DeclareRobustCommand{\Dref}[1]{\hyperref[sec:d#1]{D#1}}
\definecolor{protocolbriefback}{HTML}{DAE8FC}
\definecolor{protocolbriefframe}{HTML}{6C8EBF}
\definecolor{systembriefback}{HTML}{FFF2CC}
\definecolor{systembriefframe}{HTML}{D6B656}
\definecolor{behaviorbriefback}{HTML}{D5E8D4}
\definecolor{behaviorbriefframe}{HTML}{82B366}
\definecolor{readerpromptback}{HTML}{F8CECC}
\definecolor{readerpromptframe}{HTML}{B85450}

\tcbset{
    protocolbrief/.style={
        colback=protocolbriefback,
        colframe=protocolbriefframe,
        boxed title style={colback=protocolbriefframe, colframe=protocolbriefframe},
    },
    systembrief/.style={
        colback=systembriefback,
        colframe=systembriefframe,
        boxed title style={colback=systembriefframe, colframe=systembriefframe},
    },
    boundarybrief/.style={
        systembrief,
        frame hidden,
        borderline={0.8pt}{0pt}{behaviorbriefframe, dash pattern=on 3pt off 2pt},
    },
    behaviorbrief/.style={
        colback=behaviorbriefback,
        colframe=behaviorbriefframe,
        boxed title style={colback=behaviorbriefframe, colframe=behaviorbriefframe},
    },
}

\newtcolorbox{briefbox}[2][]{
    enhanced jigsaw, breakable, sharp corners,
    colback=black!10, colframe=black!65, boxrule=0.6pt,
    left=10pt, right=10pt, bottom=8pt, top=10pt,
    fonttitle=\footnotesize\bfseries\sffamily,
    coltitle=black,
    title={#2 \MakeUppercase{In brief}},
    attach boxed title to top left={xshift=12pt, yshift=-\tcboxedtitleheight/2},
    boxed title style={
        colframe=black!65, boxrule=0pt, sharp corners,
        colback=black!30,
        left=3pt, right=3pt, top=1pt, bottom=1pt,
    },
    #1
}

\newtcolorbox{readerpromptbox}{
    enhanced jigsaw, breakable, sharp corners,
    colback=readerpromptback, colframe=readerpromptframe, boxrule=0.5pt,
    left=8pt, right=8pt, bottom=7pt, top=9pt,
    fonttitle=\footnotesize\bfseries\sffamily,
    coltitle=black,
    title={\MakeUppercase{Try before reading}},
    attach boxed title to top left={xshift=10pt, yshift=-\tcboxedtitleheight/2},
    boxed title style={
        colframe=readerpromptframe, boxrule=0pt, sharp corners,
        colback=readerpromptframe,
        left=3pt, right=3pt, top=1pt, bottom=1pt,
    }
}

\begin{document}
\bstctlcite{bstctl:etal}

\title{Continual Reinforcement Learning Is an Integration Problem}

\author{%
    Nis~Meinert,
    Jakob~Gawlikowski,
    Edward~Grant,
    Max~Cyprys,
    Nathan~Schurr,
    Neil~Boston,
    Alfredo~Ibias,
    Milind~Tambe,
    and Janusz~Marecki
    \IEEEcompsocitemizethanks{
        \IEEEcompsocthanksitem N.\ Meinert, J.\ Gawlikowski, M.\ Cyprys, N.\ Schurr, and J.\ Marecki are with FractalBrain, London, United Kingdom (e-mail: \{nis,jakob,max,nathan,janusz\}@fractalbrain.ai).
        \IEEEcompsocthanksitem E.\ Grant is with University College London (UCL), London, United Kingdom (e-mail: edward339@gmail.com).
        \IEEEcompsocthanksitem N.\ Boston is an independent researcher, London, United Kingdom (e-mail: neil1.boston@gmail.com).
        \IEEEcompsocthanksitem A.\ Ibias is an independent researcher, Barcelona, Spain (e-mail: \mbox{alfibi@proton.me}).
        \IEEEcompsocthanksitem M.\ Tambe is with Harvard University, Cambridge, MA, USA (e-mail: milind\_tambe@harvard.edu).
    }
}

\hypersetup{
    pdftitle={Continual Reinforcement Learning Is an Integration Problem},
    pdfauthor={Nis Meinert, Jakob Gawlikowski, Edward Grant, Max Cyprys, Nathan Schurr, Neil Boston, Alfredo Ibias, Milind Tambe, Janusz Marecki},
    pdfsubject={Design and evaluation desiderata for continual reinforcement learning},
    pdfkeywords={Continual reinforcement learning, reinforcement learning, continual learning, lifelong learning, online learning, catastrophic forgetting, stability-plasticity dilemma, benchmark design}
}

\markboth{%
%IEEE Transactions on Pattern Analysis and Machine Intelligence%
}{Meinert \MakeLowercase{\textit{et al.}}: Continual Reinforcement Learning Is an Integration Problem}

\IEEEtitleabstractindextext{%
\begin{abstract}
    This paper presents an opinionated set of ten desiderata intended to inform the design and evaluation of reinforcement-learning agents that learn continually throughout a continuing lifetime.
    We use these desiderata to examine a representative selection of existing approaches.
    Progress on individual desiderata is valuable, but the longer-term aim should be to integrate all ten within a single agent.
    To our knowledge, no published evaluation has yet provided evidence for such an integration.
    Likewise, the representative benchmarks we review probe different parts of this target, but separate evaluations do not establish that the corresponding properties coexist within one agent throughout a continuing lifetime.
    As a secondary contribution, we therefore present a compact persistent-world scenario as a playable companion to the desiderata and outline a proposed benchmark that examines a broad range of associated behavioral capabilities together within one such lifetime.
\end{abstract}

}

\maketitle
\IEEEdisplaynontitleabstractindextext
\IEEEpeerreviewmaketitle

\IEEEraisesectionheading{\section{Introduction}}

\IEEEPARstart{C}{ontinual} reinforcement learning (CRL) seeks long-lived autonomous agents that keep learning from experience, extending and revising their knowledge and skills as conditions and objectives change.
An early account of this form of learning comes from Ring, who described it as open-ended development in which increasingly complex behavior is built incrementally and hierarchically from previously acquired skills \cite{ring1997}.
CRL also inherits the lifelong-learning emphasis on accumulating and transferring knowledge across problems \cite{thrun1995}, together with a system-level perspective that focuses on complete learning systems instead of isolated algorithms \cite{silver2013}.
These early ideas remain highly relevant today to multi-agent systems in which cooperating or competing agents continue to learn and adapt \cite{nekoei2021,kim2021} and, more broadly, to the development of AGI \cite{morris2024,hendrycks2025,burnell2026}.
Despite sustained progress, however, no general solution to this learning problem has yet emerged \cite{khetarpal2022,wolczyk2021,abel2023}.

\subsection{Motivation and Research Gap}

Deployment environments may have relevant properties that were not fully known during training, while operating conditions, objectives, and sensing capabilities may also change over time \cite{khetarpal2022}.
These demands motivate learning from ongoing interaction, especially when repeated retraining from scratch is impractical \cite{verwimp2024}.
Current practice falls short of the continual-learning ambition in several distinct ways.
One limitation is that standard deep-learning methods are commonly trained in a separate weight-updating phase and then deployed with fixed parameters, while sustained updating across changing data can progressively impair further learning \cite{dohare2024}.
More broadly, continual learning is not merely a chosen training protocol but is inherent to model editing, personalization, on-device learning, efficient retraining, and reinforcement learning, where new data and changing conditions arise after deployment \cite{verwimp2024}.
Continual learning across these applications, including CRL, entails the classical stability--plasticity dilemma of incorporating new experience without losing knowledge that may remain useful \cite{carpenter1987,anand2023}.
This tension appears in two distinct but opposing failure modes: catastrophic forgetting occurs when learning from new experience damages useful abilities acquired earlier \cite{mccloskey1989,french1999,kirkpatrick2017}, whereas loss of plasticity occurs when changes accumulated through earlier learning progressively reduce the agent's ability to keep learning \cite{dohare2024}.
A continual learner can also exhibit a transient failure mode known as the stability gap, in which performance on previously learned tasks drops sharply during a transition but later recovers \cite{delange2023}.
An agent can appear stable by stopping further updates, replaying all past experience, or allocating a separate model to every task.
The first strategy sacrifices plasticity, while the others shift the difficulty into ever-growing storage, model capacity, or computation \cite{wolczyk2021,harun2023}.
Accordingly, Wo{\l}czyk et al.\ argue that CRL should evaluate not only whether an agent retains old skills, but also whether those skills support forward transfer by helping it learn new ones under bounded storage, model capacity, and computation \cite{wolczyk2021}.
A credible lifelong agent must therefore retain useful knowledge, remain able to learn, transfer and revise what it knows, and do so within declared resource limits.

How strongly an evaluation supports claims of stability, plasticity, transfer, and bounded resource use depends on the conditions under which the agent learns and is tested.
In our target setting, the agent remains deployed throughout one continuing lifetime, receives no predefined segmentation of experience or repeated returns to controlled initial conditions, and cannot be configured using information from its complete lifetime.
To make controlled study tractable, common experimental formulations set aside parts of this challenge by representing non-stationarity as finite, piecewise-stationary sequences of externally delimited tasks, often keeping state and action spaces fixed and supplying task identities or boundaries \cite{khetarpal2022,wolczyk2021,powers2022}.
Further conveniences include episodic resets \cite{sharma2022} and hyperparameter tuning over complete lifetimes, which can leak information unavailable to an agent that truly has only one life \cite{mesbahi2025,chen2022,bible1611}.
Without these conveniences, actions can have persistent and irreversible consequences, observations are only a partial record of a vast world, rewards may be sparse or delayed, and neither relevant knowledge nor future objectives can be enumerated in advance \cite{schaul2018,khetarpal2022}.
This difficulty is compounded by the partly endogenous data stream: the current policy determines which states, outcomes, and learning signals are encountered, so failures of exploration or representation can restrict the evidence available for later adaptation \cite{khetarpal2022,abel2023}.
This broader view of CRL is especially relevant to adaptive multi-agent systems.
From the perspective of one agent, learning partners create endogenous non-stationarity because they change in response to experience and potentially to the agent itself, while newly encountered partners require further adaptation without erasing earlier coordination strategies \cite{nekoei2021,kim2021}.
To organize this broader view, we formulate ten desirable properties for CRL, which we call desiderata.
Several of them also describe enabling capabilities for continual multi-agent learning, including task-free online adaptation, stability and transfer under bounded resources, persistent models of changing behavior, and long-horizon decisions under concurrent objectives.
The limitations discussed above show that the ability to avoid catastrophic forgetting and preserve plasticity is necessary but not sufficient for lifelong learning.
An agent may still lack bounded mechanisms for allocating capacity and computation, memory and credit assignment over long delays, active information gathering, persistent world models, or robust decisions across risk, time scales, and competing objectives.
It may also fail to learn reusable structure before reward makes that structure immediately valuable, or fail to compose previously learned skills into new plans.
These concerns connect CRL to active sensing, model-based control, multi-objective decision making, reward-free learning, and temporal abstraction \cite{bajcsy1988,hafner2025,hayes2022,jin2020,sutton1999}.
They are often studied separately, although a continual agent must integrate them; Schaul et al.\ explicitly caution that the challenges of continual learning are unlikely to be solved entirely in isolation \cite{schaul2018}.

The same fragmentation appears in evaluation.
Existing continual-learning and CRL benchmarks make important dimensions measurable, but each necessarily isolates a subset of the target \cite{wolczyk2021,powers2022,sharma2022,platanios2020}.
Success on any one benchmark, or a high aggregate return across several tasks, does not show that all capabilities required by the target setting coexist in one agent throughout a persistent deployment.
This paper therefore asks a design-level question: \emph{what properties should inform the design and evaluation of agents intended for continual reinforcement learning?}
Our answer is an opinionated set of ten desiderata spanning behavioral capabilities, system properties, and protocol conditions.
The desiderata are not architectural prescriptions; they organize discussion, expose missing assumptions, and make progress on individual capabilities reportable without turning one benchmark score into a claim of general lifelong competence.

\subsection{Target Setting and Terminology}

More specifically, our target is an agent that learns online during a continuing interaction, without externally supplied task identities or boundaries and without resets that erase state or action consequences, while operating under declared constraints on storage, model capacity, and computation.
Environment changes may be abrupt or gradual, earlier conditions may recur, and the consequences of the agent's actions may persist.
Our target setting adds concrete protocol and resource conditions to Abel et al.'s broader definition of CRL, under which the best agents must continue learning indefinitely \cite{abel2023}.

We use reinforcement learning here as an agent-level problem setting in which an agent learns through interaction to choose actions in pursuit of objectives, not as a synonym for model-free policy learning.
Accordingly, a CRL system may learn predictive models or reusable representations without immediate reward and may select actions through planning, including model-predictive control \cite{sutton1990,sutton2018}.
Its individual components need not each be reinforcement-learning algorithms.

Following earlier work that uses \emph{desiderata} to organize requirements and research directions for continual learning \cite{schwarz2018,khetarpal2022,verwimp2024}, we use the term for properties that we regard as desirable in our target setting, not as necessary conditions for CRL.
An agent might achieve CRL without satisfying every desideratum, and satisfying all ten would not by itself prove CRL.
Together, the desiderata provide a structured design-and-evaluation perspective: they help state which properties a system targets, which it omits, and which trade-offs it accepts.
Existing methods may fully satisfy one or several individual desiderata.
Our claim is not that every desideratum remains individually unsolved, but that we are not aware of any published evaluation demonstrating a single agent that satisfies all ten simultaneously within our target setting.

\subsection{Three Contributions}

This paper is deliberately selective.
For a broad research agenda for CRL, see Schaul et al.\ \cite{schaul2018}; for a recent application-driven perspective on continual learning and its future directions, see Verwimp et al.\ \cite{verwimp2024}; for a review and taxonomy of CRL, see Khetarpal et al.\ \cite{khetarpal2022}; for a definition of CRL, see Abel et al.\ \cite{abel2023}; and for a domain-agnostic framework for characterizing lifelong-learning systems and their trade-offs, see Baker et al.\ \cite{baker2023}.
Against this background, we make three concrete contributions:
\begin{enumerate}
    \item We develop and motivate an opinionated set of ten desirable properties for the design and evaluation of CRL systems, spanning sustainable adaptation, bounded resources, information acquisition and memory, world knowledge, and long-horizon decision making and planning.
    \item We relate each desideratum to representative prior work, distinguishing demonstrated capabilities from the assumptions under which the evidence holds and identifying what remains missing from an integrated system.
    \item We complement the desiderata with a compact persistent-world scenario that illustrates how a simple set of interacting rules can make several behavioral properties relevant within the same task-free, non-episodic online lifetime.
    Together, the suggested configuration, evaluation protocol, and capability-specific readouts define its proposed use as a benchmark while keeping the resulting evidence tied to particular properties instead of a single aggregate score.
\end{enumerate}
Our aspiration is practical: we want the desiderata to give researchers a map of the CRL design space and to help guide the development of learning systems that integrate capabilities current work largely studies in isolation.

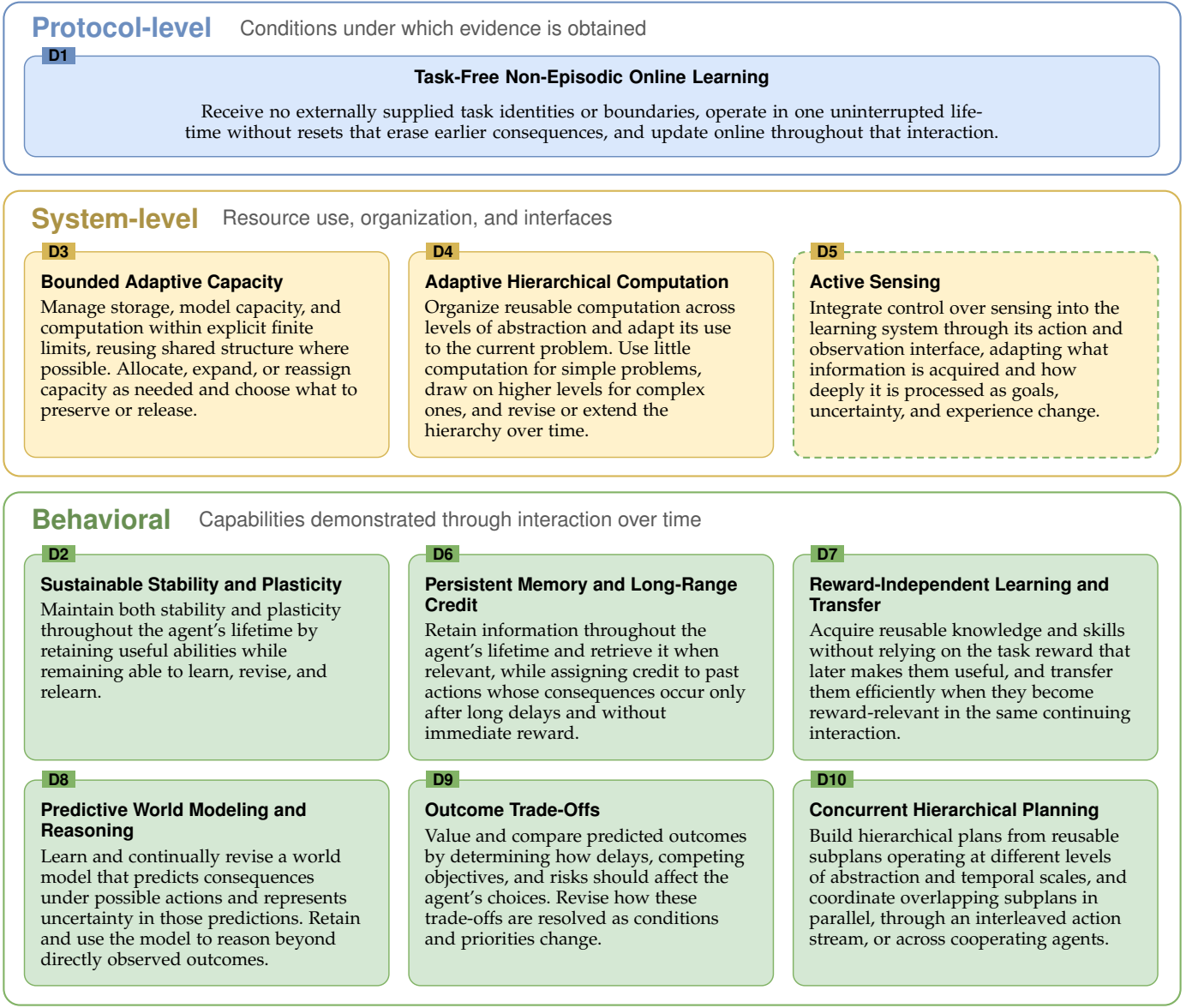
\begin{figure*}
    \centering
    \begingroup
    \newcommand{\overviewtitle}[1]{\vspace*{0.5em}{\sffamily\bfseries #1}\par\vspace{0.07cm}}
    \newcommand{\overviewstatement}[1]{#1\par}
    \newcommand{\overviewcardheight}{2.75cm}
    \newcommand{\overviewcardbody}[2]{
        \begin{minipage}[t][#1][t]{0.273\textwidth}
            \vspace*{0cm}
            \raggedright
            #2
        \end{minipage}
    }
    \newcommand{\systemcardbody}[1]{\overviewcardbody{\overviewcardheight}{#1}}
    \newcommand{\behavioralcardbody}[1]{\overviewcardbody{\overviewcardheight}{#1}}

    \begin{tikzpicture}[
        x=\textwidth,
        y=1cm,
        overview group/.style={rounded corners=0.25cm, line width=0.8pt},
        overview card/.style={
            rounded corners=0.18cm,
            line width=0.55pt,
            align=left,
            anchor=north west,
            inner xsep=0.25cm,
            inner ysep=0.21cm,
            font=\footnotesize
        },
        protocol card/.style={overview card, draw=protocolbriefframe, fill=protocolbriefback},
        system card/.style={overview card, draw=systembriefframe, fill=systembriefback},
        boundary card/.style={
            overview card,
            draw=behaviorbriefframe,
            fill=systembriefback,
            line width=0.8pt,
            dash pattern=on 3pt off 2pt
        },
        behavioral card/.style={overview card, draw=behaviorbriefframe, fill=behaviorbriefback},
        card tag/.style={
            anchor=west,
            rounded corners=0cm,
            inner xsep=0.10cm,
            inner ysep=0.04cm,
            outer sep=0cm,
            xshift=0.28cm,
            font=\scriptsize\bfseries\sffamily,
            text=black
        },
        protocol tag/.style={card tag, fill=protocolbriefframe},
        system tag/.style={card tag, fill=systembriefframe},
        behavioral tag/.style={card tag, fill=behaviorbriefframe},
        group title/.style={anchor=north west, font=\large\bfseries\sffamily},
        group description/.style={anchor=north west, font=\small\sffamily, text=black!65}
    ]
        \draw[overview group, draw=protocolbriefframe] (0.001,0) rectangle (0.999,-2.66);
        \node[group title, text=protocolbriefframe] at (0.018,-0.12) {Protocol-level};
        \node[group description] at (0.195,-0.15) {Conditions under which evidence is obtained};
        \node[protocol card, align=center, text width=0.935\textwidth] (d1) at (0.018,-0.82) {
            \overviewtitle{Task-Free Non-Episodic Online Learning}
            \overviewstatement{Receive no externally supplied task identities or boundaries, operate in one uninterrupted lifetime without resets that erase earlier consequences, and update online throughout that interaction.}
        };
        \node[protocol tag] at (d1.north west) {D1};

        \draw[overview group, draw=systembriefframe] (0.001,-2.91) rectangle (0.999,-7.31);
        \node[group title, text=systembriefframe!80!black] at (0.018,-3.05) {System-level};
        \node[group description] at (0.18,-3.08) {Resource use, organization, and interfaces};

        \node[system card] (d3) at (0.018,-3.84) {
            \systemcardbody{
                \overviewtitle{Bounded Adaptive Capacity}
                \overviewstatement{Manage storage, model capacity, and computation within explicit finite limits, reusing shared structure where possible.
                Allocate, expand, or reassign capacity as needed and choose what to preserve or release.}
            }
        };
        \node[system tag] at (d3.north west) {D3};
        \node[system card] (d4) at (0.344,-3.84) {
            \systemcardbody{
                \overviewtitle{Adaptive Hierarchical Computation}
                \overviewstatement{Organize reusable computation across levels of abstraction and adapt its use to the current problem.
                Use little computation for simple problems, draw on higher levels for complex ones, and revise or extend the hierarchy over time.}
            }
        };
        \node[system tag] at (d4.north west) {D4};
        \node[boundary card] (d5) at (0.670,-3.84) {
            \systemcardbody{
                \overviewtitle{Active Sensing}
                \overviewstatement{Integrate control over sensing into the learning system through its action and observation interface, adapting what information is acquired and how deeply it is processed as goals, uncertainty, and experience change.}
            }
        };
        \node[system tag] at (d5.north west) {D5};

        \draw[overview group, draw=behaviorbriefframe] (0.001,-7.56) rectangle (0.999,-15.47);
        \node[group title, text=behaviorbriefframe!80!black] at (0.018,-7.70) {Behavioral};
        \node[group description] at (0.16,-7.73) {Capabilities demonstrated through interaction over time};

        \node[behavioral card] (d2) at (0.018,-8.51) {
            \behavioralcardbody{
                \overviewtitle{Sustainable Stability and Plasticity}
                \overviewstatement{Maintain both stability and plasticity throughout the agent's lifetime by retaining useful abilities while remaining able to learn, revise, and relearn.}
            }
        };
        \node[behavioral tag] at (d2.north west) {D2};
        \node[behavioral card] (d6) at (0.344,-8.51) {
            \behavioralcardbody{
                \overviewtitle{Persistent Memory and Long-Range Credit}
                \overviewstatement{Retain information throughout the agent's lifetime and retrieve it when relevant, while assigning credit to past actions whose consequences occur only after long delays and without immediate reward.}
            }
        };
        \node[behavioral tag] at (d6.north west) {D6};
        \node[behavioral card] (d7) at (0.670,-8.51) {
            \behavioralcardbody{
                \overviewtitle{Reward-Independent Learning and Transfer}
                \overviewstatement{Acquire reusable knowledge and skills without relying on the task reward that later makes them useful, and transfer them efficiently when they become reward-relevant in the same continuing interaction.}
            }
        };
        \node[behavioral tag] at (d7.north west) {D7};

        \node[behavioral card] (d8) at (0.018,-11.99) {
            \behavioralcardbody{
                \overviewtitle{Predictive World Modeling and Reasoning}
                \overviewstatement{Learn and continually revise a world model that predicts consequences under possible actions and represents uncertainty in those predictions.
                Retain and use the model to reason beyond directly observed outcomes.}
            }
        };
        \node[behavioral tag] at (d8.north west) {D8};
        \node[behavioral card] (d9) at (0.344,-11.99) {
            \behavioralcardbody{
                \overviewtitle{Outcome Trade-Offs}
                \overviewstatement{Value and compare predicted outcomes by determining how delays, competing objectives, and risks should affect the agent's choices.
                Revise how these trade-offs are resolved as conditions and priorities change.}
            }
        };
        \node[behavioral tag] at (d9.north west) {D9};
        \node[behavioral card] (d10) at (0.670,-11.99) {
            \behavioralcardbody{
                \overviewtitle{Concurrent Hierarchical Planning}
                \overviewstatement{Build hierarchical plans from reusable subplans operating at different levels of abstraction and temporal scales, and coordinate overlapping subplans in parallel, through an interleaved action stream, or across cooperating agents.}
            }
        };
        \node[behavioral tag] at (d10.north west) {D10};
    \end{tikzpicture}
\endgroup
    \vspace{-1.25\baselineskip}
    \caption{Overview of the ten desiderata, grouped by their primary evidential perspective.
    \Dref{1} defines the interaction protocol under which evidence for all others is interpreted.
    \Dref{3}--\Dref{5} primarily concern resource use, organization, and interfaces of the complete learning system and require evidence beyond interaction outcomes, whereas \Dref{2} and \Dref{6}--\Dref{10} have behavioral cores that can be probed through what one agent demonstrates throughout a continuing interaction.
    This grouping neither partitions the desiderata into independent properties nor imposes an ordering among them; the desiderata interact.
    The dashed outline of \Dref{5} indicates its behavioral component despite its primarily system-level placement.}
    \label{fig:overview}
\end{figure*}

\section{Ten Desiderata}
\label{sec:10d}

The ten desiderata are the central contribution of this paper, and we denote them \Dref{1} through \Dref{10}.
They form an opinionated decomposition of properties relevant to the design and evaluation of a long-lived CRL agent.
They are neither a minimal definition of CRL nor ten independent gates that an agent must pass.
Their boundaries are intended to keep distinct design questions, sources of evidence, and failure modes visible.
The desiderata therefore overlap and constrain one another, individual methods may contribute to several, and we do not assume that each admits a unique scalar measure.
\Cref{fig:overview} provides a visual overview and summary of the proposed desiderata.

We classify the desiderata as \emph{protocol-level}, \emph{system-level}, or \emph{behavioral} according to the primary form of evidence required for each.
The protocol-level \Dref{1} specifies the interaction conditions under which evidence for the remaining properties is obtained.
A behavioral core means that the property is established primarily by what the same agent demonstrates throughout a continuing interaction; it does not imply that the supporting mechanism must itself be learned.
System-level desiderata additionally concern the organization, interfaces, or resource use of the complete learning system and therefore require evidence beyond observed outcomes.
Under this classification, \Dref{3}--\Dref{5} are primarily system-level, while \Dref{2} and \Dref{6}--\Dref{10} have behavioral cores.
These are primary perspectives rather than disjoint categories: \Dref{2} is behavioral even when stability and plasticity arise from fixed machinery, while \Dref{5} lies near the boundary because active sensing has a behavioral expression but depends on the sensing interface and its resource use.
We place \Dref{2} before the system-level properties because sustainable stability and plasticity support every capability that follows.

The literature shows a mild inflation in the number of desiderata: Schwarz et al.\ list five for continual-learning methods, Khetarpal et al.\ seven for continual learners, and we arrive at ten for CRL agents and their evaluation \cite{schwarz2018,khetarpal2022}.
The increasing count should not be read as a succession of stricter definitions or as an attempt to construct a target that every existing system must miss.
The lists differ in scope and granularity, although they share concerns about task-agnostic learning, stability, transfer, and scalability.
For example, we combine Khetarpal et al.'s separate requirements of online, task-agnostic, and incremental learning without a fixed training set into the protocol of \Dref{1}.
Conversely, \Dref{8}--\Dref{10} could be grouped under a broad notion of reasoning and planning, but predicting possible consequences, comparing them, and organizing actions to bring selected outcomes about can fail independently and require different evidence.
Our list is therefore a working decomposition: each desideratum should express a coherent question, while the divisions should remain fine enough to expose important differences in mechanisms, evidence, and failure.

We use representative prior work to clarify each desideratum, identify what has been demonstrated under which assumptions, and explain what remains missing from an integrated CRL system.
The mechanisms discussed are candidates rather than architectural requirements.
The following subsections are deliberately self-contained and need not be read as one uninterrupted checklist (a one-coffee-per-desideratum reading schedule is entirely defensible).
We begin with \Dref{1} because it sets the protocol under which evidence for all subsequent desiderata is interpreted.

\subsection{Task-Free Non-Episodic Online Learning}
\label{sec:d1}
\begin{briefbox}[protocolbrief]{D1}
    \emph{\Dref{1} specifies the protocol under which continual learning is claimed: task-free, non-episodic online learning.
    The complete learning system should therefore (1) receive no externally supplied task identities or boundaries, (2) operate in one uninterrupted lifetime without resets that restore a predefined state or erase the consequences of earlier actions, and (3) update online throughout that interaction.}
\end{briefbox}

\noindent The starting point for our desiderata is the learning protocol, which specifies the conditions under which an agent interacts and learns throughout its lifetime.
Our target protocol combines three distinct properties: it is task-free with respect to externally supplied task structure, non-episodic with respect to the continuity of interaction and its consequences, and online in that learning continues during deployment.
None of these properties implies the other two, so \Dref{1} requires all three.
Individual capabilities may be studied outside this protocol, but evidence that they are integrated in a CRL system must be obtained under all three conditions.

\noindent\textbf{(1) Task-free.}
Under \Dref{1}, the complete learning system receives neither externally supplied task identities nor boundary signals, even when the environment contains changing objectives or recurring regimes \cite{aljundi2019,caccia2023}.
These signals provide distinct forms of privileged information: a task identity can guide the acting system's choice of behavior, while a task boundary can signal when the learning process should consolidate knowledge or reorganize experience.
However, task-free does not mean objective-free; an instruction from a person or another agent may be part of ordinary interaction when it communicates a desired outcome without exposing an externally defined task identity or boundary.
Withholding them does not prevent the agent from inferring latent contexts or imposing its own segmentation, but any such structure must emerge from its interaction history.
Within this protocol, an objective that becomes relevant may call for knowledge and skills acquired before its task reward made them useful; acquiring and transferring such reusable structure is the concern of \Dref{7}.
Pursuing that objective may then require the agent to organize reusable skills and subplans into a hierarchy and coordinate them with other ongoing plans, which is the concern of \Dref{10}.
Neither behavioral capability follows merely from the absence of supplied task identities and boundaries under \Dref{1}.

\noindent\textbf{(2) Non-episodic.}
Episodic protocols divide interaction into trials that terminate and restart \cite{sutton2018}.
This organization makes learning easier and more stable: resets allow trajectories to be sampled repeatedly from controlled initial conditions, while episode boundaries can signal when to consolidate knowledge, reorganize experience, or switch behavior.
These conveniences are excluded under \Dref{1}, which places the agent in one persistent lifetime without evaluator-controlled resets to predefined states \cite{naik2019,lu2021}.
In this setting, action consequences persist across the interaction, and the relevance of earlier events must be inferred without externally supplied episode boundaries.
An intervening reset could replenish resources, undo environmental modifications, or rescue the agent from unfavorable states, thereby breaking links between earlier actions and later observations.
If such a reset is not revealed, stable world dynamics may also appear inconsistent, for example when an exhausted resource silently reappears.
Conversely, continuous interaction in a non-episodic setting avoids these discontinuities and offers a further benefit: it preserves causal and informational relationships that episodic partitioning may erase.
For example, information that initially appears irrelevant may become useful only in combination with an event or resource encountered much later.

At the same time, maintaining uninterrupted interaction across the agent's lifetime makes learning harder.
Experience arrives as one temporally correlated and potentially drifting stream in which earlier situations cannot generally be resampled at will.
The learner must therefore control catastrophic forgetting without relying on episode-wise shuffling to approximate i.i.d.\ training \cite{aljundi2019,rolnick2019}.
The substantive distinction therefore lies in what resets and boundaries make available to the learner.
Episode counts and implementation labels alone are insufficient.
Naik et al.\ connect continuing tasks to lifelong learning because an agent cannot normally return to a predefined initial state \cite{naik2019}.
Their example of very long episodes shows that the episode label alone does not determine whether a boundary matters to the agent.
Accordingly, \Dref{1} treats a boundary as substantive only when the acting or learning system depends on its signal, or when the associated reset restores state, enables repeated sampling, or erases earlier consequences.
Whether a finite uninterrupted run is recorded as one episode or several segments is irrelevant to \Dref{1} if those divisions neither inform the system nor reset the interaction.

\noindent\textbf{(3) Online.}
Under the conventional separation between training and inference, a model is optimized on a fixed or repeatedly sampled data distribution and subsequently deployed with its learned behavior frozen.
\Dref{1} instead interleaves experience, action, and updating, so that later behavior can incorporate information that became available only during the continuing interaction.
Online updating may change model parameters, memory, or other internal state, but the relevant criterion is whether new evidence can continue to influence future behavior when needed.

The combination of task-free operation, non-episodic interaction, and online learning makes \Dref{1} fundamental to our account of CRL by defining the protocol under which the remaining desiderata must be integrated.
Success under this protocol cannot rely on human-supplied segmentation, offline training followed by fixed deployment, or resets that erase earlier consequences.
Existing work approaches this protocol both by combining its three conditions in focused experiments and by treating individual conditions separately.
Anand and Precup provide a notable example of the former in Jelly Bean World (JBW) \cite{platanios2020}, which also provides the environmental basis for our playable companion in Section~\ref{sec:benchmark-proposal}.
In their control experiment, PT-DQN learns online during one uninterrupted run while the reward mapping changes on a fixed but unrevealed schedule \cite{anand2023}.
The method receives neither task identities nor boundary signals, and the world is not reset between changes.
PT-DQN therefore instantiates all three parts of the \Dref{1} protocol in a focused stability--plasticity experiment and is discussed further under \Dref{2}.

Other methods cover only parts of \Dref{1}, including approaches that reduce reliance on supplied task identities and boundaries during online learning.
One Bayesian example comes from Zeno et al., whose fixed-point online variational update handles a changing supervised stream without known task boundaries or an external replay buffer \cite{zeno2021}.
The method shows that earlier knowledge can be protected without a supplied task-boundary signal.
However, its evaluation uses supervised prediction streams and does not test an agent acting in a persistent environment.
Aljundi et al.\ address the same problem by adapting Memory Aware Synapses (MAS) \cite{aljundi2018,aljundi2019}.
MAS estimates each parameter's importance from the sensitivity of the learned function to changes in that parameter, and penalizes subsequent changes to important parameters, creating a soft distinction between relatively stiff degrees of freedom supporting learned outputs and more plastic ones available for new learning.
Where the original method updates importance after each task, Aljundi et al.\ ask when importance should be re-estimated, which data should support the estimate, and how successive estimates should be accumulated.
Their method uses a plateau in the observed loss to trigger an update of the parameter-importance estimates, computes the update from a small buffer of difficult examples, and combines successive estimates with a cumulative moving average.
This makes consolidation boundary-free and online, although the objective and output interface remain fixed and learning still relies on target signals accompanying the observations.
Veness et al.'s Gated Linear Networks (GLNs) provide a complementary supervised example in which fixed context gates select locally updated parameters during a single online pass without task-boundary input \cite{veness2021}.
Their stability--plasticity results are discussed under \Dref{2}.
For \Dref{1}, GLNs provide evidence for task-free online prediction, and Sezener et al.\ extend them to contextual bandits, providing an early success in a restricted reinforcement-learning setting with fully online action selection \cite{sezener2020}.
Contextual bandits have no action-dependent state dynamics, however, so this result does not yet show an agent learning through continuing interaction with a persistent environment.

In reinforcement learning, Caccia et al.'s 3RL \cite{caccia2023} combines the off-policy soft actor--critic algorithm \cite{haarnoja2018} with a recurrent neural network and experience replay, a standard continual-learning mechanism for mitigating forgetting \cite{rolnick2019}.
The recurrent representation summarizes recent states, actions, and rewards for the actor and critics, allowing them to infer latent task context and adapt without receiving a task identity, while replay of trajectories from previous tasks mitigates catastrophic forgetting.
The results also suggest that the recurrent context can relate a new task to components of earlier ones, supporting forward transfer.
The acting policy is therefore task-free and learning remains online, but the reported replay scheme prioritizes current experience using separate current-task and old-task buffers and is therefore task-aware at the level of training.
As an alternative, the authors suggest that oversampling recently collected experience could remove the need for an explicit task-change signal.
The environments nevertheless retain ordinary episodes and resets, so 3RL does not establish the non-episodic component of \Dref{1}.

Other methods target the non-episodic component by reducing online interaction's dependence on external resets.
Eysenbach et al.\ introduce the Leave No Trace method \cite{eysenbach2018}, which jointly learns a forward policy and a reset policy that returns the environment to reusable states, and uses the reset policy's value estimate to avoid transitions from which autonomous recovery is unlikely.
This substantially reduces manual intervention, but it retains an explicit decomposition into forward and reset objectives and repeated attempts.
By contrast, Lu et al.'s LiSP framework goes further toward non-episodic learning: rather than relying on a reset after failure, the agent remains operational while a learned model, reusable skills, and skill-space planning help it avoid states or regions from which autonomous recovery is unlikely \cite{lu2021}.
Its online Lifelong Hopper and Ant tests continue across scheduled target-velocity changes without resets or an additional task identity or boundary signal.
These tests instantiate \Dref{1} under our distinction between supplied objectives and task structure, while assuming an initial dataset and a known reward function.
We discuss its model learning, skill discovery, and hierarchical planning in more detail under \Dref{10}.

Taken together, these works provide evidence for the complete \Dref{1} protocol in focused control settings and for its individual components under broader assumptions.
In Section~\ref{sec:benchmark-coverage}, we examine how closely existing benchmarks instantiate this three-part protocol.

\subsection{Sustainable Stability and Plasticity}
\label{sec:d2}

\begin{briefbox}[behaviorbrief]{D2}
    \emph{\Dref{2} asks an agent to maintain both stability and plasticity throughout its lifetime by retaining useful abilities while remaining able to learn, revise, and relearn.
    Immutability may preserve earlier abilities but prevents further learning, while excessive plasticity may erase abilities that still matter.}
\end{briefbox}

\noindent Stability should not be confused with immutability.
Change is necessary when the environment changes, and even in a stationary environment, further learning or consolidation may alter the system in ways that make its useful behavior more stable.
For \Dref{2}, what matters is whether a change harms abilities that remain applicable or may be needed again, not whether it overwrites parameters, predictions, or stored records.
Plasticity allows a learner to acquire new abilities and revise knowledge invalidated by new evidence, but the resulting updates may also damage abilities that remain applicable or may be needed again.
Carpenter and Grossberg describe this tension as the classical stability--plasticity dilemma, in which new experience must be incorporated without destabilizing useful existing representations \cite{carpenter1987}.
Continual learners can fail on either side of this dilemma: catastrophic forgetting occurs when learning from new experience damages useful abilities acquired earlier \cite{mccloskey1989,french1999}, whereas loss of plasticity occurs when changes accumulated through earlier learning make the agent increasingly unable to adapt to new situations through further learning \cite{dohare2024}.
Having enough model capacity to represent both old and new estimates does not by itself resolve this dilemma because the learner must still determine which estimates should persist across situations and which should track current conditions.
Anand and Precup \cite{anand2023} show that this issue is not specific to neural networks: even in tabular reinforcement learning, tracking current conditions requires overwriting earlier estimates, whereas preserving those estimates reduces accuracy on the current task.

A later decline in performance on a previously encountered problem indicates a failure of stability under \Dref{2} only if new learning has damaged an ability that remains useful.
Deliberately discarding or compressing low-value information under a declared resource budget is instead capacity management under \Dref{3}.
Performance can also decline without a failure of stability when the relevant ability remains intact but the learner cannot determine when it applies.
Kessler et al.\ distinguish this objective interference from catastrophic forgetting in settings where the same observation calls for incompatible actions under different objectives \cite{kessler2022}.
They also address the problem algorithmically by retaining the incompatible policies in separate heads and using reward feedback to select the applicable head during evaluation.
This allows their evaluation to separate the loss of an earlier policy from a failure to identify the current objective.
For our broader CRL target, however, using known training boundaries to trigger head creation and replay clearing leaves \Dref{1}'s task-free condition open, while adding one head per task causes linear model-capacity growth and leaves \Dref{3}'s bounded-capacity requirement open.

Correctly attributing performance loss is only one part of evaluating \Dref{2}.
Stability and plasticity must also be assessed jointly over time because a method can preserve earlier abilities by restricting further learning or promote new learning at the cost of forgetting.
On the stability side, the stability gap occurs when performance on earlier tasks drops sharply at the start of training on a new task and partially recovers during subsequent updates \cite{delange2023}.
Evaluations performed only at task endpoints can therefore miss this transient forgetting.
On the plasticity side, retaining earlier abilities says little if the system can no longer acquire anything new.
Evaluating \Dref{2} therefore requires performance measurements throughout transitions and tests of new learning at progressively later lifetime stages \cite{dohare2024}.

We group existing methods contributing to stability, plasticity, or both into five families.
The first family uses parameter and functional consolidation to constrain changes to learned parameters or functions.
EWC uses the diagonal Fisher information to estimate each parameter's importance to a learned task and penalizes changes to important parameters more strongly during subsequent learning \cite{kirkpatrick2017,huszar2018}.
SI tracks each parameter's contribution to loss reduction along the optimization trajectory \cite{zenke2017}, and MAS measures the sensitivity of learned outputs to parameter changes \cite{aljundi2018}.
The task-free extension of MAS changes when and from which data these importance estimates are updated \cite{aljundi2019}.
Zeno et al.'s fixed-point online variational method, also discussed under \Dref{1} for its task-free online updates, provides a Bayesian counterpart by carrying an approximate Gaussian weight posterior from one minibatch to the next \cite{zeno2021}.
Its uncertainty estimates protect high-confidence weights while leaving other weights more plastic, providing finite supervised evidence for boundary-free retention but not sustainable plasticity over a lifetime.
Learning without Forgetting instead constrains the function directly by distilling the old network's responses on new-task data \cite{li2018}, while Encoder Based Lifelong Learning combines such an output constraint with task-specific undercomplete autoencoders that protect low-dimensional feature subspaces \cite{rannen2017}; like the parameter-importance methods above, both mainly preserve existing input--output behavior.
Leaving some model capacity available for new learning demonstrates local plasticity, but accumulated protection without a learned criterion for relaxing obsolete constraints does not establish sustainable revision over a lifetime.

The second family uses rehearsal to protect previously learned abilities by reintroducing information about earlier learning during later updates.
Experience replay is its most direct form, revisiting stored experience rather than estimating which parameters should remain fixed.
In reinforcement learning, Rolnick et al.'s CLEAR method implements this idea by combining on-policy learning from current experience with off-policy learning and behavioral cloning on replayed experience, explicitly targeting plasticity and stability without task identities or boundaries \cite{rolnick2019}.
In experiments that cycle through several reinforcement-learning tasks, CLEAR retains earlier performance and acquires a new probe task at a similar rate when it is introduced at progressively later stages.
This provides unusually direct evidence for \Dref{2}, although the episodic protocol covers only a small task family and does not test revision when old behavior becomes invalid.
Wo{\l}czyk et al.\ likewise show that behavioral cloning can reduce forgetting while improving forward transfer and that reusing earlier policies for exploration provides an additional transfer benefit \cite{wolczyk2022}.
However, their experiments assume known task transitions and use task-specific output heads and retained data from earlier tasks.
GEM instead uses stored examples from earlier tasks to constrain gradient steps, modifying any step that would increase their loss \cite{lopezpaz2017}.
Deep Generative Replay instead interleaves current data with pseudo-examples from a learned generator, replacing stored examples with generator capacity and training computation in finite classification sequences \cite{shin2017}.
CORe3, discussed under \Dref{9}, extends rehearsal beyond stored or generated examples by using a learned reward model to reconstruct unavailable reward signals for earlier objectives on current experience, allowing the policy to continue training toward those objectives without replaying earlier transitions \cite{li2024a}.
The protection supplied by these forms of rehearsal depends on what the stored or generated experience and reconstructed signals preserve from earlier learning.
At the same time, different rehearsal mechanisms shift the cost of stability among storage, model capacity, and computation rather than eliminating it.
Beyond these resource trade-offs, much of the evidence for consolidation and rehearsal comes from finite supervised vision or prediction protocols organized around announced tasks or task-specific output layers \cite{parisi2019,aljundi2019}.
Even the robot-control experiment in task-free MAS learns by imitating a sensor-based expert rather than from rewards.
EWC broadens the evidence to sequential Atari control, but its complete system still relies on scheduled game blocks, task recognition, game-specific biases and gains, and separate replay buffers.
These studies therefore demonstrate retention under their respective protocols, not sustained stability and plasticity under the complete \Dref{1} setting.

The third family uses structural separation to limit interference between old and new learning.
PackNet \cite{mallya2018} prunes a shared network after each task, freezing the retained weights and making the remaining capacity available for later tasks.
HAT \cite{serra2018} instead learns task-conditioned attention masks that restrict which units each task uses and updates.
Veniat et al.\ explicitly broaden continual-learning evaluation beyond forgetting to include transfer and the scaling of storage and computation, and propose a modular method that reuses earlier modules or instantiates new ones under a task-driven prior \cite{veniat2021}.
Freezing allocated parameters or modules provides strong retention at the cost of later revision, while reliance on supplied task boundaries and identities does not meet \Dref{1}'s task-free requirement.
These methods also expose a trade-off in bounded model capacity under \Dref{3}: parameter isolation progressively consumes a fixed pool, whereas modular expansion increases model size as new tasks arrive.
A different approach to parameter separation is provided by GLNs, introduced under \Dref{1}.
Instead of relying on task identities, they use a boundary-free and backpropagation-free form of input-dependent parameter selection.
Fixed gates map the current input directly to a context and select one weight vector per neuron.
Across layers, these vectors form a path of input-dependent weight matrices through which the prediction is computed.
Each neuron directly predicts the target and can therefore update its selected weights locally through online convex optimization, without backpropagating errors through the network.
This combination of contextual parameter separation and local online learning makes GLNs strongly resilient to catastrophic forgetting without requiring task identities or boundaries \cite{veness2021}.
However, GLNs are general-purpose learning architectures rather than complete CRL methods.
They provide an alternative to conventional neural networks trained with backpropagation and could serve as one component of a broader CRL agent.

The fourth family uses associative memory to retain learned representations, with modern Hopfield networks serving as a representative example.
These comprise several high-capacity formulations, including the higher-order Dense Associative Memory (DAM) \cite{krotov2016} and the continuous formulation of Ramsauer et al.\ \cite{ramsauer2021}.
McAlister et al.\ study the former and show that it can suffer catastrophic forgetting under sequential training while evaluating standard mitigation methods based on rehearsal, replay-constrained gradients, and parameter regularization \cite{mcalister2025}.
These methods improve the measured stability--plasticity balance of the DAM over the evaluated sequence, but have not been shown to satisfy \Dref{2} sustainably.
These results do not directly test Ramsauer et al.'s formulation; their broader implication is that high associative capacity alone should not be taken as evidence of a continual-learning rule.
We return to modern Hopfield networks as candidate persistent-memory mechanisms under \Dref{6}.

The fifth family focuses directly on maintaining or recovering adaptability.
Anand and Precup instantiate permanent--transient value learning in PT-DQN \cite{anand2023}.
It decomposes each action-value estimate into a slowly changing permanent component, which captures structure recurring across experience, and a transient component, which rapidly learns corrections for the current conditions.
The two estimates are added when selecting actions, so stability and adaptation contribute to one policy rather than being handled by separate task-specific agents.
In the boundary-free variant, the permanent component is updated at fixed intervals and the transient component is decayed rather than reset at an announced task change.
PT-DQN therefore provides a concrete task-agnostic RL mechanism that addresses stability and adaptation jointly.
Its finite experiments on value prediction and discrete-action control under scheduled non-stationarity provide partial evidence for \Dref{2}, but do not establish sustainable stability and plasticity over an open-ended lifetime.
This limited scope is deliberate: PT-DQN is a value-learning component rather than a complete agent architecture, and its permanent--transient decomposition could in principle be combined with complementary mechanisms that address additional desiderata.

Other methods in this family target adaptability through the learning process itself.
Luketina et al.'s contextualized meta-gradients use signals about agent performance and environmental change to adapt optimization hyperparameters \cite{luketina2022}.
They improve adaptation in non-stationary environments, but target how the learner updates rather than whether it retains previously learned abilities, leaving the stability side of \Dref{2} open.
Continual backpropagation selectively reinitializes low-utility units, releasing and reusing capacity while keeping network size fixed \cite{dohare2024}.
Its update rule requires neither task identities nor boundaries, and its extended experiments provide direct evidence of late-life learning ability.
It nevertheless targets loss of plasticity rather than catastrophic forgetting: the utility-based replacement rule has no explicit objective for retaining earlier abilities that current experience no longer exercises.
Moreover, the supervised benchmarks announce task or class increments and either reset the classification head or add output units at those boundaries, while the reinforcement-learning benchmarks use ordinary episodic environment resets.
These protocols therefore do not establish the complete \Dref{1} setting.

Taken together, these mechanisms address different parts of the stability--plasticity requirement and are therefore complementary rather than interchangeable.
Parameter and functional consolidation protect established behavior, rehearsal reintroduces earlier experience or learning signals, structural separation isolates capacity, associative memory supports retrieval, permanent--transient decomposition balances slow retention with rapid adaptation, and adaptive optimization or unit recycling support continued plasticity.
None of the results reviewed here establishes \Dref{2} as a whole.
Evidence should jointly track retention, revision, reacquisition, and new learning at multiple lifetime stages.
Retaining earlier abilities does not show that learning can continue, while successful late-life acquisition does not show that those abilities remain available, especially when the learner receives newly allocated task-specific components.
Because these mechanisms can trade resources for stability or plasticity, their compatibility with \Dref{3} depends on how storage, model capacity, and computation scale over the lifetime.

\subsection{Bounded Adaptive Capacity}
\label{sec:d3}

\begin{briefbox}[systembrief]{D3}
    \emph{\Dref{3} asks a learning system to manage storage, model capacity, and computation within explicit finite limits as it encounters previously unknown problems.
    When problems share structure, compression and reuse should allow resource requirements to scale sublinearly as more such problems are encountered.
    Within its available budget, the system should allocate, expand, or reassign capacity as needed and deliberately choose what to preserve and what to release if it cannot retain everything.}
\end{briefbox}

\noindent Over a lifetime, retaining experience, preserving learned structure, and acquiring new abilities consume storage, model capacity, and computation.
\Dref{3} asks a learning system to manage these resources within explicit finite limits as it encounters previously unknown problems.
Within these limits, the system may expand, compress, or reassign capacity as its needs change, but unconstrained expansion merely postpones the resource problem.
When problems share structure, reuse should reduce the additional resources needed for each new problem.
Following Veniat et al., model capacity and computation should then grow sublinearly with the number of problems rather than linearly through the addition of an independent fixed-size model for each one \cite{veniat2021}.
Partially related problems may permit only partial reuse, while unrelated problems may still require new capacity, so \Dref{3} ties the expected resource savings to the amount of shared structure instead of demanding unconditional sublinear growth.

\Dref{3} complements \Dref{2} by asking whether stability and plasticity remain sustainable under finite resource limits.
A system can protect earlier abilities by retaining all earlier experience for replay or by allocating a frozen model for every new problem, yet eventually exhaust its storage, model capacity, or computational budget.
Conversely, remaining within a fixed budget says little if the system forgets or ceases to learn.
Without such a strategy, consolidation and isolation eventually leave too little unconstrained capacity for further learning, while replay and architectural expansion eventually exhaust their storage, model-capacity, or computational budgets.
\Dref{3} therefore asks what the complete system does when these resources become scarce.

The resource budget in \Dref{3} has three distinct dimensions: storage, model capacity, and computation.
Here, we use \emph{storage} for retained data such as replay samples and external records, \emph{model capacity} for learned parameters and modules, and \emph{computation} for the processing required to update and operate the learning system.
We reserve the term \emph{memory} for the behavioral capability under \Dref{6} to preserve, retrieve, revise, and release information so that past experience can guide behavior.
Storage can support memory, but does not by itself establish that capability.
These resource dimensions interact without being interchangeable; replay, search, or repeated optimization can increase computation without adding stored data or model parameters.
These distinctions matter in practice because evaluations do not constrain all three resources equally.
Verwimp et al.\ find that stored samples are commonly bounded, whereas computational cost receives much less attention \cite{verwimp2024}.
Bounding retained data alone does not make a learning system resource-bounded if its model capacity or computational requirements can grow without limit.
Because every physical system faces finite limits on all three, \Dref{3} requires each limit to be explicit in the evaluation but does not prescribe how it is communicated to the learning system.
A concrete setting may supply a budget, expose resource pressure, or enforce bounded components; what matters is whether the complete system sustains learning within those limits.

The protocol-level \Dref{1} adds an important constraint to capacity management.
A task-free learner cannot wait for an announced task transition, use a supplied task identity to select a module, or allocate capacity simply because an external task counter advanced.
Its decisions to reuse, expand, compress, or release capacity must instead be driven by signals available from its own experience and learning process.
An evaluator may still count encountered problems when studying scalability, but that count must not become privileged input to the learning system.

The literature does not provide a tidy split between \Dref{2} and \Dref{3}.
To the best of our knowledge, existing work does not yet demonstrate \Dref{3} in full within the CRL setting considered here: managing storage, model capacity, and computation from experience under explicit finite budgets.
It does, however, provide useful pieces, including bounded replay, learned growth, parameter isolation, modular reuse, capacity recycling, and model compression.
Most of these mechanisms were developed at least partly to preserve earlier abilities or maintain plasticity, so their strong overlap with \Dref{2} is not accidental.

Replay provides a direct storage-bounded example.
CLEAR uses reservoir sampling to maintain an approximately uniform fixed-size sample of past experience and retains most of its performance with substantially smaller buffers \cite{rolnick2019}.
This demonstrates retention under a fixed storage budget without task information, but does not decide which experiences are especially valuable or obsolete, nor does it adaptively manage model capacity or computation.
Deep Generative Replay provides a complementary trade-off by moving much of the historical-data burden from retained samples to a learned generator, whose parameters and repeated training count against model-capacity and computation budgets \cite{shin2017}.

Constructive neural networks provide an early precedent for learned growth.
Cascade-Correlation starts with no hidden units and adds a new one when training no longer reduces residual error sufficiently, after which the new unit's incoming weights are frozen \cite{fahlman1990}.
It therefore learns the size and topology needed for one static supervised problem, but does not address how a continual agent should revise or reclaim capacity as its world changes.
Progressive Networks transfer features through lateral connections from frozen earlier columns while adding a complete new column for every announced task \cite{rusu2016}.
This prevents forgetting by construction and was evaluated in reinforcement learning, but the original architecture grows its hidden units linearly and its parameter count quadratically with the number of tasks, while selecting a column at inference requires the task label.
CORe3, discussed under \Dref{9}, provides a narrower reinforcement-learning example by sharing encoders while adding value and reward-model heads for every new objective \cite{li2024a}.
This is more economical than adding a complete network, but model capacity still grows linearly with the number of encountered objectives while expansion remains tied to supplied task changes.
Dynamically Expandable Networks make growth more selective by retraining relevant units, adding units only when the current task loss remains above a threshold, and splitting units that have drifted too far \cite{yoon2018}.
Their supervised protocol still announces each task, records the training stage at which each unit was added, and uses this information during task-specific inference.
It also provides no mechanism for reclaiming previously allocated capacity when resources become scarce.
Han et al.\ combine task-triggered allocation, pruning, and freezing within a single spiking network, reusing retained neurons while returning redundant newly allocated neurons to a fixed capacity pool \cite{han2023}.
Previously retained neurons remain frozen, however, and the procedure relies on externally separated classification tasks, leaving autonomous capacity revision and the \Dref{1} protocol open.

Modular approaches make reuse more explicit.
PathNet searches for a pathway through a fixed pool of modules, freezes the pathway selected for one task, and allows a later task to reuse parts of it \cite{fernando2017}.
This bounds total model capacity and limits active computation, but the evaluated transfer protocol explicitly switches tasks and does not determine what should happen when the fixed pool fills.
Veniat et al.\ instead search for a composition of frozen modules from earlier supervised tasks and newly instantiated modules, allowing related tasks to share a compact path through the network \cite{veniat2021}.
Their experiments on long sequences of tasks provide direct evidence that reuse can reduce the additional parameters and computation required for each new task, but task descriptors are supplied during both training and evaluation, and structural changes begin when a new task arrives.

CompoNet brings growable modularity into continual reinforcement learning by freezing each learned policy module and letting the next module compose the outputs of earlier policies \cite{malagon2024}.
Its parameter count grows linearly rather than quadratically with the number of tasks, and its experiments demonstrate transfer without sacrificing the newly added module's plasticity.
However, it introduces one module at every known task transition and assumes that task boundaries and identifiers are available.
Moreover, because each policy module depends on earlier modules, the paper's own complexity analysis gives quadratic theoretical inference cost in the number of tasks.
CompoNet is therefore a useful step from independent policies toward reusable growth, but it does not yet provide task-free allocation or prevent inference from becoming progressively more expensive, as required by \Dref{3}.

Even reusable growth can accumulate model capacity and computation over time, so \Dref{3} also asks whether previously allocated structure can be compressed or released.
Network pruning shows that many trained connections can be removed and the remaining network retrained without losing performance, while lottery-ticket experiments show that dense networks can contain much smaller subnetworks that remain trainable from suitable initializations \cite{han2015,frankle2019}.
These results establish that a learned function may require only a fraction of a dense model's parameters, but their standard train--prune--retrain procedures are offline and tied to fixed classification datasets.
They do not tell a continual learner what to retain when resource limits force a trade-off, how released capacity should be reused, or how to avoid damaging abilities that may be needed again.

Taken together, these methods show that capacity can be added selectively, partitioned, reused, recycled, or compressed.
None, however, manages the full capacity lifecycle required by \Dref{3}: deciding from experience when to apply each operation and what to preserve or release as resource use approaches its declared limits.

\Dref{3} does not prescribe growth, pruning, sparsity, modularity, or any other particular mechanism, and its evaluation need not inspect the learner's internal signals.
Instead, we can treat the complete learning system largely as a black box.
The behavioral \Dref{2} asks whether the system retains useful abilities while continuing to learn, whereas the system-level \Dref{3} asks whether it can sustain this ability under explicit limits on retained data and model capacity, together with a bound on computation per unit of interaction.
Evidence for \Dref{3} therefore comes from \Dref{2} continuing to hold when these resource limits actually constrain the learner.
The limits must be tight enough to make reuse, compression, or selective release necessary; otherwise, a sufficiently large fixed model can hide the capacity problem.
Evaluations across problem counts and resource budgets can then show whether the system accommodates related problems with sublinear resource growth.

\subsection{Adaptive Hierarchical Computation}
\label{sec:d4}

\begin{briefbox}[systembrief]{D4}
    \emph{\Dref{4} asks a learning system to organize reusable computation across levels of abstraction and adapt its use to the current problem.
    It should use little computation for simple problems and may draw on higher levels of abstraction for complex ones.
    Across its lifetime, it should be able to revise and extend this hierarchy.}
\end{briefbox}

\noindent Over its lifetime, a continual learner may face tasks that use the same observation modalities and action space while requiring computation at different levels of abstraction, from direct responses to multistep reasoning.
A functional hierarchy can accommodate this range by placing sensory and action interfaces at a shared low-abstraction level, reusing them across compatible tasks and extending them when new modalities or primitive actions become available.
The sensory and action interfaces can occupy separate modules and still anchor a hierarchy in which more complex tasks may engage higher levels of abstraction and additional reusable computation.
During interaction, the learning system should adapt both the amount of computation it uses and the parts of the hierarchy it engages.
For example, a visible obstacle might trigger a direct avoidance response, while deciding whether it blocks a route may recruit object recognition and spatial reasoning at higher levels.
Across its lifetime, the system should revise and extend the hierarchy when its existing computations are insufficient.
Of course, what counts as difficult may also change over the agent's lifetime, since learning may create lower-level computations that now allow the agent to handle a previously demanding problem through a shorter path in the hierarchy.

Decisions about using the hierarchy during interaction and revising it over a lifetime connect the system-level \Dref{4} to preceding desiderata of different types.
Under the protocol-level \Dref{1}, they must be driven by the agent's experience rather than supplied task identities or boundaries.
The behavioral \Dref{2} requires the hierarchy to remain revisable without destroying useful abilities or its ability to learn, while the system-level \Dref{3} limits the capacity and computation used to maintain it.
Hierarchy also appears in the behavioral \Dref{10}, but with a different role:
\Dref{4} concerns how the system organizes, allocates, and extends computation, whereas \Dref{10} concerns whether the agent can compose skills and coordinate plans across levels of abstraction and temporal scales.
A functional hierarchy may support this behavior efficiently, but its presence does not establish hierarchical planning, nor does such behavior reveal a particular internal organization.

Beyond helping an agent handle diverse tasks efficiently, a functional hierarchy may support the gradual construction of new capabilities over its lifetime.
Ring \cite{ring1997} argues that increasingly complex behavior can be constructed from previously learned skills.
Drawing on Piaget's constructivist account of cognitive development, Drescher \cite{drescher1991} develops a complementary view in which a learner begins with little prior world knowledge and creates new representational elements when its existing vocabulary becomes insufficient.
For \Dref{4}, the two accounts jointly motivate a functional hierarchy that begins modestly and becomes more capable by adding representational elements and composing previously learned skills as needed.

Elements of this design hypothesis are also consistent with broad findings from neuroscience, although their implications for artificial-agent design remain open.
Felleman and Van Essen describe the primate cortex as a distributed, reciprocal hierarchy, in which information flows between lower and higher areas in both directions instead of along a one-way processing chain \cite{felleman1991}.
Hasson et al.\ find that different cortical regions integrate information over progressively longer temporal windows \cite{hasson2008}, while Badre and D'Esposito report a prefrontal hierarchy of increasingly abstract action selection \cite{badre2007}.
Together, these findings suggest that biological hierarchies may organize processing by both temporal context and abstraction.
Work on cognitive control provides a possible biological analogue of adaptive selection.
Miller and Cohen propose that goal representations in the prefrontal cortex bias processing toward pathways appropriate to current goals \cite{miller2001}, while Stocco et al.\ model the basal ganglia as a conditional router that selects the source and destination of information transfers between cortical regions \cite{stocco2010}.
Zador et al.\ connect these biological ideas to artificial intelligence by highlighting modular hierarchies with partly autonomous lower-level components \cite{zador2023}.
Merel et al.\ explain how this organization can support efficient motor control, with lower-level controllers handling behavior locally and being reused across contexts while higher levels provide guidance \cite{merel2019}.
They also describe \emph{amortized control}, in which a faster reactive controller learns behavior initially generated through costly planning, reducing the need to repeat that planning.

To examine these ideas further in artificial systems, we organize relevant methods into three strands, beginning with approaches that select and compose reusable modules:
PathNet, discussed under \Dref{3}, searches for task-indexed paths through a fixed module pool, whereas Rosenbaum et al.'s learned router selects function blocks recursively for each input \cite{fernando2017,rosenbaum2018}.
Outside reinforcement learning, Neural Module Networks use the linguistic structure of each question to assemble a question-specific computation graph from jointly trained reusable modules \cite{andreas2016}.
Together, these approaches show that computation can be routed through or assembled from reusable modules according to the current problem.
Their module inventories and possible compositions are nevertheless fixed in advance, so they do not establish that a functional hierarchy can be revised or extended over a lifetime.

Other methods in this strand make reuse and structural growth across task sequences explicit:
Veniat et al.\ select earlier modules to reuse and instantiate new ones for a known task, explicitly linking transfer to sublinear growth of model capacity and computation when tasks share structure \cite{veniat2021}.
Mendez et al.\ compose sensor-processing, planning, and motor-control modules in lifelong reinforcement learning, demonstrating reuse, forward transfer, and unseen combinations \cite{mendez2022b}.
CompoNet grows a cascade by adding a policy module at each task change and allowing it to combine the outputs of earlier policies \cite{malagon2024}.
These methods establish that learned modules can be reused and composed, but their structures are created or selected at announced task changes.
They therefore leave task-free operation under \Dref{1} open and do not establish that an agent can vary its functional hierarchy during interaction or revise and extend it autonomously over its lifetime, as required by \Dref{4}.

A second strand concerns conditional and adaptive computation.
Shazeer et al.\ use learned sparse gating to select a small subset from a large pool of experts, decoupling model capacity from the computation used for one input \cite{shazeer2017}.
The expert pool is nevertheless trained in a static supervised protocol, and the number of selected experts is generally fixed rather than adapted to problem complexity.
Graves instead proposes a learned halting mechanism that lets a recurrent network choose how many updates to perform for the current input while penalizing unnecessary steps \cite{graves2016a}.
This mechanism demonstrates input-dependent computation depth within a fixed architecture.
Each internal step reuses the same learned recurrent transition function, so the method varies the number of steps but does not explicitly compose distinct functional levels or provide a mechanism to revise and extend a hierarchy over a lifetime.

A third strand concerns hierarchical predictive coding, which combines bidirectional information flow with local, Hebbian-type learning.
Higher levels send predictions downward, while prediction errors from lower levels propagate upward and drive iterative inference through local updates that can occur asynchronously \cite{mumford1992,rao1999,salvatori2026}.
Classical predictive coding therefore illustrates the reciprocal hierarchical computation relevant to \Dref{4}, but its primarily perceptual formulation does not provide the shared low-abstraction observation and action interfaces proposed above.
Under specific assumptions, predictive-coding dynamics can converge to the same gradients as backpropagation while implementing credit assignment through local updates \cite{millidge2022}.
Burst-dependent plasticity offers a related biologically grounded account in which feedback from higher levels, multiplexed with feedforward activity through bursts, coordinates local learning across a hierarchical circuit \cite{payeur2021}.
Local learning therefore changes how credit assignment is carried out, but does not by itself prevent catastrophic forgetting.
Particular predictive-coding systems that add mechanisms such as lateral competition have nevertheless shown substantially reduced forgetting \cite{ororbia2022}.
Active Predictive Coding, discussed under \Dref{8} for its action-conditioned model, extends this picture by using predictive-coding circuits with local learning in an online robotic-control agent \cite{ororbia2023}.
Like the other implementations in this strand, it uses a fixed computational organization, leaving problem-dependent selection and lifetime revision and extension open under \Dref{4}.

Although the Dreamer architecture is primarily relevant to \Dref{8}, where we discuss it in more detail, it also admits a curious interpretation in the context of \Dref{4}.
Dreamer encodes observations into latent model states and decodes those states back into observation space \cite{hafner2020}.
Through the lens of \Dref{4}, this can be read as an upward path from observations to a latent abstraction and downward paths from that abstraction to reconstructed observations and concrete actions.
The connection remains only an analogy, however, because a latent representation does not, by itself, establish several functional levels or their adaptive selection, growth, and revision.
Moreover, decoder-free extensions of Dreamer support rollouts in latent space without reconstructing observations \cite{okada2021,deng2022}, showing that the apparent bidirectional organization is not essential to the approach.
Still, Dreamer offers an instructive example of how perception, prediction, and action selection can meet at a shared latent level.

Collectively, the methods in all three strands show that modules can be reused and composed, that computation can be routed or iterated according to the current input, and that reciprocal hierarchies can support abstraction and control.
Yet the stronger combination proposed by \Dref{4} remains unestablished: a single continual agent that selects, revises, and extends a functional hierarchy from experience without supplied task structure.

Assessing \Dref{4} is not straightforward because the evidence remains incomplete whether we inspect a system's internal organization or treat it as a black box.
Internal inspection may seem direct, but even seemingly straightforward indicators such as model depth do not by themselves establish hierarchical representations: for example, conventional convolutional neural networks exhibit increasingly complex and task-specific features in deeper layers \cite{zeiler2014,yosinski2014}, yet blocks in convolutional residual networks tolerate removal or reordering, interventions that should be destructive if each block formed an indispensable level of a strict hierarchy \cite{veit2016}.
Vision transformers provide a related cautionary example because tokens derived from local image patches can be repurposed as internal registers instead of continuing to represent those patches \cite{darcet2024}.
Their place in the architecture therefore does not necessarily reveal their functional role.

Black-box proxies are no more decisive.
As under \Dref{3}, computation time could provide one such proxy: simple problems should permit quick responses, while complex problems may require more computation and take longer.
This measure remains difficult to interpret because runtime depends on the implementation, learning may make a formerly complex problem fast, and timing cannot reveal whether the difference arises from a functional hierarchy or whether that hierarchy can be revised and extended.

We therefore view \Dref{4} as a design hypothesis for making \Dref{1}--\Dref{3} jointly viable rather than as a property established by one architectural or behavioral test.
Its value should appear indirectly if an agent continues to learn under \Dref{1}, sustains \Dref{2}, and operates within the resource limits of \Dref{3} while efficiently handling problems of different complexity.

\subsection{Active Sensing}
\label{sec:d5}

\begin{briefbox}[boundarybrief]{D5}
    \emph{\Dref{5} asks a learning system to treat sensing as an active process.
    Instead of merely receiving a fixed observation stream at a predetermined level of detail, an agent should choose what information to seek and how much processing to devote to it as its goals, uncertainty, and experience change.}
\end{briefbox}

\noindent Action, sensing, and perception are tightly coupled because action controls sensing, while perception interprets the acquired information and guides subsequent action and learning \cite{bajcsy1988,bajcsy1992}.
The terminology is not entirely uniform, and the meanings of \emph{active sensing} and \emph{active perception} often overlap.
Here, we use \emph{sensing} to mean acquiring sensory information and \emph{perception} for interpreting it.
Accordingly, we call sensing active when the learning system controls which sensory information it acquires.

In reinforcement learning, exploration can concern not only which action produces reward, but also which action makes useful information available.
Seeking such information can be essential because different world states may produce the same observation even though they require different actions, a problem Whitehead and Ballard call perceptual aliasing \cite{whitehead1990}.
They address this ambiguity with an agent that learns both how to act and where to direct its attention to acquire the necessary sensory information.
This result directly supports our view that active sensing should be learned together with action selection during interaction with the environment.

\Dref{5} arguably sits between our system-level and behavioral categories.
Its placement depends partly on where we draw the agent boundary, since a controllable sensor may be treated as part of the agent or as part of the environment reached through its action and observation interface.
For this paper, we place sensors on the environment side of that interface rather than inside the learner.
Behavioral observations can provide useful but only partial evidence because actions that control the world may also change what the agent can sense.
A movement around an obstacle may serve information acquisition, goal pursuit, or both, and the action stream alone does not distinguish these roles.
Conversely, merely providing a dedicated sensing action does not show that the agent learns to use it according to its current information needs.
We therefore place \Dref{5} among the system-level desiderata as a pragmatic choice, while acknowledging its clear behavioral component.
Our system-level classification does not, however, imply a separate sensing interface, because active sensing can be represented within the existing action and observation spaces.

Active sensing can improve inference by revealing evidence that a fixed viewpoint cannot provide, as when an agent directs a camera or moves around an obstacle \cite{bajcsy2018}.
It can also change the inference problem itself: Aloimonos et al.\ show that controlling viewing geometry makes several low-level vision problems well-posed or more tractable \cite{aloimonos1988}.

Having access to all the necessary sensory information does not mean that an agent should also process all of it at the same level of detail.
Instead, it may treat most sensory input coarsely and reserve detailed sensing and processing for selected parts.
Bajcsy et al.\ illustrate this through the different perceptual needs of navigation and manipulation, with coarse spatial information often sufficient for the former but finer-grained sensory information about particular objects needed for the latter \cite{bajcsy2018}.
The appropriate level of sensory detail therefore depends on what the agent is doing, and this principle extends to all controllable sensory modalities, including sound and touch.
Under the resource limits of \Dref{3}, the agent should learn when additional information is worth its cost.
Bajcsy and Campos frame this choice as an economy of energy, time, and memory, although Tsotsos emphasizes that selecting and combining observations also requires computation even when active sensing reduces sensory processing \cite{tsotsos1992}.
The relevant question is therefore not whether active sensing is always cheaper, but whether its informational and behavioral benefits justify its total cost.
Over a lifetime, this balance also depends on memory, since exploration without remembering earlier experience is possible but uneconomical \cite{bajcsy1992}.
In our framework, this connects the system-level framing of \Dref{5} to the behavioral persistent-memory capability of \Dref{6}: the former asks what sensory information to acquire next, whereas the latter asks how acquired information can be preserved and retrieved when it becomes relevant later.

Another aspect of active sensing concerns what guides attention.
Bajcsy and Campos distinguish task-driven from exception-driven attention, contrasting support for the current activity with sensitivity to events outside the system's expectations \cite{bajcsy1992}.
This distinction adds another dimension alongside the familiar exploration--exploitation trade-off in reinforcement learning.
Exploration versus exploitation concerns whether an action seeks new information or uses current knowledge to pursue reward, whereas task-driven versus exception-driven attention concerns what directs sensing during either activity.
The two choices are distinct, so either form of attention can accompany exploration or exploitation.
For example, task-driven attention may support exploration or a familiar action, while exception-driven attention may prompt information gathering, replanning, or an immediate response based on current knowledge.
In CRL, both choices must remain adaptive over a lifetime: the agent must balance exploration and exploitation while shifting between task-driven and exception-driven attention as its knowledge, objectives, and conditions change.

Over a lifetime, which information is worth acquiring may change as the agent acquires knowledge or encounters new objectives.
A distinction that was previously irrelevant may become essential, so the agent should be able to seek different information or observe it at a finer level of detail.
Active sensing therefore forms a feedback loop with learning: what the agent already knows guides what information it seeks, while new evidence may reveal distinctions it has not yet learned to make.
The specific role of \Dref{5} in this feedback loop is to control which evidence becomes available, while retaining and incorporating that evidence depend on learning and memory mechanisms considered elsewhere.

Biological vision provides a concrete example of why active sensing can be useful, extending the biological inspiration discussed under \Dref{4} \cite{zador2023}.
The primate retina must balance a wide field of view and high acuity against limited neuronal resources \cite{mccann2011}.
The resulting arrangement combines a small retinal region of high visual acuity, the \emph{fovea}, with a lower-resolution periphery, producing the nonuniform visual resolution known as \emph{foveation}.
Rapid eye movements between fixation points, called \emph{saccades}, bring different parts of the scene onto the fovea for detailed inspection.
Peripheral signals nevertheless retain useful information and can help select locations for those movements \cite{mccann2011}.
During visual search, human fixation choices can approach an information-efficient strategy that directs this limited high-resolution region toward locations expected to be most informative \cite{najemnik2005}.
During natural activity, gaze is instead coordinated with the current action, allowing information to be acquired shortly before it is needed instead of maintaining a detailed representation of the entire scene \cite{hayhoe2003}.
Ballard similarly argues that gaze control can turn low-resolution imaging into high effective resolution while simplifying visual computation \cite{ballard1991}.
Together, these works illustrate how the system-level resource limits captured by \Dref{3} can shape the behavioral aspect of \Dref{5} in a natural system: nonuniform retinal resolution concentrates limited neuronal resources in a small region of high visual acuity, while saccades direct that region toward information needed for the current task.
A similar combination of nonuniform sensory resolution and active control may offer artificial systems a benefit beyond efficiency.
Image classifiers are vulnerable to small adversarial perturbations and common corruptions, and small changes to visual observations can also substantially degrade learned reinforcement-learning policies \cite{goodfellow2015,hendrycks2019,huang2017}.
More directly, Shah et al.\ find that a biologically inspired foveated transformation improves robustness to adversarial attacks and common corruptions, while Mukherjee et al.\ report robustness gains for active-vision systems that combine foveated glimpses from multiple fixation points \cite{shah2023,mukherjee2025}.
These results do not show that biological foveation evolved for robustness or establish the same benefit in CRL, but they support robustness as a plausible additional benefit of active sensing.

We include active sensing among our desiderata because we regard control over information acquisition as pivotal for continual learners, yet it remains, in our view, underrepresented in modern CRL.
Neural examples of active sensing nevertheless exist outside CRL, including Mnih et al.'s recurrent attention model, which learns to select a sequence of spatial glimpses \cite{mnih2014}.
Besides reducing processing costs, the learned attention policy searches for a digit placed at an unknown location and, when clutter is present, avoids irrelevant regions by not looking at them.
Once found, the digit is observed in a common local coordinate frame, supporting recognition independently of its position in the full image while retaining spatial information for choosing subsequent fixations.
Its recurrent state integrates information across fixations before producing a decision.
In translated and cluttered digit tasks, this combination outperforms a CNN baseline with a comparable number of parameters while keeping its computation independent of the input-image size.

GFNet provides a related example at ImageNet scale: under a computational budget, it first processes a coarse global view, then uses a policy trained by reinforcement learning to select high-resolution patches, and stops once the classifier is sufficiently confident \cite{wang2020}.
This directly instantiates the control over sensory detail required by \Dref{5} and connects it to the resource efficiency of \Dref{3} and adaptive computation of \Dref{4}, although the system is trained for a fixed classification task.

Malinowski et al.'s bootstrapped hard attention provides a complementary example for visual question answering \cite{malinowski2018}.
It constructs a binary mask over the encoded spatial feature map and passes only the selected feature vectors to subsequent reasoning, with an adaptive variant also determining how many locations remain unmasked.
The authors argue that removing irrelevant features can reduce interference among visual entities and simplify credit assignment, and report improved performance on questions involving relations between objects.
For \Dref{5}, this shows that learning to restrict subsequent processing to task-relevant features can improve relational reasoning while reducing its computational and memory costs.
This allocation remains internal to the model, however, because the mask is computed only after the full input image has been encoded and cannot control which image regions enter the model in the first place.

Returning to reinforcement learning, Shang and Ryoo study active vision in settings where sensory and motor actions can be selected separately and the agent may adjust its view at every step \cite{shang2023}.
They jointly learn a motor policy and a sensory policy that selects image regions or controls a movable camera.
Alongside task reward, an intrinsic signal favors views from which a learned model can infer the motor action between observations, encouraging the agent to observe the effects of its own actions.
The learned sensory policies exhibit fixation and tracking and outperform random viewing and systematic scanning in their limited-observation experiments.
With peripheral vision, they also report higher overall scores across the evaluated Atari games than the full-image baseline, providing further evidence that selective observation can help beyond compensating for restricted input.

To integrate partial observations over time, Shang and Ryoo also introduce a Persistence-of-Vision Memory (PVM) with two variants: one stitches recent partial views into a common spatial representation, while the other combines learned representations of those views recurrently \cite{shang2023}.
PVM preserves information while the sensory policy looks elsewhere, and the authors' ablations show that this contributes to performance.
The resulting system illustrates the boundary between \Dref{5} and \Dref{6}: the sensory policy selects what to observe under \Dref{5}, while PVM combines recent observations into a useful representation of the current situation.
This working combination of learned sensing, acting, and short-term integration is the main contribution to \Dref{5}.
PVM's limited window nevertheless leaves the persistent retrieval required by \Dref{6} unaddressed.

Friston et al.'s active-inference framework complements these learned mechanisms by offering a principled account of what makes an observation worth seeking \cite{friston2017}.
It extends the predictive-coding approach introduced under \Dref{4} to action selection and favors policies both for their expected information gain and for their expected outcomes, allowing an agent to reduce uncertainty while pursuing its objectives.
Da Costa et al.\ review robotics applications in which this coupling supports state estimation, uncertainty reduction, and control \cite{dacosta2022}.

Information acquired to reduce uncertainty may matter even when it yields no immediate task reward.
This connects \Dref{5} to the task-agnostic learning signals of \Dref{7}: exploratory observation policies and curiosity-driven or model-based exploration provide objectives for acquiring information before its usefulness becomes apparent through reward \cite{jayaraman2018,pathak2017,sekar2020}.
If a sensing action receives external reward only much later, choosing that action remains part of \Dref{5}, while attributing the delayed outcome to it invokes the long-range-credit component of \Dref{6}.

In summary, we have argued that active sensing belongs within continual learning and that established methods already provide useful ways to implement it.
The main gap is therefore not a lack of active-sensing mechanisms, but their limited integration into the representative CRL methods considered throughout this paper, which generally take their observation interfaces as given.
\Dref{5} calls for connecting these mechanisms to a continual learner that can keep adapting how it acquires information over its lifetime.

\subsection{Persistent Memory and Long-Range Credit}
\label{sec:d6}

\begin{briefbox}[behaviorbrief]{D6}
    \emph{\Dref{6} asks an agent to retain information throughout its lifetime and retrieve it when it becomes relevant, while assigning credit to past actions when their consequences occur only after long delays and without an immediate reward.}
\end{briefbox}

\noindent In a continuing lifetime, information may become relevant long after it was acquired, and the consequences of an action may emerge only after substantial intervening experience.
Following the memory--storage distinction introduced under \Dref{3}, (1) persistent memory addresses the first gap by preserving and retrieving information for later decisions, while (2) long-range credit addresses the second by attributing delayed consequences to earlier actions so that they can influence subsequent behavior and learning.
\Dref{6} asks for both capabilities in the same agent throughout a continuing interaction.
Under \Dref{1} and \Dref{3}, both capabilities must operate without supplied task structure or consequence-erasing resets and within finite storage and computation budgets.
\Dref{2}, by comparison, asks whether useful abilities survive further learning, whereas \Dref{6} concerns access to particular information across delays and the attribution of delayed consequences to earlier actions.
Satisfying either does not establish the other: an ability may remain after the relevant event has been forgotten, while an event may remain retrievable after the ability needed to use it has degraded.

\noindent\textbf{(1) Persistent Memory.}
Information needed to interpret the current situation may come from recent observations, while information needed for a later decision may have been acquired much earlier.
Persistent memory under \Dref{6} covers the latter case by maintaining access across substantial delays and intervening experience.
Long short-term memory (LSTM) provides a canonical recurrent neural-network mechanism for carrying information across such intervals: gated memory cells can hold information over long periods, and during training they let error signals travel back across long time lags without fading \cite{hochreiter1997}.
LSTMs thus address both parts of \Dref{6} in their simplest form, but with clear limits: everything the agent retains must fit into a fixed-size state that new input can overwrite, and training relies on backpropagation through time, whose cost we return to below.

Learned read--write memories remove the first limit by separating storage from the recurrent state.
Graves et al.'s Neural Turing Machine \cite{graves2014} and Differentiable Neural Computer \cite{graves2016b} couple a neural controller to an external memory matrix and learn how to access it.
At each step, the controller emits read and write heads that address memory locations softly, by content similarity and, in the latter architecture, also by recency and allocation of free slots.
Every operation is differentiable, and the entire system can be trained end to end with gradient descent.
Hence, these networks learn what to store, and when and where to write and read it.
MERLIN extends learned read--write memory to partially observed reinforcement learning by coupling an external memory with an unsupervised predictive objective, so the agent stores information that helps it predict what it will observe next without relying on reward alone to decide what is worth remembering \cite{wayne2018}.
This predictive objective also connects \Dref{6} to the reward-independent acquisition of \Dref{7}.
Relative to our broader CRL target, the evidence across these architectures comes from episodic protocols with bounded horizons, leaving \Dref{1} open, and establishes neither lifetime retention of memory contents nor a release policy under the fixed budget of \Dref{3}.

Storing information is useful only if the agent can retrieve it when it becomes relevant to the current situation.
Attractor-based associative memories provide such state-driven retrieval directly: a partial or noisy input settles into the closest stored pattern such that retrieval is a fixed-point computation rather than a learned addressing policy.
Classical Hopfield networks store patterns as attractors of a recurrent energy-based network \cite{hopfield1982}, Dense Associative Memory raises capacity through higher-order interactions \cite{krotov2016}, and the continuous formulation of Ramsauer et al.\ retrieves patterns in effectively one step with an update rule equivalent to transformer attention \cite{ramsauer2021,vaswani2017}.
This equivalence lets us read a transformer's context window as associative memory, with the contents being discarded once they leave the window.
For \Dref{6}, these models contribute content-addressed retrieval, an important component of persistent memory, but do not establish the broader capability: their high capacity holds only for sufficiently separated patterns, they do not decide which patterns to store or release under the budget of \Dref{3}, and storing patterns sequentially can cause forgetting, as discussed under \Dref{2} \cite{mcalister2025}.
Titans provides a recent step from retrieval toward persistent memory by adding a neural long-term memory that keeps learning at test time \cite{behrouz2025}.
The memory is a small network whose parameters update during inference, with larger updates for surprising inputs and a learned forgetting mechanism that releases outdated content, while attention serves as short-term memory over the current context.
Deciding what to store and what to release thus becomes part of the model, as \Dref{6} requires under the budget of \Dref{3}.
However, the supporting evidence so far comes from finite sequence-modeling tasks such as language modeling and long-context retrieval, not from a continual-learning agent.

Hyperdimensional computing provides a complementary approach to representing and retrieving structured information.
Its algebraic operations allow related information to be combined in one vector and individual associations to be recovered through a suitable query \cite{kanerva2009}.
Teeters et al.\ compare storing associations in a single vector with using a sparse distributed associative memory \cite{teeters2023}.
Their experiments with large finite-state automata show that the associative memory can reduce storage and retrieval costs at matched recall reliability.
For \Dref{6}, these mechanisms support the organization and efficient retrieval of memory contents, but leave open how a continual agent decides what to retain, update, or discard under bounded resources.

The Forager environment probes state construction under partial observability in a single continuing, non-episodic stream \cite{tang2026}.
Two findings from this work bear directly on \Dref{6}.
First, an exponential reward trace, a simple form of state construction introduced by Rafiee et al.\ \cite{rafiee2023}, substantially improves continual control under partial observability in Forager.
Recurrent Trace Units, trained fully online with real-time recurrent learning, yield a similar improvement \cite{elelimy2024}.
By contrast, a deep recurrent Q-network with a gated recurrent unit does not yield the same improvement \cite{tang2026}, so recurrence alone is not evidence of state construction.
Second, methods that mitigate loss of plasticity keep the network trainable but do not recover the missing state information, which separates state construction from \Dref{2}.

Recurrent states, external or associative memories, and learned traces can support persistent memory, but their presence alone does not establish \Dref{6}.
Evidence for persistent memory should show that information acquired at one point in a lifetime remains available and affects behavior much later, without an evaluator marking what to retain and within a declared budget.

\noindent\textbf{(2) Long-Range Credit.}
Pignatelli et al.\ formalize temporal credit assignment as learning how actions influence outcomes \cite{pignatelli2024}.
Its effective horizon may differ from the agent's memory span, as Ni et al.\ show by formalizing separate memory and credit-assignment lengths and demonstrating in controlled reinforcement-learning tasks that transformers improve long-term memory without improving long-range credit assignment \cite{ni2023}.
Long-range credit also differs from long-horizon planning because anticipating distant consequences before acting does not establish that an observed consequence can later be attributed to the earlier actions that caused it.
A classical mechanism for propagating credit backward is the eligibility trace, which maintains a decaying record of recent states or actions and distributes error along it \cite{sutton2018}.
Traces are inexpensive and fully online, and therefore fit \Dref{1} well.
However, their reach is limited: a trace-decay rate determines how quickly the record of past actions fades, and the discount factor sets how much reward feedback is reduced over time, further reducing feedback on earlier actions and decisions.
Actions taken long before an outcome therefore receive little credit.
This use of the discount factor can make eligibility traces appear closely related to the temporal trade-offs discussed under \Dref{9}, where the same factor commonly weights delayed outcomes in the objective.
The two horizons nevertheless run in opposite directions: \Dref{9} concerns how strongly delayed outcomes influence a current choice, whereas the backward horizon under \Dref{6} limits whether credit for a later outcome reaches the earlier actions that caused it.
An agent may therefore value distant consequences and still fail to assign them credit.

Various approaches target longer temporal distances between actions and their consequences.
RUDDER trains a return predictor on completed episodes and redistributes the delayed return along the trajectory, so that the learner receives an immediate signal whenever the predicted return changes \cite{arjona2019}.
Hindsight credit assignment instead estimates how much an outcome depends on an earlier action, rather than distributing the return over all preceding steps \cite{harutyunyan2019}.
Temporal Value Transport uses an attention-based memory to identify earlier events associated with later reward and transports value to them, linking the retrieval and credit components relevant for \Dref{6} within one architecture \cite{hung2019}.
Relative to our CRL target, all three approaches rely on episodic structure in their reported formulations or evaluations, though they do not give the agent oracle access to future reward at decision time.
RUDDER depends on it most directly.
Its signal at each step is the change between two successive return predictions, each computed only from the trajectory observed so far, so the agent learns to anticipate the return rather than being told it.
Episodes are needed to train this predictor to predict the total return from a partial trajectory, with completed trajectories whose final return is already known as training examples.
Hindsight credit assignment and value transport depend on episodes less directly.
Both revisit an earlier action once later events along the same trajectory have been observed, and only the return-conditioned variant of hindsight credit assignment requires the final return.
Their reliance on episodes lies mainly in the evaluated protocols, which use episodic tasks.
Under \Dref{1}, however, no episode end marks when an outcome is complete or how far back its credit should reach, so this interval must itself be inferred.

Long-range credit assignment must also fit within the storage and computation budgets of \Dref{3}.
Backpropagation through time retains an unrolled computational graph whose size grows with the interval and propagates gradients through it, so its resource cost grows with the credit horizon.
Eligibility propagation combines local eligibility traces with online learning signals to approximate recurrent gradients at a fixed per-step cost, consistent with the bounded-computation requirement of \Dref{3} \cite{bellec2020}.
Like classical eligibility traces, however, its effective horizon is limited by the trace decay.
Burst-dependent plasticity, discussed under \Dref{4}, coordinates learning across the levels of a hierarchical circuit and thus addresses credit across levels rather than across time \cite{payeur2021}.
Equilibrium propagation computes gradient-like updates locally in an energy-based network, but it addresses where the learning signal comes from, not how far back in time it can reach \cite{scellier2017}.
Together, these works illustrate possible mechanisms for local credit assignment, but none shows a local rule that bridges longer temporal delays than backpropagation through time.

Overall, memory architectures and credit-assignment methods provide complementary parts of \Dref{6}, but the reviewed evidence does not establish their integration.
A complete demonstration should show one task-free, non-episodic online agent retrieving information across substantial intervening experience and assigning delayed outcomes to earlier actions while operating within bounded resources.
The two effects should be tested separately so that evidence for one is not mistaken for evidence for the other.

\subsection{Reward-Independent Learning and Transfer}
\label{sec:d7}

\begin{briefbox}[behaviorbrief]{D7}
    \emph{\Dref{7} asks an agent to acquire reusable knowledge and skills from experience without relying on the task reward that later makes them useful, and to transfer them efficiently when they become reward-relevant within the same continuing interaction.}
\end{briefbox}

\noindent Task reward indicates which outcomes currently matter, but it is an incomplete learning signal for a continual agent.
In a continuing lifetime, rewards may be sparse or delayed, and future objectives cannot be enumerated in advance \cite{schaul2018,khetarpal2022}.
If learning is driven only by current reward, reusable structure that is irrelevant to the present objective may not be acquired until a later objective makes it useful, by which time the experience needed to learn it may no longer be available.
Environmental dynamics, object effects, and behaviors that reliably change state are often shared across objectives and can be observed before any one of them becomes reward-relevant.
\Dref{7} asks the agent to acquire such structure independently of the task reward that later makes it useful and to transfer it when that relevance arises.

We call a learning signal reward-independent when it does not depend on the task reward of the objective for which the acquired structure later becomes useful.
This includes self-supervised prediction and reconstruction as well as intrinsic objectives such as novelty, prediction error, or learning progress, even when these are implemented as rewards.
\Dref{7} therefore does not require a globally reward-free world: the agent may receive reward for other objectives while acquiring structure whose relevance lies elsewhere, so reward independence must be assessed relative to the structure in question.
Representations shaped jointly by prediction and task reward provide only partial evidence, because the task reward may already have selected what they contain.
\Dref{7} therefore combines two requirements: (1) reward-independent acquisition and (2) transfer once the acquired structure becomes reward-relevant.
Acquisition without later transfer does not show that the structure is useful, while rapid improvement after reward appears may reflect efficient reward-driven learning instead of earlier acquisition.

Animal learning provides a classical precedent.
Tolman reviews latent-learning experiments in which rats that explored a maze without food reward improved abruptly once food was introduced, rapidly approaching the performance of rats rewarded from the start \cite{tolman1948}.
Their knowledge of the maze remained invisible in behavior until reward made it relevant and could be inferred only from the speed of subsequent adaptation relative to a control group, which illustrates both the pattern \Dref{7} targets and why evaluating it is subtle.

A common evaluation design for reward-independent learning divides interaction into a reward-free pretraining phase followed by a reward-labeled adaptation phase \cite{jin2020,sekar2020,laskin2021}.
The transition explicitly tells the learning system when reward-free acquisition ends and reward-directed adaptation begins, supplying the kind of phase boundary withheld by \Dref{1}.
Under \Dref{1}, a new objective or reward may become relevant during interaction, for example when a resource becomes scarce, but this change should occur without an externally supplied task boundary or reset.
Reward-independent acquisition must therefore continue alongside reward pursuit instead of being confined to an earlier phase.
The resulting knowledge and skills must remain usable until they become relevant, connecting \Dref{7} to stability under \Dref{2} and bounded resource use under \Dref{3}.
More generally, \Dref{7} concerns how reusable structure is acquired and later transferred, regardless of whether that structure takes the form of retrievable information, a predictive world model, or a reusable skill.
Establishing that this structure supports persistent memory, predictive reasoning, or planning with reusable skills still requires separate evidence under \Dref{6}, \Dref{8}, and \Dref{10}, respectively.

\noindent\textbf{(1) Reward-independent acquisition.}
Existing work spans three strands: predictive knowledge, intrinsically motivated exploration, and unsupervised skill discovery.
Within the first strand, Horde demonstrates reward-independent predictive learning during continuing interaction on a mobile robot, which learns many general value functions in parallel, off-policy, and in real time without any task reward \cite{sutton2011}.
The questions defining those predictions are supplied by the designer, and the resulting knowledge is not shown to transfer to a later objective.
Veeriah et al.\ address the first limitation by using non-myopic meta-gradients to discover predictive auxiliary tasks whose answers improve the representation learned by an actor--critic agent \cite{veeriah2019}.
However, the discovery criterion is optimized for performance on the current rewarded task, so it remains reward-dependent and does not establish that useful predictive structure can be acquired before its reward relevance is known.
UNREAL trains auxiliary pixel-control and reward-prediction objectives concurrently with task reward on a shared representation, avoiding a separate reward-free pretraining phase \cite{jaderberg2017}.
Its auxiliary objectives are fixed in advance, reward prediction is tied to task reward, and its reported episodic evaluations do not establish the complete \Dref{1} protocol.

World Models trains a variational autoencoder and a recurrent dynamics model on random-policy rollouts before optimizing a compact controller for task reward \cite{ha2018}, providing an early and clean but staged separation of predictive learning from control.
Dreamer instead trains its latent representation jointly through observation reconstruction and reward prediction, while its actor and critic are optimized for task reward \cite{hafner2020}.
Dreamer is therefore relevant to \Dref{8}'s predictive world modeling and reasoning, but provides only partial evidence for \Dref{7} because task reward helps shape the representation whose later reuse would have to be attributed to reward-independent learning; MERLIN's predictive memory, discussed under \Dref{6}, raises the same question \cite{wayne2018}.
DreamZero, discussed in more detail under \Dref{8}, provides related evidence that predictive learning without task reward can support transfer: its video-action model achieves nontrivial zero-shot progress on tasks absent from its robot training data \cite{ye2026}.
Its offline protocol nevertheless does not establish acquisition and use within one continuing lifetime.

The second strand uses intrinsic motivation to guide which experience an agent seeks when task reward provides no such direction.
Oudeyer et al.\ reward learning progress, the decrease of the agent's own prediction error, so that it avoids both well-predicted and unlearnable situations and develops behaviors of increasing complexity \cite{oudeyer2007}.
Pathak et al.'s Intrinsic Curiosity Module rewards prediction error in a feature space learned through inverse dynamics, which discounts aspects of the environment the agent cannot influence; its agents explore without extrinsic reward, and the resulting exploration policy accelerates learning on unseen game levels \cite{pathak2017}.
Burda et al.\ find at scale that purely curiosity-driven agents often make substantial progress on extrinsic game scores \cite{burda2019}.
They also note that dying returns the agent to a familiar, well-predicted starting state, so the episode structure itself discourages dying, and that prediction-error curiosity can be captured by irreducible stochasticity such as a noisy television.
Never Give Up combines an episodic novelty signal, computed from a memory cleared at every episode start, with a life-long novelty signal that decays as the environment becomes familiar \cite{badia2020}.
The former depends on resets withheld by \Dref{1}, and the latter eventually vanishes, so a continual agent needs intrinsic signals that remain informative throughout its lifetime, a requirement that learning progress addresses more directly than raw novelty.

Within this strand, Plan2Explore shifts the emphasis from reward-free exploration itself to transfer.
It reuses a world model learned without task reward to train policies for reward functions revealed afterward \cite{sekar2020}.
It measures novelty as disagreement among an ensemble of learned one-step models, uses rollouts from its learned model to train exploration toward states of high expected disagreement, and later adapts to downstream tasks with little or no further interaction.
However, the downstream reward is supplied only after a separate exploration phase.
A related acquisition-and-transfer example comes from active perception.
Jayaraman and Grauman train a look-around policy to reduce uncertainty about unobserved parts of a scene without access to the downstream tasks and show that it transfers to tasks unknown during training \cite{jayaraman2018}, linking \Dref{7} to \Dref{5}.
Theoretically, Jin et al.\ show that in tabular, episodic, finite-horizon settings, reward-free data collection suffices to compute near-optimal policies for arbitrary reward functions specified afterwards \cite{jin2020}, formalizing the promise of reward-independent exploration while embodying the phase separation that \Dref{1} excludes.
Collectively, these works show that intrinsic objectives can guide exploration without task reward, and several demonstrate benefits on later tasks.
The reported evidence nevertheless either evaluates reward-free exploration directly, tests transfer after a separate training or exploration stage, or relies on episodic mechanisms such as reset memory.
It therefore remains open whether one continual agent can sustain reward-independent acquisition and transfer what it learns when reward relevance emerges without an announced phase change.

The third strand concerns the reward-independent acquisition of reusable behavior through unsupervised skill discovery.
DIAYN maximizes the mutual information between a latent skill and the states it visits and reuses the resulting skills for hierarchical control or fine-tuning \cite{eysenbach2019}.
DADS instead favors skills with predictable consequences, learning a skill-dynamics model that supports zero-shot planning over skills once a reward is specified \cite{sharma2020}; LiSP adopts a related objective inside a learned model, as discussed under \Dref{10} \cite{lu2021}.
METRA learns a compact latent space whose distances reflect temporal reachability and discovers skills that traverse it, extending this line to high-dimensional pixel-based locomotion and manipulation \cite{park2024}.
Because these objectives reward skill distinguishability, predictable consequences, or temporal-metric coverage rather than usefulness for an as-yet unknown objective, skill discovery alone addresses only the acquisition half of \Dref{7}; the other half requires transfer to a reward that was absent during acquisition.
Some DIAYN and DADS navigation experiments further bias acquisition toward designer-selected position coordinates, so those results do not show that the relevant structure would have been identified without advance guidance about what later becomes useful.
These methods also choose the dimensionality or cardinality of the latent skill space in advance, leaving autonomous expansion and revision of a reusable skill repertoire under \Dref{10} untested.

\noindent\textbf{(2) Transfer once reward becomes relevant.}
One route to transfer reuses predictive structure acquired before the task reward was known.
A learned dynamics model predicts the consequences of actions independently of how those consequences will later be valued \cite{sekar2020,jin2020}.
Skill-dynamics models provide analogous predictions for temporally extended behaviors \cite{sharma2020}.
Once a reward becomes relevant, these predictions can be used to compare candidate actions or skills under the new objective.
A second route separates reward-independent from reward-dependent factors of value.
Building on Dayan's successor representation \cite{dayan1993b}, successor features allow the values of previously learned policies to be evaluated immediately for any reward that is linear in a set of features, while generalized policy improvement combines these policies into one at least as good as each \cite{barreto2017}.
Successor features do not require this feature map to be supplied, but their transfer guarantees assume that future rewards can be represented by a shared map.
In Barreto et al.'s experiments, the map is either given or learned from data produced on earlier reward-defined tasks, so the results do not establish reward-independent acquisition of features whose later relevance was initially unknown.

VISR learns controllability-based features and skill-conditioned policies without access to downstream rewards, then infers a new task's reward weights after brief reward exposure and applies generalized policy improvement across the learned policies \cite{hansen2020}.
APS combines variational successor features with nonparametric entropy maximization, jointly learning exploratory coverage and a representation intended for rapid downstream adaptation \cite{liu2021}.
Forward--backward representations go further toward zero-shot transfer, obtaining near-optimal policies for arbitrary rewards from a small number of reward samples within the limits of a finite-rank approximation \cite{touati2021}.
These methods learn reusable structure in a designated reward-free phase and evaluate its transfer after a downstream reward is introduced under fixed dynamics, leaving continual revision as objectives and dynamics change untested.

A third route reuses earlier policies to guide exploration.
Wo{\l}czyk et al.\ isolate the contribution of exploration to forward transfer by replacing SAC's initial random exploration on each new task with policies learned on earlier tasks.
This improves forward transfer on average, although the benefit varies across task pairs and continual-learning methods \cite{wolczyk2022}.
Because those policies were learned from earlier task rewards, the experiment isolates a transfer mechanism without establishing reward-independent acquisition.
Continual-Dreamer moves closer to both parts of \Dref{7} by combining Plan2Explore-style intrinsic motivation with selective replay across a sequence of tasks \cite{kessler2023}.
Its evaluations remain finite and episodic, leaving the non-episodic condition of \Dref{1} untested.
They also use the intrinsic signal alongside task reward, without testing whether earlier reward-independent acquisition supports later transfer.

Across this literature, separate experiments show that predictive knowledge, exploration policies, and skills can be acquired without the downstream task reward, and can reduce or sometimes avoid later reward-driven learning.
The evidence is strongest for staged protocols in which a reward-free phase precedes an announced adaptation phase with a fixed downstream reward.
The reviewed evidence does not establish one continual agent that keeps acquiring reward-independent structure, retains it under bounded resources, and transfers it when reward relevance arises within the continuing interaction without being announced.

As Tolman's latent-learning experiments illustrate, reward-independent structure may remain invisible in behavior until reward makes it relevant, so evidence for \Dref{7} requires comparison with an otherwise matched condition lacking the relevant earlier experience.
It should show that experience gathered while no reward depended on the relevant structure later lets the agent act appropriately sooner than it otherwise could, rather than merely showing rapid improvement once a reward appears.
Such evidence is interpretable only when it is clear which rewards were active during acquisition and which structure the designer supplied, since either may already have shaped what the agent learned.
Under \Dref{1}, reward relevance should arise during the continuing interaction without an announced phase switch.
The acquired structure must also remain usable until then, invoking stability under \Dref{2} or persistent memory under \Dref{6} depending on whether transfer relies on preserved knowledge and skills or retrieval of particular earlier information.
Successful transfer therefore depends on acquiring reusable structure, retaining it until it becomes relevant, and using it to guide behavior.
Transfer performance alone does not identify the mechanism by which that structure was acquired.

\subsection{Predictive World Modeling and Reasoning}
\label{sec:d8}

\begin{briefbox}[behaviorbrief]{D8}
    \emph{\Dref{8} asks an agent to learn and continually revise a predictive world model that relates its current situation and possible actions to their future consequences.
    It should retain and use this model throughout its lifetime to reason about consequences beyond those it has observed directly.}
\end{briefbox}

\noindent An agent that can only react to observed outcomes must learn about a course of action after executing it.
A predictive world model instead allows the agent to reason internally about the plausible consequences of actions before trying them in the environment \cite{lecun2022}.
\Dref{8} concerns learning and continually revising the world knowledge that supports these predictions; \Dref{9} compares the resulting possible outcomes, and \Dref{10} organizes actions and subplans to bring selected outcomes about.
In the standard Markov decision process formulation, a tabular model of the transition and reward dynamics relates the current state and a possible action to a distribution over the next state and reward, providing one realization of this predictive role \cite{sutton2018}.
More generally, a predictive model relates the information currently available to the agent and possible action sequences to distributions over their future consequences; \Dref{8} does not require a neural, generative, or separately named world-model module.
Retaining observed transitions can provide data for such a model, but does not by itself establish that the agent can use learned relations to predict beyond those exact records.

Prediction can remain uncertain because outcomes are stochastic, experience leaves the dynamics incompletely known, or the current observation does not identify the underlying situation.
A Markov state contains the information needed to predict future outcomes, but the agent must still learn how states and actions determine those outcomes.
In an arbitrary tabular Markov decision process, experience with one state--action pair says nothing about an unobserved pair.
Predictions beyond observed interactions therefore depend on assumptions about shared structure and on regularities learned from other experience that connect different states and actions \cite{young2023}.
\Dref{8} therefore requires the agent to exploit such structure when it supports prediction in unfamiliar situations and to represent the uncertainty that remains; correct extrapolation to arbitrary unseen state--action pairs lies outside its scope.
Under partial observability, the current observation may also leave the underlying situation ambiguous, so prediction may depend on a belief or internal state inferred from earlier observations and actions \cite{kaelbling1998}.

Beyond uncertainty, reasoning with a world model raises two action-centered questions: what may follow from an action considered now, and how an observed outcome might have differed had another action been taken in the past.
In the terminology of causal inference, the former is an interventional query and the latter a counterfactual query \cite{pearl2010,pearl2018}.
Interventional queries express the prospective use central to \Dref{8}.
Rezende et al.\ demonstrate this problem for partial models that omit state variables influencing both the actions selected by the behavior policy and later outcomes.
As a result, such a model can fit the observed trajectories yet predict incorrectly when planning evaluates actions under a different policy \cite{rezende2020}.
Counterfactual queries are more demanding than prospective interventional predictions because they ask how a particular observed outcome would have differed under another past action, and the available evidence may support only an uncertain answer.
Accordingly, \Dref{8} requires prospective predictions under candidate actions and treats counterfactual reasoning about past alternatives as an additional capability where the available evidence and causal assumptions support it.

Beyond supporting these queries at a given time, a continual learner must keep its predictive relations both persistent and revisable.
Relations that remain valid should survive further learning, while prediction errors and changes in the environment should permit inaccurate relations to be revised.
Persistent memory under \Dref{6} can retain information needed to infer the current situation, whereas \Dref{8} uses learned relations to predict possible consequences.

Dyna provides an early foundation by integrating online learning of an action-conditioned model with planning from simulated experience \cite{sutton1990}.
In small navigation experiments, Dyna also adapts after changes to the environment, providing early evidence for online model revision in a narrow tabular setting \cite{sutton1990}.
World Models provides an early latent-dynamics approach for high-dimensional observations \cite{ha2018}, and the Dreamer family extends this approach by learning behavior within a latent dynamics model \cite{hafner2020,hafner2021,hafner2025}.
By rolling their learned dynamics forward under candidate actions, these methods provide evidence for the prospective, action-conditioned use of a world model central to \Dref{8}.
Methods in the Dreamer family train an actor and critic on trajectories generated by an action-conditioned latent model, allowing behavior to improve without executing every model-generated trajectory in the environment \cite{hafner2020}.
At deployment, the learned actor selects actions directly without comparing alternative trajectories online.
These methods therefore provide evidence for action-conditioned prediction and policy learning from model-generated trajectories under \Dref{8}; they do not implement decision-time planning under \Dref{10}.
The third generation of this family, DreamerV3, broadens the empirical evidence by applying the same algorithm and hyperparameter configuration in separate training runs for more than 150 tasks with predefined objectives \cite{hafner2025}.
Active Predictive Coding \cite{ororbia2023} offers a complementary, backpropagation-free realization in which a predictive-coding dynamics circuit learns online within episodic tasks to predict the next observation from the current observation and executed action, with prediction error supplying an epistemic exploration signal.
The model is updated after executed transitions instead of being rolled forward to compare candidate actions, and the authors identify continual RL as future work, so the evidence covers online action-conditioned model learning without establishing the prospective reasoning or lifetime revision sought by \Dref{8}.

Predictions relevant to \Dref{8} can be made at different representational levels using different architectures.
DreamZero provides a generative example by adapting a pretrained video diffusion backbone to predict future video together with continuous action sequences \cite{ye2026}.
For cross-embodiment transfer, the authors additionally co-train the model on video demonstrations from humans or another robot without corresponding action labels.
This improves the target robot's performance on tasks absent from its original robot training data, linking the world-model evidence here to reward-independent transfer under \Dref{7}.
The model nevertheless generates a visual future jointly with an associated action sequence under the current instruction, instead of predicting and comparing outcomes conditioned on alternative candidate actions, and its offline training does not establish continual revision.
MuZero and V-JEPA 2 take a different approach by avoiding reconstruction of future observations.
MuZero learns action-conditioned latent dynamics from reward, policy, and value targets and uses them for Monte Carlo tree search \cite{schrittwieser2020}, while V-JEPA 2 predicts future representations in a learned embedding space and uses its action-conditioned model for image-goal robotic planning \cite{assran2025}.
Together, MuZero and V-JEPA 2 show that latent predictions can guide action without reconstructing future sensory input, consistent with LeCun's proposal that world models should predict action-relevant consequences in an abstract representation space instead of functioning as complete world simulators \cite{lecun2022}.
These systems were not proposed as continual learners, and their evaluations do not establish counterfactual reasoning or continual revision as environmental relations change.

Methods developed specifically for continual learning bring persistence closer to the center of the problem.
Continual-Dreamer uses task-agnostic exploration and selective replay to retain and extend learned world models across sequential episodic environments \cite{kessler2023}.
ARROW extends this analysis to DreamerV3, finding that DreamerV3 substantially forgets earlier Atari tasks when compared under the same replay-memory budget, whereas ARROW's distribution-matching replay preserves performance on those tasks more effectively \cite{alyahya2026}.
DRAGO combines synthetic rehearsal with incentives to revisit states relevant to earlier tasks, improving retention and state-space coverage \cite{fu2025}.
Liu et al.\ learn a shallow world model online and use model-predictive control across sequentially presented tasks, directly connecting online model learning to continual planning \cite{liu2025}.
Across these methods, replay, directed revisitation, and online model learning help preserve previously learned predictive relations across sequential tasks.
The reported settings remain finite and episodic, either switching between predefined environments or varying rewards under a shared transition function.
They therefore leave non-episodic operation under \Dref{1} open and do not establish whether a continual agent can selectively revise changed environmental relations while preserving those that remain valid.

LoCA examines a model-level instance of the stability--plasticity tension discussed under \Dref{2}: outdated predictions must be revised without losing relations that remain valid \cite{wan2022}.
Its diagnostic asks whether a locally observed reward change can propagate through the model to states that the agent does not revisit.
Wan et al.\ find that PlaNet and DreamerV2 adapt poorly in this setting because a large replay buffer preserves stale experience, while clearing the buffer causes their neural models to forget predictions for unvisited regions \cite{wan2022}.
A modified linear Dyna method succeeds, whereas its nonlinear neural extension does not.
The LoCA results therefore show that learning a model and using it to generate trajectories or perform search do not by themselves establish the selective revision required by \Dref{8}.
Its episodic resets leave the non-episodic component of \Dref{1} open, and its narrow local reward-change test does not establish the broader \Dref{8} capability.

Structured models provide more direct evidence for causal and compositional reasoning.
Counterfactually-Guided Policy Search uses structural causal models to evaluate alternative actions for individual recorded episodes of a partially observable Markov decision process \cite{buesing2019}.
This provides direct evidence for the stronger counterfactual capability included in \Dref{8}, but only in a fixed offline setting that leaves continual revision of the causal model open.
Schema Networks learn an object-oriented generative model that separates multiple causes and transfers without further learning across altered Breakout environments \cite{kansky2017}.
Their specialized structure and fixed game family support causal composition and transfer, but not revision throughout a continuing lifetime.
Ke et al.\ study causal discovery from visual observations in model-based reinforcement learning and show that explicit structure and modularity can help recover high-level causal variables and relations \cite{ke2021}.
CausalWorld complements such methods with controlled interventions and combinatorial transfer tests for causal structure learning \cite{ahmed2021}.
Together, these works study causal discovery and transfer in controlled finite settings, while persistent autonomous revision of learned relations remains untested.
Theory-based reinforcement learning and EMPA combine theory-guided exploration with symbolic causal-model learning and hierarchical planning \cite{tsividis2026}.
EMPA's learned theories support prospective prediction by simulating alternative action sequences, and its hierarchical search then uses those predictions to plan, linking \Dref{8} to \Dref{10}.
Still, EMPA operates with structured symbolic states, substantial priors and heuristics, and episodic games, and it neither transfers knowledge between games nor learns from one lifelong stream.

Reasoning systems built around large language models form an important boundary case for \Dref{8}.
At inference time, chain-of-thought prompting and related methods can improve performance on multistep problems by allocating additional autoregressive computation to intermediate tokens \cite{wei2022,li2024b}.
Reinforcement-learning post-training can shape which intermediate traces a model generates, while distillation can train a smaller model on traces generated by a stronger one \cite{guo2025}.
Both training routes can be combined with additional inference-time computation, and some reasoning-related behavior can also arise during pretraining (including some ability to detect and correct deliberately introduced errors in reasoning traces \cite{shah2025}).
Performance gains on reasoning tasks do not by themselves establish that the model learns or continually revises an action-grounded world model.
Evidence from internal representations complements these observable reasoning behaviors and connects language models more directly to world modeling.
Li et al.\ identify an encoded Othello board state that causally affects predictions, while Gurnee and Tegmark recover coherent spatial and temporal representations from language models \cite{li2023,gurnee2024}.
In Hao et al.'s framework, the language model proposes alternative reasoning steps and serves as a world model by predicting the reasoning state that would follow each step, allowing Monte Carlo tree search to compare the resulting paths before selecting a solution \cite{hao2023}.
These results demonstrate useful components of reasoning and prediction, but do not establish the causal reasoning discussed above.
CLadder evaluates associational, interventional, and counterfactual questions expressed in language and finds that these formally defined queries remain challenging for language models \cite{jin2023}.
Generated token traces may contain information useful for monitoring without faithfully revealing the computation that produced an answer \cite{korbak2025,turpin2023}.
Longer traces alone are likewise not reliable evidence of better reasoning \cite{ballon2026}.

Reasoning with a frozen language model and transient context is not itself continual learning: it supports inference over currently available information but does not persistently update the complete learning system or continually revise an action-grounded world model.
Even information that remains within a long context may not be used reliably, as performance can depend strongly on where relevant information appears in the input \cite{liu2024}.
Agents built around language models can nevertheless retain information beyond a transient context through external episodic memory, which carries textual reflections across trials without changing model parameters \cite{shinn2023}.
Rehearsal during sequential fine-tuning can likewise preserve earlier language abilities as new tasks are learned, whereas removing it produces catastrophic forgetting \cite{scialom2022}.
Sequential parameter updates still face the stability--plasticity problem, as illustrated by Huang et al.'s observation of substantial forgetting during sequential adaptation and its mitigation through self-synthesized rehearsal \cite{huang2024}.
Language models may therefore be components of a continual learner, while the reasoning results above do not by themselves establish CRL.

The methods reviewed in this section establish action-conditioned prediction in latent and structured spaces, model-generated experience, causal discovery, compositional transfer, and partial mechanisms for persistence and revision.
Evidence is strongest for prospective prediction under candidate actions, while counterfactual reasoning is usually studied separately in controlled or textual settings.
The work reviewed here does not yet establish their integration in one task-free, non-episodic online agent that continually revises a reusable action-grounded world model while retaining relations that remain valid.

Assessment should begin from predictive behavior because an explicit world-model module is neither necessary nor sufficient evidence for \Dref{8}.
Controlled behavioral tests can provide evidence that an agent uses predictions about the consequences of candidate actions, but they cannot determine how those predictions are represented or computed internally; claims about internal structure require inspection, ablation, or intervention on the relevant components.
Tests should distinguish prediction on observed trajectories, prediction after interventions or policy changes, and, where supported by the available evidence and assumptions, counterfactual questions that hold an observed situation fixed while changing an earlier action.
They should also examine composition of learned relations, selective revision after localized changes, retention of unchanged relations, and whether the resulting predictions improve behavior.
The roles of \Dref{6} and \Dref{8}--\Dref{10} are conceptually distinct, but their behavioral assessment can be difficult to disentangle.
An incorrect action may reflect a failure to retrieve relevant information under \Dref{6}, to predict possible consequences under \Dref{8}, to compare those consequences under \Dref{9}, or to construct and coordinate a suitable plan under \Dref{10}.
Direct prediction, intervention, and counterfactual probes can provide more specific evidence for \Dref{8}, while end-to-end control performance should be treated as joint evidence unless the neighboring contributions are controlled or ablated.
Changes should not be announced through task identities or boundary signals under \Dref{1}, and the resources used to maintain the model and generate action-conditioned predictions should be reported under \Dref{3}.

\subsection{Outcome Trade-Offs}
\label{sec:d9}

\begin{briefbox}[behaviorbrief]{D9}
    \emph{\Dref{9} asks an agent to value and compare possible outcomes across time, objectives, and uncertainty.
    In making these trade-offs, it should determine how delayed consequences, competing objectives, and rare but severe risks affect its choices, and revise how it resolves them as conditions and priorities change.}
\end{briefbox}

\noindent When considering a course of action, an agent may use its knowledge of the world to project several ways in which the interaction could unfold.
\Dref{8} supplies predictions about what may happen under candidate actions and represents uncertainty where those predictions are not determined.
\Dref{9} addresses the separate problem of comparing the projected possibilities across time, objectives, and risk.
This decision problem gives rise to three related conflicts:
\textbf{(1)} An action may provide an immediate benefit while causing an adverse effect much later.
\textbf{(2)} The same action may advance one objective while undermining another.
\textbf{(3)} Favorable outcomes may be common while an unlikely outcome is severe enough to change the preferred choice.
To describe how an agent resolves these conflicts, we use the term \emph{utility} for a rule that ranks possible policy outcomes by specifying how consequences across time, objectives, and uncertainty are weighed.
Under this broad usage, discounting and average-reward criteria, objective scalarization, and expectation or risk measures are different ways of defining utility.
Vamplew et al.\ connect these choices within one utility-based account and motivate learning policies across their parameterizations \cite{vamplew2024}.
For CRL, \Dref{9} additionally requires one agent to maintain and revise the resulting trade-offs throughout a continuing interaction as its conditions and priorities change.

\noindent\textbf{(1) Long horizons.}
The first conflict concerns how immediate and delayed consequences are combined when evaluating a course of action.
Established criteria define the temporal component of utility differently: finite-horizon criteria impose a cutoff, geometric discounting progressively reduces the weight of distant consequences, and average reward assigns equal weight across time \cite{puterman1994,sutton2018}.
Under geometric discounting, a consequence occurring $d$ steps ahead receives weight $\gamma^d$, thereby encoding a substantive temporal preference.
This becomes consequential when a decision is irreversible: a plan providing an immediate benefit $B$ but making a penalty $C$ unavoidable after $d$ steps is preferred to a safe alternative with zero return whenever $B > \gamma^d\,C$.
Any fixed $\gamma<1$ can thus make a sufficiently delayed finite penalty negligible when the commitment is made, even though its weight increases as it approaches and changing course may no longer be possible.
If the delay is instead a random variable $D$, its expected discounted contribution is $\mathbb{E}[\gamma^D]\,C$, so discounting represents timing uncertainty but may still value the same certain consequence differently according to when it occurs.
Fixed geometric discounting is therefore not a neutral model of general sequential preferences \cite{pitis2019}, and the penalty required to enforce safety can depend strongly on the time to failure \cite{massiani2023}.
When an outcome should be avoided independently of its timing, an explicit safety or risk constraint is more appropriate than an arbitrarily chosen finite penalty.

For \Dref{9}, $\gamma$ is part of the utility used to compare outcomes because it determines how strongly delayed consequences influence a choice.
Koopmans provides one foundation for this weighting through stationary preferences that favor earlier outcomes \cite{koopmans1960}.
In reinforcement learning, geometric discounting also admits an episodic interpretation: $\gamma$ is the probability that an episode continues for another step, while $1-\gamma$ is the probability that it terminates and no later rewards are counted \cite{sutton2018}.
This interpretation can be useful when such termination is part of the problem, but it is not implied by the continuing, non-episodic interaction required by \Dref{1}.
When no such termination model is intended, average reward provides an alternative that gives immediate and delayed rewards equal weight \cite{wan2021}.
This removes attenuation by delay, but an isolated finite harm can vanish in the long-run average, so policies with the same reward rate may still differ in important transient consequences \cite{mahadevan1996}.
It also does not by itself resolve risk or competing objectives.
Choosing $\gamma$ sufficiently close to one can recover average-reward-optimal behavior in some fixed environments, but the required value still depends on their dynamics and consequence delays and can make learning more difficult \cite{naik2019,dewanto2022}.
Taken together, these observations show that no temporal criterion is neutral: finite horizons, geometric discounting, and average reward each reduce the influence of some consequences.
\Dref{9} therefore does not prescribe one of them, but asks whether the agent can make and revise the intended temporal trade-offs without relevant delayed outcomes disappearing through the chosen criterion.

\noindent\textbf{(2) Multiple objectives.}
The second conflict arises when possible outcomes cannot be ranked along a single fixed objective.
Multi-objective reinforcement learning (MORL) evaluates outcomes separately for each objective and uses preferences to compare the resulting trade-offs \cite{roijers2013,hayes2022}.
Pareto optimality can identify policies for which no objective can be improved without worsening another.
Choosing among the resulting Pareto-optimal policies requires preferences, which a utility function or scalarization can encode.
A common choice is a linear scalarization that weights and sums the objective returns, whereas nonlinear utilities can express thresholds, interactions between objectives, or other trade-offs that no weighted sum captures.
Nonlinearity also matters under uncertainty because a policy preferred when utility is applied to its average outcome may not be preferred when utility is first assigned to each possible outcome and then averaged.
\Dref{9} therefore cannot always separate trade-offs between objectives from risk under uncertainty.
The appropriate learning problem depends on when preferences become available and whether they may later change: known fixed preferences can support learning one policy, whereas unknown or variable preferences may call for a coverage set or a policy conditioned on the current preferences \cite{hayes2022}.

Existing MORL methods cover several parts of this space.
Pareto Q-learning constructs a set of Pareto-dominating policies in one run under an episodic formulation \cite{vanmoffaert2014}.
Preference-conditioned methods instead learn a policy or value function that adapts to supplied linear objective weights \cite{abels2019,yang2019}, while other methods address changing or uncertain preferences over a fixed collection of objectives \cite{natarajan2005,buetgolfouse2023}.
Deschamps et al.\ provide a continual-learning example in which utilities change over a fixed set of reward components, adapting a fixed-capacity policy to preferences that may be nonlinear \cite{deschamps2026}.
Their method discretizes the current utility over the objective space and converts it into soft masks over shared policy parameters, allowing related utilities to reuse overlapping parts of the network.
This demonstrates continual adaptation to changing nonlinear preferences without model growth, yet the current utility remains externally supplied.
Together, these methods avoid committing the learner to one fixed linear scalarization, either by preserving several possible trade-offs or by conditioning behavior on preferences that may be supplied, changed, or uncertain.
They show that such a scalarization can be restrictive because it collapses objective-specific returns into one score and fixes one weighting in advance, so changing priorities may require another policy to be learned.

Beyond changing preferences over a fixed set of objectives, a continual learner may encounter new objectives or lose access to reward signals for earlier ones.
CORe3 directly studies this form of continual MORL by adding objective-specific heads as new objectives appear and learning a reward model for all objectives encountered so far.
For each current transition, CORe3 combines observed signals for the current objectives with reward-model predictions for earlier ones, allowing the policy to continue training toward all of them \cite{li2024a}.
The experiments show that the agent can incorporate new objectives while retaining performance on earlier ones even after their reward signals become unavailable.
Within \Dref{9}, this establishes adaptation to an expanding objective set with externally supplied preferences and linear scalarization, while leaving nonlinear trade-offs and preference inference from experience untested.
Learning remains online within each stage, while supplied task-change signals and episodic benchmark interactions leave the task-free and non-episodic parts of \Dref{1} open.
Its expanding architecture also leaves the bounded-resource requirement of \Dref{3} open, and the experiments do not address the long-horizon or risk components of \Dref{9}.

Independently of whether objectives or preferences change, a more basic question is whether a single scalar reward can encode every relevant trade-off.
Skalse and Abate answer this question largely in the negative: their necessary and sufficient conditions show that most multi-objective and risk-sensitive tasks cannot be represented by scalar Markovian rewards \cite{skalse2023}.
This does not make scalar rewards universally inadequate, but it cautions against treating one scalar signal as evidence that all relevant trade-offs have been represented.

\noindent\textbf{(3) Risk under uncertainty.}
The third conflict arises because expected or average return can hide a low-probability severe outcome behind frequent small gains.
Tamar et al.\ optimize a risk measure called conditional value at risk, which evaluates the expected return among a specified fraction of the worst outcomes \cite{tamar2015}.
Chow et al.\ formulate related constrained problems that optimize expected performance while limiting either the probability of cumulative cost exceeding a harmful threshold or the expected cost among the worst outcomes \cite{chow2018}.
Hayes et al.\ connect risk and multiple objectives more directly through planning algorithms that optimize nonlinear utility over distributions of possible returns, demonstrating that these two parts of \Dref{9} can be addressed jointly \cite{hayes2023}.
Together, these methods provide practical ways to account for risk once the relevant criterion and planning problem have been specified, but do not show how the resulting trade-offs can be revised during continual learning.
Estimating a hazard's probability and consequences stays within \Dref{8}, while under \Dref{9} the agent uses this descriptive world knowledge to determine and revise how strongly the hazard should influence action.

Related formulations illustrate why these evaluations may depend on more than an externally fixed reward function.
In homeostatic reinforcement learning, the value of an outcome depends on the agent's internal physiological state, allowing behavior to anticipate future imbalance \cite{keramati2014}.
This is not a CRL method, but it motivates continuing objectives whose importance changes with the agent's state.
Active inference, as discussed under \Dref{5} for its treatment of information-seeking actions, is also relevant here because it can evaluate policies according to both how well their predicted outcomes match preferences and how much uncertainty they are expected to reduce \cite{friston2017}.
Both perspectives broaden how outcomes may be valued without demonstrating the complete long-horizon, risk-aware, multi-objective requirement in one continual learner.

Existing reinforcement-learning methods address each component of \Dref{9}, and some connect multiple components.
Across the work reviewed here, the unresolved challenge is their integration: none evaluates one task-free, non-episodic online agent that maintains and revises temporal, objective, and risk trade-offs throughout a continuing lifetime.
Evidence for \Dref{9} should therefore examine all three considerations within the same continuing deployment, including whether the agent can revise their trade-offs as conditions and priorities change.
Aggregate return alone is insufficient because it can hide how delayed consequences, competing objectives, and low-probability severe outcomes shaped the agent's decisions.

\subsection{Concurrent Hierarchical Planning}
\label{sec:d10}

\begin{briefbox}[behaviorbrief]{D10}
    \emph{\Dref{10} asks an agent to build hierarchical plans from reusable subplans operating at different levels of abstraction and temporal scales, and to coordinate such subplans when their execution overlaps.
    We call this coordination concurrent whether the subplans are executed in parallel, interleaved through a sequential action stream, or distributed among cooperating agents.}
\end{briefbox}

\noindent Organizing behavior hierarchically can make long plans more tractable by packaging lower-level behavior into reusable subplans.
A high-level plan can therefore commit to a subgoal while postponing its lower-level realization, keeping distant parts of the plan abstract while uncertainty is high and refining them as their execution approaches and new information becomes available.
Doing so requires the agent to monitor feasibility and replan when necessary.
Decomposing a complex plan into reusable subplans does not necessarily make those subplans independent.
Their executions may share action patterns or depend on the same preparatory conditions, so coordinating them can avoid repeated work and allow one action or intermediate result to advance several subplans.
Interruptibility and interleaving are not merely ways to improve efficiency.
When temporal constraints or dependencies cross subplan boundaries, there may be no feasible schedule that executes each subplan to completion before another begins; the agent must instead interrupt one, advance another, and later resume the first.
Several concurrent subplans may jointly serve one objective, while a single course of action may balance several objectives.
A single agent may execute such subplans in parallel or interleave their actions, while cooperating agents may distribute them among themselves.
When a subplan initiates a durative process whose result becomes available only later, an agent with no other active plan may have nothing useful to do before that process completes.
Maintaining several ongoing plans lets the agent use this interval to advance compatible work.
The agent must therefore decide when new work can begin while another process remains underway, accounting for remaining durations, shared resources, possible interference, and conditions that must continue to hold.
Existing planning methods provide partial solutions to the coordination problems described above.
Classical plan-merging algorithms coordinate separately generated plans by sharing actions or resources \cite{yang1992}, while probabilistic temporal-planning methods coordinate simultaneous and durative actions under stochastic durations \cite{mausam2008}.
These approaches capture complementary aspects of the coordination required by \Dref{10}, but leave the learning and revision of reusable subplans outside their scope: plan-merging algorithms receive completed plans as input, while probabilistic temporal planners assume specified goals and action models.
A continual agent would need to learn and revise a repertoire of reusable subplans over its lifetime while coordinating their shared and temporally overlapping execution.

In reinforcement learning, \emph{options} provide one established formalization of reusable, temporally extended subplans.
Formally, an option is defined by the states in which it may begin, a policy followed while it is active, and a rule for its termination \cite{sutton1999}.
It packages learned behavior into a reusable choice that a higher-level controller can select and sequence, and options may themselves invoke other options to form a hierarchy \cite{barto2003}.
Yet options are only one way to realize reusable subplans and do not by themselves satisfy \Dref{10}.
Goal-conditioned policies, learned skills, and other subtasks can play the same role.
Conversely, a collection of options does not by itself show that an agent can organize them hierarchically, plan with them, or maintain, interrupt, and resume several plans concurrently.
Still, their explicit initiation, policy, and termination structure makes options a useful foundation for learning and planning with reusable behavior.

The options relevant to \Dref{10} should themselves emerge through learning, together with the abstractions that organize them into a planning hierarchy.
Drescher's constructivist account provides an early conceptual motivation for this idea by separating predictions of what would follow from possible actions from the selection of predicted outcomes according to current value \cite{drescher1991}.
It treats intermediate concepts and recursively composed actions as products of learning instead of assuming them in advance.

Early hierarchical reinforcement-learning methods formalize multilevel control and reusable lower-level behavior, providing part of what \Dref{10} requires, but they generally assume much of the hierarchy in advance rather than learning its abstractions and organization.
An early example is Feudal reinforcement learning, which lets higher-level managers issue commands to subordinate learners operating at finer spatial and temporal resolutions \cite{dayan1993a}.
By hiding the external reward from lower levels, it allows them to learn reusable command-conditioned behavior independently of whether a manager's current command serves the overall objective, but this learning occurs within a state hierarchy and information structure specified in advance.
All levels also implement one nested command chain instead of maintaining several ongoing plans.

MAXQ provides another early formalization of hierarchical control by decomposing a Markov decision process and its value function into a programmer-specified hierarchy of subtasks \cite{dietterich2000}.
The value of choosing a child subtask separates into the return produced while executing that child and the return from completing its parent afterward.
This decomposition allows several parents to reuse the same child, while state abstraction lets a subtask ignore variables that do not affect its local problem.
MAXQ makes subtask policies compact and reusable by allowing them to ignore the context in which a parent invokes them.
This creates a composition trade-off because, under recursive optimality, each subtask is optimized given its descendants, but their combination need not produce the best policy expressible by the hierarchy as a whole.
When a subtask's best behavior depends on why its parent invoked it, MAXQ can mitigate this restriction by reconsidering the primitive action at every step, but it still selects one current path through a fixed task hierarchy.
Its hierarchy, termination conditions, pseudo-rewards, and relevant state abstractions remain externally specified instead of being discovered and revised over a lifetime.

Barto and Mahadevan show how options, MAXQ, and related early methods can be understood using semi-Markov decision processes and identify automatic hierarchy construction, activity-dependent abstraction, transfer, and evaluation in larger applications as open directions \cite{barto2003}.
Later, Pateria et al.\ provide a taxonomy that distinguishes learning a policy within handcrafted subtasks from unified policy and subtask learning, task-independent subtask discovery, and transfer across tasks \cite{pateria2021}.
Their assessment is particularly relevant here because skill discovery, transfer, composition, reuse, and selective initiation are largely addressed by separate methods.
Integrating these operations into a lifelong skill repertoire with high-level planning remains open.

Later hierarchical reinforcement-learning methods learn some of the temporal abstractions and goals that Feudal reinforcement learning and MAXQ assume in advance.
Option-Critic jointly learns option policies, their termination conditions, and the policy that selects among them without requiring supplied subgoals or auxiliary rewards \cite{bacon2017}.
FeUdal Networks learn manager and worker policies operating at different temporal scales, while HIRO lets a high-level policy propose goals to a low-level policy and corrects replay as ongoing training changes the lower-level behavior \cite{vezhnevets2017,nachum2018}.
Building on the DreamerV2 architecture discussed under \Dref{8}, Hafner et al.'s Director method trains a high-level policy to select latent goals and a low-level policy to reach them from visual observations \cite{hafner2021,hafner2022}.
These methods demonstrate that temporal abstractions and their policies can be learned, but their finite and mainly episodic evaluations use a fixed number of levels, and the agents follow one active chain of subgoals at a time without maintaining several interruptible and resumable plans.

Vezhnevets et al.\ extend learned hierarchical control to multi-agent games by separating the selection of a strategic response from its low-level implementation \cite{vezhnevets2020}.
A high-level policy uses the agent's interaction history to select among learned response options, allowing familiar behaviors to be recombined against unfamiliar opponents.
This improves generalization to held-out opponents and shows how multi-agent interaction can ground a reusable behavioral hierarchy.
However, its two-level structure and number of options are fixed, concealed opponent information supplies a training signal, and the evaluation is episodic.
The results therefore demonstrate generalization through learned response options, but not the autonomous construction and revision of a planning hierarchy throughout a continuing lifetime.

Learned options and other reusable subplans should form reliable new combinations even as continued learning changes the individual components.
The Option Keyboard realizes a restricted form of such composition by synthesizing options for objectives formed from linear combinations of the pseudo-rewards associated with previously learned options \cite{barreto2019}.
Tasse et al.\ instead use Boolean operations over learned task value functions to obtain zero-shot policies for conjunctions, disjunctions, and negations of previously learned tasks \cite{tasse2022}.
Under \Dref{4}, we discussed Mendez et al.'s modular method and CompoNet as examples of learned modules being reused and recombined across lifelong or continual reinforcement-learning task sequences \cite{mendez2022b,malagon2024}.
For \Dref{10}, these methods provide direct evidence of compositional reuse.
Across their evaluations, however, composition is constrained by supplied or restricted task structures, composition rules, or fixed module architectures, and each agent addresses only a single task at a time.
Bagaria et al.\ expose a further difficulty: options learned independently can cease to compose reliably when their policies or regions of applicability change, and additional mechanisms are required to preserve reliable composition \cite{bagaria2021}.

LiSP combines reusable skill acquisition with a learned model and decision-time planning over skill sequences, providing a useful contrast with MuZero's latent-model search over primitive actions discussed under \Dref{8} \cite{schrittwieser2020,lu2021}.
Real transitions train an ensemble dynamics model, model-generated rollouts support the acquisition of reusable task-independent skills, and model-predictive control is implemented via Model Predictive Path Integral optimization \cite{williams2017} to evaluate skill sequences and replan after every action.
LiSP therefore connects reward-independent skill acquisition under \Dref{7}, model-based lookahead under \Dref{8}, and skill-space planning under \Dref{10} within one system.
Its evidence nevertheless concerns sequencing skills rather than coordinating overlapping subplans, and it does not test the continued revision and reliable composition of an expanding skill repertoire.
Another example of skill-level model-based planning is provided by Liang et al., who learn effect models for parameterized skills and use them in high-level search, yet assume supplied skills and task specifications and do not study the coordination of overlapping subplans in continuing interaction \cite{liang2022}.

Concurrent planning under \Dref{10} concerns the temporal coordination of several ongoing subplans, whether one agent runs several processes or cooperating agents carry out different parts of a shared plan.
It does not require literal parallel execution: when only one action can be emitted at a time, actions belonging to different ongoing subplans may be intermingled while the state and progress of each are retained.
Rohanimanesh and Mahadevan explicitly formulate their concurrent-action model to cover both a single agent executing several processes and multiple agents acting in parallel \cite{rohanimanesh2002}.
Their multi-actions combine primitive or temporally extended actions, while the \emph{all}, \emph{any}, and \emph{continue} termination schemes determine when a new decision is made and what happens to unfinished components.
The \emph{continue} scheme is especially relevant to \Dref{10}, since completed components can be replaced while unfinished ones remain active.
Barto and Mahadevan present a related construction in terms of multi-options \cite{barto2003}.
These constructions assume a supplied action repertoire and a fixed episodic task, and do not study the discovery, interruption and later resumption, or lifelong revision of concurrent subplans.
Saxe et al.\ introduce a module based on multitask linearly solvable Markov decision processes that combines several learned macro-actions concurrently and can itself be stacked to form hierarchies across space and time \cite{saxe2017}.
This realizes concurrency as a simultaneous weighted composition of behaviors, but does not maintain separate plans that can be interrupted, interleaved, and resumed during one continuing interaction.
The evaluated hierarchy and task basis are also fixed, leaving autonomous lifelong discovery and revision open.

Several capabilities needed for \Dref{10} are addressed by other desiderata, but none of them establishes concurrent hierarchical planning on its own.
The system-level \Dref{4} describes the machinery for organizing and allocating computation, while the behavioral \Dref{10} asks whether the agent can use that machinery to build and coordinate plans from reusable components.
A functional hierarchy can therefore support \Dref{10} without demonstrating concurrent hierarchical planning on its own.
Reusing subtasks and ignoring locally irrelevant state variables can reduce storage and computation under \Dref{3}, while revising a reusable hierarchy without destroying useful components requires the stability and plasticity of \Dref{2}.
Viewed through the needs of \Dref{10}, \Dref{7} concerns acquiring reusable options or option-like skills, \Dref{8} predicting their consequences, and \Dref{9} determining which objectives and risks should guide their use.
\Dref{10} then concerns how these skills and subplans are sequenced, composed, interrupted, and coordinated, whether they support one objective or several.

A behavioral evaluation of \Dref{10} should require reuse in unfamiliar combinations, planning across levels of abstraction or temporal scales, and coordination when temporally extended subplans overlap.
Comparisons with strictly sequential execution and independently planned subplans would show whether coordination reduces unproductive waiting or duplicated preparation.
Interrupting one plan with another and later requiring its efficient resumption would test concurrent maintenance more directly than a single long-horizon return.
The evaluation should state whether concurrency is realized through parallel execution, interleaving, or distribution across cooperating agents.
In the distributed case, simultaneous team activity alone is insufficient; changing teammates, roles, or plan dependencies should test whether the focal agent can reorganize its own contribution to the shared plan.
Evaluation should distinguish composition from training a separate policy for every combination and state which subgoals, task identities, skill boundaries, and composition rules are supplied.
An explicitly hierarchical architecture can support an explanation of the result but is neither necessary nor sufficient evidence by itself.
Existing work demonstrates substantial parts of the desideratum, but we are not aware of one agent that autonomously discovers, composes, revises, and concurrently coordinates a reusable planning repertoire under the complete \Dref{1} protocol.

\section{Evaluation Landscape}
\label{sec:benchmark-coverage}

\definecolor{probed}{HTML}{1C5670}
\definecolor{notprobed}{HTML}{A0A6AC}
\definecolor{ours}{HTML}{FFF0CD}
\definecolor{block}{HTML}{D6DEE6}

\DeclareRobustCommand{\CoverageMark}[2]{%
    \ensuremath{\vcenter{\hbox{%
        \tikz[x=0.55ex, y=0.55ex, line width=0.08ex]{
            \fill[#1] (0,0) -- (0,1) arc[start angle=90, end angle={90+360*#2}, radius=1] -- cycle;
            \draw[#1] (0,0) circle[radius=1];
        }%
    }}}%
}
\DeclareRobustCommand{\Full}{\CoverageMark{probed}{1}}
\DeclareRobustCommand{\Half}{\CoverageMark{probed}{0.5}}
\DeclareRobustCommand{\None}{\CoverageMark{notprobed}{0}}

\definecolor{notesbg}{HTML}{F3F5F7}
\begin{table*}[t]
    \centering
    \caption{Coverage of the desiderata by representative benchmarks and by our proposed benchmark.
        Ratings refer to the protocols and readouts described in the cited studies and in our proposal.
        The last row gives the intended coverage of the benchmark proposed for our playable companion.
        For \Dref{1}, the columns describe whether the task-free, non-episodic, and online conditions are instantiated: \Full~fully; \Half~partially; \None~not instantiated.
        For the behavioral columns, \Full~indicates that a dedicated protocol or readout directly probes the property; \Half~indicates incomplete component coverage or indirect or confounded evidence; \None~indicates that the property is not probed.
        The system-level desiderata \Dref{3}--\Dref{5} are omitted.}
    \label{tab:benchmark-coverage}
    \small
    \begin{tabularx}{\textwidth}{>{\raggedright\arraybackslash}X ccc cccccc}
        \toprule
        & \multicolumn{3}{c}{Protocol (\Dref{1})} & \multicolumn{6}{c}{Behavioral} \\
        \cmidrule(lr){2-4} \cmidrule(l){5-10}
        Benchmark & \shortstack{Task-\\free} & \shortstack{Non-\\episodic} & \shortstack{On-\\line} & \Dref{2} & \Dref{6} & \Dref{7} & \Dref{8} & \Dref{9} & \Dref{10} \\
        \midrule
        \addlinespace
        \rowcolor{block}\multicolumn{10}{l}{\textbf{Predefined task sequences} (Sec.~\ref{sec:landscape-sequences})} \\
        CLEAR (imagery)~\cite{lin2021} & \None & \None & \Full & \Full & \None & \None & \None & \None & \None \\
        Continual World~\cite{wolczyk2021} & \None & \None & \Full & \Full & \None & \Half & \None & \None & \None \\
        CORA~\cite{powers2022} & \Full & \None & \Full & \Full & \None & \Half & \None & \None & \None \\
        COOM~\cite{tomilin2023} & \None & \None & \Full & \Full & \None & \Half & \None & \None & \None \\
        Continual multi-objective suite~\cite{li2024a} & \None & \None & \Full & \Full & \None & \Half & \None & \Half & \None \\
        \addlinespace
        \rowcolor{block}\multicolumn{10}{l}{\textbf{Reset-free and continuing testbeds} (Sec.~\ref{sec:landscape-continuing})} \\
        EARL~\cite{sharma2022} & \Full & \Half & \Full & \None & \None & \None & \None & \None & \None \\
        Lifelong Hopper/Ant (online runs)~\cite{lu2021} & \Full & \Full & \Full & \Half & \None & \Half & \Half & \None & \Half \\
        JBW (PT-DQN experiment)~\cite{anand2023} & \Full & \Full & \Full & \Half & \None & \None & \None & \None & \None \\
        Forager~\cite{tang2026} & \Full & \Full & \Full & \Half & \Half & \None & \None & \None & \None \\
        AgarCL~\cite{mohamed2025} & \Full & \Full & \Full & \Half & \Half & \None & \None & \None & \None \\
        \addlinespace
        \rowcolor{block}\multicolumn{10}{l}{\textbf{Focused diagnostics} (Sec.~\ref{sec:landscape-complementary})} \\
        URLB~\cite{laskin2021} & \None & \None & \Full & \None & \None & \Full & \None & \None & \None \\
        CausalWorld (zero-shot tests)~\cite{ahmed2021} & \Full & \None & \None & \None & \None & \Half & \None & \None & \None \\
        CompoSuite (zero-shot tests)~\cite{mendez2022a} & \None & \None & \None & \None & \None & \Half & \None & \None & \Half \\
        LoCA~\cite{wan2022} & \Full & \None & \Full & \Half & \None & \None & \Half & \None & \None \\
        T-Maze diagnostics~\cite{ni2023} & \Full & \None & \Full & \None & \Full & \None & \None & \None & \None \\
        \addlinespace
        \rowcolor{block}\multicolumn{10}{l}{\textbf{Related research areas} (Sec.~\ref{sec:landscape-complementary})} \\
        ARC-AGI-3~\cite{arcprize2026} & \Half & \None & \Full & \None & \None & \None & \Half & \None & \Half \\
        Melting Pot (social generalization tests)~\cite{leibo2021} & \Full & \None & \Full & \None & \None & \None & \None & \Half & \Half \\
        MineRL~\cite{guss2019} & \Full & \None & \Full & \None & \None & \None & \None & \None & \None \\
        MineDojo~\cite{fan2022} & \Full & \None & \Full & \None & \None & \Half & \None & \None & \None \\
        \midrule
        \rowcolor{ours}
        \textbf{Our companion (Sec.~\ref{sec:benchmark-proposal})} & \Full & \Full & \Full & \Full & \Half & \Half & \Half & \Half & \Half \\
        \bottomrule
    \end{tabularx}
    \par\bigskip
    \setlength{\fboxsep}{5pt}
    \noindent\colorbox{notesbg}{%
        \begin{minipage}{\dimexpr\linewidth-2\fboxsep\relax}
            \footnotesize
            The following notes clarify \Dref{1} ratings that require additional context about the evaluated protocols:
            \par\smallskip
            \begin{tabular}{@{}ll@{}}
                \emph{CLEAR (imagery)} & The online rating concerns streaming prediction with updates from each new temporal bucket. \\
                \emph{CORA} & The task-free rating concerns the CLEAR runs; other baselines use supplied task boundaries. \\
                \emph{EARL} & Runs are task-free with supplied goals; sparse external resets limit the non-episodic rating. \\
                \emph{LoCA} & The task-free rating concerns standard runs; buffer-clearing variants use supplied change signals. \\
                \emph{ARC-AGI-3} & Level transitions are externally marked, limiting task-free coverage to interaction within levels. \\
                \emph{Melting Pot} & The online rating includes recurrent-state updates with fixed parameters and game rules. \\
                \emph{MineDojo} & Single-stage combat runs are task-free; other groups use staged training with policy-head resets. \\
            \end{tabular}
        \end{minipage}%
    }
\end{table*}

The evaluation landscape reflects the breadth of CRL, with different protocols probing different capabilities and supporting different conclusions.
Predefined task sequences enable controlled comparisons of adaptation, forgetting, and transfer, whereas reset-free and continuing testbeds expose the consequences of persistent interaction and state distributions shaped by the agent's own behavior.
Focused diagnostics help isolate particular capabilities, while benchmarks developed for other purposes, such as general intelligence and social generalization, provide additional tests of adaptation, transfer, and planning under different protocol assumptions.
Their contribution to our CRL target depends on both the range of capabilities assessed and the extent to which those capabilities are evaluated together.

As in our discussion of the desiderata in \cref{sec:10d}, we focus on representative benchmarks, testbeds, and diagnostic studies relevant to our CRL target and do not aim for comprehensive coverage.
We assess each benchmark through its reported protocols and readouts, without extending the ratings to other possible uses of its underlying environment.
Table~\ref{tab:benchmark-coverage} summarizes which \Dref{1} conditions are met and which behavioral desiderata are probed, including the intended coverage of our companion.
Each row represents a single benchmark, with some rows covering multiple reported setups that differ in whether they meet the task-free, non-episodic, and online conditions of \Dref{1}.
For these entries, a full rating means that at least one included setup meets the condition; the three ratings may refer to different setups, as explained in the table notes and discussion below.
Notably, the ratings for our proposed companion refer to a single core configuration that combines all three \Dref{1} conditions with the indicated behavioral diagnostics.
(Details of the companion's proposed readouts are given in Table~\ref{tab:companion-diagnostics}.)
The system-level \Dref{3}--\Dref{5} lie outside this comparison because their assessment requires additional evidence about implementation, sensing operations, or resource use.

\subsection{Predefined Task Sequences}
\label{sec:landscape-sequences}

Many continual-learning and CRL benchmarks organize tasks or domains into predefined sequences.
The CLEAR imagery benchmark provides a useful non-interactive boundary case for this pattern \cite{lin2021}.
Not to be confused with the CLEAR replay method discussed under \Dref{2}, this benchmark evaluates continual prediction on a fixed collection of Internet images ordered by upload time from 2004 to 2014.
Its reported protocol divides this stream into temporal buckets with known training boundaries.
Under its streaming protocol, predictions on the next bucket precede updates from that bucket, instantiating online prediction as later data become available.
The chronological ordering captures how visual concepts change gradually over time in real-world imagery and supports evidence about stability, plasticity, and adaptation to distribution shift under \Dref{2}.
The learner does not, however, acquire observations through action or alter its subsequent experience.
Its fixed data and supplied labels therefore leave the interactive and non-episodic parts of \Dref{1}, active sensing, persistent action consequences, and the action-centered behavioral desiderata open.

Interactive CRL suites retain this staged format while adding control and increasingly rich observations.
Continual World uses predefined episodic Meta-World manipulation tasks to measure forgetting, forward transfer, performance during training, and computational or capacity costs \cite{wolczyk2021}.
CORA standardizes baselines and metrics for online performance, isolated forgetting, and zero-shot forward transfer across collections derived from Atari, Procgen, NetHack, and CHORES \cite{powers2022}.
Within each collection, observation and action spaces have the same dimensions across tasks, allowing one model with a shared policy output layer to operate throughout the sequence.
The CORA benchmark uses predefined task sequences to evaluate learning methods, including CLEAR, which learns without task identities or boundary signals, and other methods that rely on supplied task boundaries.
In CHORES, the current objective is supplied to the agent, so the benchmark tests learning how to achieve changing objectives without requiring the agent to infer which objective currently applies.
COOM introduces pixel-based embodied control through ViZDoom sequences that vary visual conditions, objectives, or both \cite{tomilin2023}.
Continual World and COOM evaluate EWC, MAS, and PackNet, while CORA evaluates EWC and CLEAR, directly connecting these suites to methods discussed under \Dref{2} \cite{wolczyk2021,powers2022,tomilin2023}.
The reported Continual World and COOM setups supply task boundaries and use task identities either to select task-specific output heads or as inputs in their single-head variants \cite{wolczyk2021,tomilin2023}.
Together, these suites provide reproducible evidence for \Dref{2} and transfer-related parts of \Dref{7}.
The suites nevertheless retain episodic resets, so task-free runs do not instantiate the complete \Dref{1} protocol.

Continual multi-objective benchmarks extend this staged design to changes in which objectives provide reward signals \cite{li2024a}.
Li et al.\ adapt discrete navigation tasks and MuJoCo locomotion environments (Hopper and Ant) to sequences of objective sets, testing policies across supplied linear preferences over all objectives encountered so far.
Pareto-front and backward-transfer measurements probe outcome trade-offs under \Dref{9} and retention and adaptation under \Dref{2}, while comparisons with learning from scratch provide transfer-related evidence under \Dref{7}.
The protocol supplies task-change signals and retains episodic resets, while the trade-off measurements leave the long-horizon and risk components of \Dref{9} untested.

\subsection{Reset-Free and Continuing Testbeds}
\label{sec:landscape-continuing}

As discussed under \Dref{1}, reducing dependence on external resets places greater demands on autonomous exploration and recovery, because the agent must learn from states produced by its own earlier behavior.
EARL examines this problem under sparse external interventions, providing a boundary case between ordinary episodic evaluation and continuing interaction \cite{sharma2022}.
Within each autonomous-learning run, supplied goals describe desired outcomes without an additional task-identity or task-change signal.
Its reported protocols allow long stretches of autonomous interaction but retain external resets to initial conditions, leaving open how learning would proceed without that assistance.
Its forward-backward reinforcement-learning baseline is closely related to the Leave No Trace method discussed under \Dref{1}, since both learn a forward policy together with a policy for autonomous recovery \cite{eysenbach2018,sharma2022}.
The evaluations probe autonomous learning and recovery in specified control tasks, but do not establish sustained adaptation across unannounced objective changes.

Lifelong Hopper and Lifelong Ant adapt MuJoCo locomotion to persistent interaction in which target velocities change while the agent's physical state carries forward \cite{lu2021}.
The online runs start from an initial transition dataset and supply the current reward function without an additional task identity or boundary signal.
Comparisons between LiSP and action-space model-predictive control, together with prediction-error measurements, test whether learned skills make model-based planning more reliable.
These readouts connect transfer across supplied velocity targets under \Dref{7}, predictive model use under \Dref{8}, and skill-space planning under \Dref{10}, but do not isolate reward-independent acquisition or the coordination of overlapping subplans.
Performance across velocity changes provides partial evidence for \Dref{2} without separately measuring retention.

Jelly Bean World (JBW) provides a configurable testbed for never-ending learning through procedurally generated worlds, local observations, and persistent interaction \cite{platanios2020}.
It can additionally support multi-task, multi-agent, multimodal, and curriculum settings, although which desiderata a JBW experiment probes depends on its configuration and evaluation protocol.
Anand and Precup's PT-DQN experiment uses JBW for one continuing interaction with recurring reward preferences and no observed task identity \cite{anand2023}.
The reward reversals follow a fixed schedule but are not announced, so this experiment instantiates \Dref{1}.
Its reward-rate measurements probe adaptation to recurring changes under \Dref{2}, but do not separately isolate retention of earlier abilities.
JBW also provides the environmental basis for the playable companion in Section~\ref{sec:benchmark-proposal}, where we adapt its continuing procedural world to combine selected behavioral pressures in one scenario.

Forager and AgarCL both study continual learning in continuing environments under partial observability but make different experimental trade-offs.
Forager provides a lightweight and configurable grid world with constant simulator memory, adjustable fields of view, hidden reward changes, and an unending variant in which the agent's behavior helps generate new foraging tasks \cite{tang2026}.
Its experiments connect the \Dref{1} protocol to plasticity under \Dref{2} and state construction relevant to \Dref{6}: plasticity mitigations can preserve trainability, while reward traces and recurrent agents additionally help recover information missing from the current observation.
These experiments do not, however, test persistent retrieval across substantial delays or isolate long-range credit assignment.

AgarCL instead provides a richer visual environment with partial observability, hybrid control, endogenous dynamics, and opponents \cite{mohamed2025}.
Its primary game continues when the controlled agent dies, although the agent respawns, while accompanying mini-games isolate selected sources of difficulty under episodic and continuing protocols.
The reported evaluations include continual backpropagation and recurrent agents, but primarily establish that the resulting learning problem remains difficult rather than demonstrating stability, persistent memory, or long-range credit.
AgarCL therefore offers a broad challenge and several focused diagnostics, while aggregate performance in its full game remains difficult to attribute to any one desideratum.

\subsection{Complementary Benchmarks and Diagnostics}
\label{sec:landscape-complementary}

Several benchmarks define specialized evaluation protocols for existing environments.
Continual World organizes Meta-World tasks into continual-learning sequences, COOM does the same with ViZDoom, and URLB separates reward-free pretraining from downstream adaptation in the DeepMind Control Suite \cite{wolczyk2021,tomilin2023,laskin2021}.
Their relevance to the desiderata follows from these experimental protocols, including their resets, changes, supplied information, learning opportunities, and measurements, rather than from the underlying environments alone.
These focused protocols are valuable because they isolate particular capabilities and dependencies under controlled conditions.
Relative to our broader CRL target, they provide complementary evidence about individual desiderata without establishing that the relevant capabilities coexist in one agent throughout a continuing lifetime.

URLB's separation of reward-free pretraining from downstream adaptation directly tests reward-independent acquisition and later reuse under \Dref{7} \cite{laskin2021}.
Its baselines include the Intrinsic Curiosity Module and DIAYN, both discussed under \Dref{7}.
CausalWorld's reported zero-shot tests assess trained policies under controlled interventions and combinatorial variation in robotic manipulation \cite{ahmed2021}.
This provides controlled evidence relevant to transfer under \Dref{7} and a setting for investigating causal structure learning under \Dref{8}.
CompoSuite's zero-shot tests evaluate generalization to held-out combinations of robot, object, objective, and obstacle, providing focused evidence for reusable knowledge under \Dref{7} and compositional behavior relevant to \Dref{10} \cite{mendez2022a}.
Its modular baseline uses the functional decomposition studied by Mendez et al.\ under \Dref{4} \cite{mendez2022a,mendez2022b}.
For both benchmarks, the generalization tests considered here assess trained policies without further learning.
URLB's separate pretraining and adaptation phases, CausalWorld's controlled causal factors, and CompoSuite's held-out evaluation splits make the respective capabilities easier to isolate and evaluate.
These evaluations nevertheless do not instantiate the complete task-free, non-episodic online protocol of \Dref{1}.

Some evaluations use focused diagnostics to distinguish capabilities that a single overall performance score cannot separate.
LoCA evaluates whether PlaNet and DreamerV2 can revise one local relation in an otherwise stable environment without relearning unaffected dynamics, thereby isolating the stability--plasticity requirement for predictive world models under \Dref{8} \cite{wan2022}.
Standard runs do not receive phase-change signals; separate buffer-clearing variants use that information to remove earlier data.
Controlled memory and credit-assignment diagnostics, including Ni et al.'s passive and active T-Maze tests \cite{ni2023}, vary observation delay, action--outcome delay, distractors, or relevant history to separate retrieval from temporal credit assignment under \Dref{6} \cite{pignatelli2024}.
These designs distinguish memory span from credit-assignment span, but evaluating them separately does not establish both capabilities in the same agent throughout a continuing interaction.

Benchmarks from related research areas broaden the evaluated behavior without adopting a CRL protocol.
The ARC-AGI series is explicitly presented as measuring progress toward AGI through the efficiency with which systems adapt to unfamiliar problems.
ARC-AGI-1 and ARC-AGI-2 use static grid transformations, asking systems to infer novel rules from a few examples and apply them to new inputs, whereas ARC-AGI-3 moves this evaluation into an interactive setting \cite{arcprize2026}.
ARC-AGI-3 consists of novel, turn-based games in which agents discover the rules and goals through interaction, with later levels designed to require reuse and composition of what they have learned.
Public demonstration games and private evaluation games share the same interaction format and scoring protocol, with performance measured by action efficiency relative to human players \cite{arcprize2026}.
The behavioral ratings in Table~\ref{tab:benchmark-coverage} are based on publicly documented evidence and may be incomplete because the mechanics of individual private games are not publicly available.
Reported stateful agents accumulate interaction history or update learned state during play, instantiating online adaptation under \Dref{1}.
The partial task-free rating distinguishes unsegmented interaction within a level from the benchmark's externally marked progression between levels.
The interactive setting provides evidence relevant to \Dref{5}, \Dref{8}, and \Dref{10}, while the benchmark's separate levels, terminal success conditions, and reset action do not constitute one uninterrupted lifetime with persistent cross-level consequences.

Melting Pot's social generalization tests assess behavior with unfamiliar co-player populations across diverse strategic interactions, making the benchmark relevant to the multi-agent motivation for CRL and coordination under \Dref{10} \cite{leibo2021}.
Its comparisons of individual and shared-reward training in social dilemmas probe how the reward criterion affects cooperation and susceptibility to exploitation, providing partial evidence for \Dref{9}.
Each test keeps the underlying game fixed without supplying the focal agent with the background population's identity.
Policy parameters remain fixed during zero-shot testing, while recurrent policies update their internal state from interaction history, instantiating the online component of \Dref{1}.
Its test scenarios preserve familiar environment rules and dynamics while varying the other agents, allowing social generalization to be isolated but leaving adaptation to changing dynamics or objectives open.

MineRL pairs a large dataset of human demonstrations with a Minecraft simulator for tasks including navigation, resource gathering, and crafting \cite{guss2019}.
Its reported agent baselines cover a small set of tasks, while its analysis of crafting hierarchies concerns human demonstrations.
The reinforcement-learning baselines update from new interaction; a separate behavioral-cloning baseline learns from demonstrations alone.
MineDojo builds on MineRL's codebase with unified observation and action spaces, thousands of language-specified tasks, and an internet-scale multimodal knowledge base \cite{fan2022}.
Policies learn online from interaction while the pretrained visual encoder and reward model remain fixed.
The task-free rating concerns single-stage training on combat tasks; other task groups use externally scheduled stages with policy-head resets.
The reported multitask and zero-shot visual generalization results provide partial evidence for transfer under \Dref{7}.
The shared Minecraft setting offers opportunities to study skill reuse under \Dref{7} and hierarchical planning under \Dref{10}, since different objectives can depend on shared resources, tools, or intermediate goals.
However, the reported agent evaluations do not directly test learned hierarchical composition or the coordination of overlapping subplans.
Their episodic evaluation protocols also leave open whether one agent can continually acquire, preserve, and compose skills under the complete \Dref{1} protocol.

\subsection{The Remaining Evaluation Gap}

The evaluations reviewed above provide complementary evidence rather than incomplete versions of one ideal benchmark.
Benchmarks based on predefined task sequences support controlled and reproducible measurements of online performance, forgetting, and transfer, while reset-free and continuing worlds expose persistent consequences, autonomous exploration, state construction, and long-term adaptation.
Focused diagnostics strengthen attribution for individual mechanisms, whereas richer environments can reveal integration failures that isolated tests miss.
AgarCL and Melting Pot illustrate why multi-agent systems provide natural but difficult settings for evaluating aspects of CRL \cite{mohamed2025,leibo2021}.
Other agents generate endogenous feedback, and when they also learn, their changing policies make experience non-stationary, so multi-agent testbeds naturally place several behavioral desiderata under simultaneous pressure \cite{nekoei2021,kim2021}.
Performance in such a testbed can therefore provide suggestive evidence that several capabilities work together, but a high score neither establishes our broader CRL target nor identifies which desiderata account for success.
A low score is equally difficult to diagnose without controlled variations and capability-specific readouts.
Within this benchmark selection, \Dref{9} receives only partial diagnostic coverage through tests of competing objectives and indirect comparisons in social dilemmas.
None of the reviewed evaluations establishes all the behavioral cores of \Dref{2} and \Dref{6}--\Dref{10} together in a single agent operating under the task-free, non-episodic online protocol of \Dref{1}.
The playable companion introduced in Section~\ref{sec:benchmark-proposal} builds on JBW's design to combine selected behavioral pressures under \Dref{1} (other testbeds could likewise be adapted to study such combinations).
Its purpose is not to certify CRL or close the entire evaluation gap, but to contribute a compact combination of rules and capability-specific readouts.

\section{A Playable Companion to the Desiderata}
\label{sec:benchmark-proposal}

The ten desiderata are the central contribution of this paper.
The compact game introduced below serves as a playable companion that helps illustrate their interplay.
Its scenario, suggested default configuration, evaluation protocol, and diagnostic readouts also define a proposed benchmark.
This dual role illustrates how several behavioral desiderata can matter simultaneously within one task-free, non-episodic online lifetime.
Its deliberately compact rules make their interactions easier to inspect and provide a shared example through which the behavioral desiderata can be examined together.

We present the companion to illustrate how the desiderata interact and to propose readouts for studying those interactions.
Providing an agent-evaluation package, calibrating the protocol, and comparing representative agents under declared resource budgets extend beyond this explanatory focus and remain future work.
The diagnostic scope reflects the aim of retaining compact, interpretable rules.
The core does not directly test counterfactual reasoning or continual world-model revision, and extending this coverage while preserving simplicity remains an open design problem.
Full assessment of the system-level aspects of \Dref{3}--\Dref{5} also requires additional implementation-specific evidence.
The compact rules nevertheless create a setting in which several behavioral desiderata matter together, with decisions about exploration, preparation, and resource use linking their associated capabilities.

\begin{readerpromptbox}
    Try the game as a human player at \href{https://fractalbrain.ai/10d/game}{https://fractalbrain.ai/10d/game} before reading the rules below.
    Confronting the world without prior explanation mirrors the information initially available to an evaluated agent, whereas the remainder of this section necessarily contains spoilers.
\end{readerpromptbox}

\noindent The scenario draws on the seeded procedural world and uninterrupted interaction of Jelly Bean World (JBW) \cite{platanios2020}, although its rules are engine-independent and could be instantiated in JBW or another suitable environment.
The human-playable implementation linked above is one realization of these rules.

\begin{figure}[!t]
    \centering
    \includegraphics[width=\columnwidth]{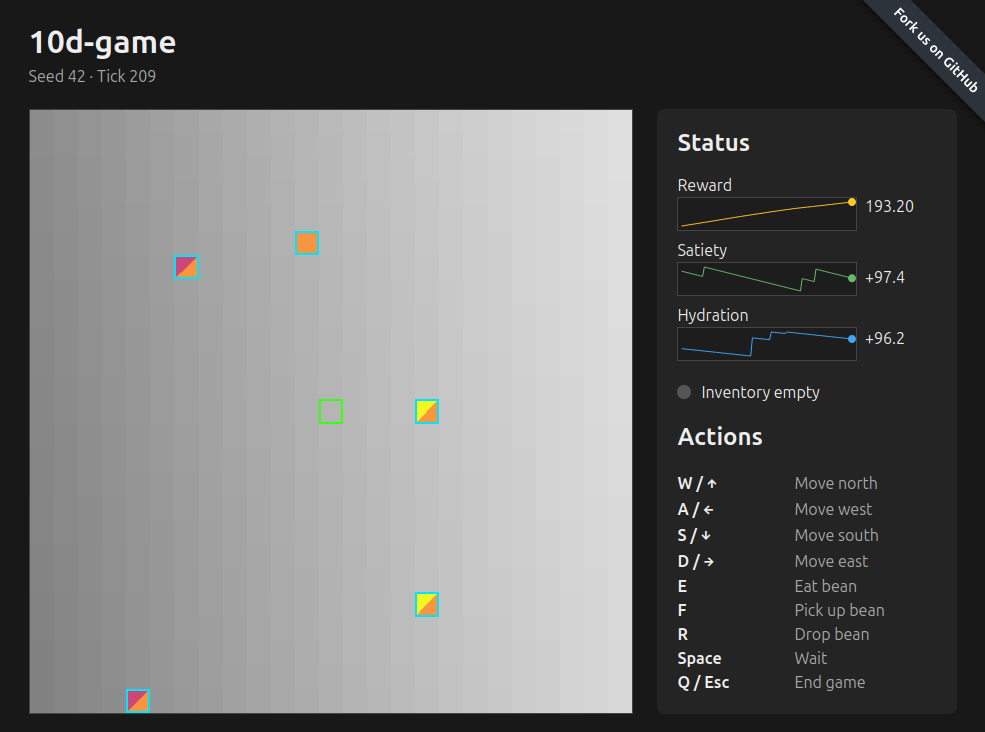}
    \caption{Human-playable realization of the companion scenario.
    The grid shows the agent's egocentric visual field with color-coded beans.
    For visual clarity, the interface renders the blue-channel illumination as a grayscale gradient.
    The status panel shows cumulative reward, satiety, hydration, inventory occupancy, and the available controls.
    The graphical interface is illustrative; the scenario and proposed benchmark remain engine-independent.}
    \label{fig:companion-screenshot}
\end{figure}

The core scenario is parsimonious by design.
Richer testbeds such as AgarCL deliberately combine many sources of difficulty, whereas this companion asks how much of the interplay among \Dref{2} and \Dref{6}--\Dref{10} can be expressed through compact, interpretable game rules in a two-dimensional grid world \cite{mohamed2025}.
Human players may consequently find the result exceptionally boring, but entertainment was never a design objective.
The aim is to examine how the targeted capabilities work together before adding complexity that could make their contributions and failures harder to interpret.
Collectively, the core mechanics were chosen to make the targeted desiderata behaviorally relevant and to create interactions among their associated capabilities.
Consequently, the fence diagnostic described in Section~\ref{sec:fence-diagnostic} remains optional because it only makes existing memory and planning probes more demanding and leaves the core scenario's desideratum coverage unchanged.
The proposed benchmark does not fix readouts for the system-level \Dref{3}--\Dref{5}, whose assessment requires implementation-specific evidence about resource use, functional organization, and sensing operations.
Users can supplement the shared behavioral diagnostics with suitable measurements for their system.

Forager already combines partial observability, biome structure, hidden reward changes, and an unending variant in which the agent's behavior helps generate new tasks \cite{tang2026}.
AgarCL and JBW provide further precedents for persistent interaction and interacting game mechanics \cite{platanios2020,mohamed2025}.
Here, the salty bean's delayed hydration cost and the one-slot inventory create preparation and recovery problems while the moving blue-light objective competes with resource maintenance.
Recurring color variants across biomes create opportunities to test retention and transfer.
The intended contribution is this combination of mechanics and capability-specific readouts within one continuing lifetime (Table~\ref{tab:companion-diagnostics}).

\subsection{World and Observation}

The scenario uses a seeded procedural two-dimensional world with uninterrupted interaction.
The seed and spatial coordinates determine the initial content of each cell independently of the order in which locations are visited, so movement reveals a persistent world instead of determining which world is generated.
The agent observes a fixed $25\times25$ egocentric RGB field of view together with its current satiety, hydration, and whether its one-slot inventory is occupied.

The first two visual channels encode bean color.
Every two-channel color identifies exactly one flavor throughout a lifetime, although one flavor may have several color variants.
The world is partitioned into unannounced square color biomes, within each of which a flavor has one consistent color.
The color variants are drawn from a configurable finite set, so a color encountered in one biome may recur in another.
Crossing a biome boundary may expose either previously encountered or unfamiliar colors for known flavors, while returning to an earlier biome restores its previous visual mapping.

The blue channel carries no flavor information and instead represents illumination from a distant ``blue sun.''
At each tick, the agent receives a reward contribution proportional to the illumination at its current position.
With position and time measured relative to the agent's initial state, so that $\boldsymbol{x} = (0, 0)^{\!\top}$ and $t=0$ initially, its intensity is
\begin{equation}
    I_{\!B}(\boldsymbol{x}, t) = 127.5 \big[ 1 + \cos(\boldsymbol{k} \cdot \boldsymbol{x} + \omega t) \big],
\end{equation}
with $\boldsymbol{k} = 2\pi/\lambda \, (\sin\phi, \cos\phi)^{\!\top}$ and $\omega = 2\pi / T$.
The configurable parameters $\lambda>0$ and $T>0$ set the spatial and temporal periods, respectively, while the world seed determines the direction $\phi$.
The default periods make the bright bands move slowly enough that tracking them creates an ongoing objective while resource maintenance creates shorter-term demands.
The light field is independent of bean colors and flavors, and the default reward contribution at the agent's position is $I_{\!B}(\boldsymbol{x},t)/255$.
Coordinates, global time, light-field parameters, biome boundaries, and bean flavors are absent from the observation.
The fixed action set contains no inspection or other dedicated sensing action.

\subsection{Beans and Resources}

Four bean flavors affect the agent's resources or reward as summarized in Table~\ref{tab:bean-catalogue}.
Their visual colors vary across biomes, but their effects remain fixed throughout the lifetime.

\begin{table*}[!t]
    \caption{Bean catalogue and resource interactions for the core scenario.
    Each color in visual channels 1--2 identifies exactly one flavor, although a flavor may be represented by multiple colors.}
    \label{tab:bean-catalogue}
    \centering
    \renewcommand{\arraystretch}{1.2}
    \begin{tabularx}{\textwidth}{
        @{}
        >{\raggedright\arraybackslash}p{0.12\textwidth}
        >{\raggedright\arraybackslash}p{0.25\textwidth}
        >{\raggedright\arraybackslash}p{0.24\textwidth}
        >{\raggedright\arraybackslash}X
        @{}
    }
        \toprule
        \textbf{Flavor} & \textbf{Immediate effect} & \textbf{Delayed effect} & \textbf{Functional role} \\
        \midrule
        Satiety & Increases satiety by a modest amount.& None.& Supports routine satiety maintenance as satiety decays.\\
        Hydration & Increases hydration by an amount calibrated to offset the delayed hydration loss from one salty bean.& None.& Supports routine hydration maintenance as hydration decays and provides prepared recovery after salty-bean use.\\
        Bitter & Applies a reward penalty.& None.& Makes indiscriminate consumption costly, rewarding agents that learn and use color--effect associations.\\
        Salty & Increases satiety; this benefit is calibrated independently of the later hydration loss.& Decreases hydration after a fixed delay.& Creates a useful but risky action: exploit it when hydration recovery is prepared and avoid it otherwise.\\
        \bottomrule
    \end{tabularx}
\end{table*}

The suggested default configuration is summarized in Table~\ref{tab:companion-defaults}.
Let $s_t$ and $h_t$ denote satiety and hydration, both bounded to $[-100,100]$, initialized at $100$, and reduced by a fixed amount on every tick.
Values below zero contribute the resource penalty
\begin{equation}
    p_t=\min(s_t,0)+\min(h_t,0).
\end{equation}
The reward on a tick is the sum of this penalty, the blue-light contribution, and any immediate bitter-bean penalty.
Resource changes saturate at these bounds, and consuming a restorative bean at the upper bound still removes it from the world.
The bitter bean instead imposes an immediate reward penalty without changing either resource, making indiscriminate consumption costly.
Encountering an unfamiliar color may require one risky first trial, but retaining the resulting color--effect association can prevent the same mistake later.

The action space consists of cardinal movement, a no-op, eating the bean under the agent, and picking up or dropping a bean at the current location.
Every action advances the same continuing lifetime.
The one-slot inventory can hold a bean of any flavor, but no additional bean can be picked up while the slot is occupied.
The observation reveals only whether the slot is occupied, not which flavor it contains.
The inventory allows the agent to reserve a hydration bean until the delayed hydration loss from a salty bean occurs, since consuming it earlier may waste some of its restorative effect when hydration is near its upper bound.
The occupancy-only observation additionally requires the agent to retain which flavor it collected.

The salty bean separates an immediate satiety benefit from a hydration loss after a configurable delay.
One hydration bean offsets this loss under the default configuration, making a salty bean useful when recovery has been prepared---for example, by carrying a hydration bean or remaining near one---but usually harmful when recovery must begin from scratch.
The nonzero delay connects an action to a later consequence across intervening behavior, although the short default separation is not by itself evidence of long-range credit assignment.
Increasing this delay could create a more focused credit-assignment diagnostic, but no architecture-neutral threshold determines when a delay becomes ``long.''
A long-delay variant would need separate calibration to avoid defining the test merely through one model's context length or making informative consequences too rare within a finite evaluation.

Consumed beans do not regrow.
Sustaining both resources over a sufficiently long lifetime therefore requires leaving depleted regions and eventually encountering new color biomes.
The resulting visual changes depend on the agent's path through the persistent world instead of an imposed task schedule.
Encounters with unfamiliar and recurring color variants across biomes create opportunities to test acquisition of new mappings, retention and reuse of earlier ones, transfer among variants, and continued plasticity.

Restorative and salty-bean effects can change internal state without producing an immediate resource-related reward or penalty because $p_t=0$ whenever $\min(s_t,h_t)\geq 0$.
Bitter beans remain immediately penalized, and the unrelated blue-light reward remains active, so this experience is reward-neutral only with respect to the resource-changing effects.
When ordinary decay or a salty-bean effect later makes a resource negative, knowledge acquired earlier becomes relevant to reward without an announced phase transition.
The first-recovery diagnostic therefore combines reward-independent acquisition under \Dref{7} with retention, retrieval, and credit under \Dref{6}; its behavioral result does not isolate either capability or identify a unique internal mechanism.

\begin{table}[!t]
    \caption{Suggested default configuration for the core scenario.
    Signs denote changes to resource levels or reward.}
    \label{tab:companion-defaults}
    \centering
    \begin{tabularx}{\columnwidth}{
        @{}
        >{\raggedright\arraybackslash}X
        >{\raggedright\arraybackslash}p{0.32\columnwidth}
        @{}
    }
        \toprule
        \textbf{Setting} & \textbf{Suggested default} \\
        \midrule
        Field of view & $25\times25$ cells \\
        Inventory capacity and initial state & $1$ bean; empty \\
        Color variants per flavor & $4$ \\
        Bean probability per cell & $0.01$ \\
        Color-biome side length & $1000$ cells \\
        Resource range & $[-100,100]$ \\
        Initial satiety and hydration & $100$ each \\
        Decay of each resource & $0.1$ per tick \\
        Satiety-bean effect & $+3$ satiety \\
        Hydration-bean effect & $+10$ hydration \\
        Salty-bean immediate effect & $+10$ satiety \\
        Salty-bean delayed effect & $-10$ hydration \\
        Salty-bean delay & $10$ ticks \\
        Bitter-bean effect & $-10$ reward \\
        Sunlight spatial period $\lambda$ & $200$ cells \\
        Sunlight temporal period $T$ & $400$ ticks \\
        Maximum sunlight reward & $+1$ per tick \\
        \bottomrule
    \end{tabularx}
\end{table}

Because the four flavors are equally likely, the default bean probability of $0.01$ corresponds to a probability of $0.01/4=0.0025$ per cell for any particular flavor.
At the default decay rate, the $+3$ gain from one satiety bean offsets only $30$ ticks of satiety decay.
Relying on satiety beans alone is therefore intended to leave little slack once search and travel are taken into account, even for an agent that pursues them aggressively.
A salty bean instead offsets $100$ ticks of satiety decay; when its later hydration cost is anticipated and compensated, this additional margin permits more exploration and more time near the moving sunlight maximum.
The salty bean is consequently not merely a delayed hazard, but a useful resource whose value depends on preparation and the agent's competing objectives.

With these defaults and no bean effects, each resource takes $1000$ ticks to decay from its initial level to zero, compared with the salty bean's $10$-tick delay.
The long separation in the joint \Dref{6}--\Dref{7} diagnostic therefore comes from observing resource effects while the initial reserve keeps them reward-neutral and applying that knowledge at a much later deficit, not from treating the salty-bean delay itself as long-range credit assignment.

The sunlight and resource mechanics also create plans at different temporal scales.
Tracking a moving bright band is an ongoing objective, but falling resources may make it advantageous to interrupt that pursuit, forage for beans, and later resume tracking.
Safe salty-bean use admits a nested plan in which the agent secures a hydration bean, retains it while exploiting the salty bean, and completes recovery after the delayed loss.
The recovery plan can remain pending while actions serve the sunlight objective or other resource needs, so the sequential action stream interleaves several ongoing concerns.
Reusing this preparation-and-recovery pattern in unfamiliar configurations and resuming sunlight tracking after maintenance would provide behavioral evidence relevant to both the hierarchical and concurrent parts of \Dref{10}, without revealing whether the agent represents the hierarchy explicitly.
The associated readouts provide partial coverage of \Dref{10} because they do not isolate reusable composition from behavior specialized to this preparation-and-recovery pattern.

These mechanics also exercise two of the three outcome trade-offs under \Dref{9}.
The salty bean couples an immediate benefit to a delayed cost, while sunlight reward, satiety, and hydration create competing state-dependent objectives.
Its delayed loss is deterministic and the default preferences remain fixed, so the core offers little evidence about rare risks or continual revision of preferences.

\begin{table*}[!t]
    \caption{Mapping from six behavioral desiderata to situations that may arise during a continuing lifetime and to associated readouts.
    The core situations may all occur within the same task-free, non-episodic online interaction under \Dref{1}, and whether or when they occur depends on the agent's behavior.
    The mapping is not one-to-one: \Dref{6} and \Dref{7} overlap in the same lifetime situation, while the salty-bean situations couple retention and credit under \Dref{6}, prediction under \Dref{8}, and evaluation under \Dref{9}.
    These proposed readouts provide the basis for the companion's coverage ratings in Table~\ref{tab:benchmark-coverage}.}
    \label{tab:companion-diagnostics}
    \centering
    \renewcommand{\arraystretch}{1.2}
    \begin{tabularx}{\textwidth}{
        @{}
        c
        >{\raggedright\arraybackslash}p{0.50\textwidth}
        >{\raggedright\arraybackslash}X
        @{}
    }
        \toprule
        & \textbf{Lifetime situation} & \textbf{Associated readouts} \\
        \midrule
        \Dref{2} & Depletion of non-regrowing local beans and crossings among biomes that expose unfamiliar or recurring color variants after intervening experience. & Acquisition of new mappings, retained color--flavor knowledge when variants recur, forward transfer, and late-life plasticity. \\
        \Dref{6} \& \Dref{7} & Early observation of resource-changing bean effects while both resources remain nonnegative, followed much later by the first endogenous deficit. The salty bean adds a short action--consequence separation, while inventory use and the optional fence introduce additional memory demands. & First corrective action, appropriateness of the first bean eaten, steps and cumulative deficit until recovery, and timely use of a stored hydration bean after the delayed loss, stratified by prior exposure and elapsed interval; separate summaries for the short-delay and memory-dependent situations. \\
        \Dref{8} & First encounters with previously unseen salty-bean configurations after the agent has separately experienced salty- and hydration-bean effects, with recovery either already prepared or requiring additional search. & First-encounter conditional salty-bean use, stratified by prior exposure to the component effects; exploiting prepared opportunities while rejecting unprepared ones provides evidence for composition beyond retrieval of an exact earlier episode. \\
        \Dref{9} & A deterministic immediate satiety benefit followed by a delayed hydration loss that requires prepared recovery, while both resources decay and the blue-light objective continues. & Resource-wise outcomes, frequency and cost of unprepared salty-bean use, blue-light reward forgone during preparation and recovery, and state-conditional quality of recovery preparation. \\
        \Dref{10} & A long-term objective of tracking moving regions of strong blue light combined with short-term resource maintenance and intermediate preparation for salty-bean use. & Blue-light tracking while resources are safe, appropriate interruption for maintenance, successful preparation, and resumption of blue-light pursuit. \\
        \bottomrule
    \end{tabularx}
\end{table*}

\subsection{Persistent Lifetime Protocol}

The protocol consists of one continuing lifetime without a terminal signal, reset, task identity, or revealed biome boundary.
Resource levels, inventory contents, pending hydration losses, consumed beans, and other modifications to the world persist throughout it.
Crossing a biome boundary changes only the visible color variants encountered at new locations; it does not change the objective, action, or observation spaces or the learning procedure.
Biome crossings are therefore changes within one environment instead of substantive task boundaries under \Dref{1}.
This differs from the predefined task-sequence protocols reviewed in Section~\ref{sec:benchmark-coverage}, which organize piecewise-stationary tasks into externally defined sequences.

An evaluation necessarily observes only a finite prefix of this continuing world.
A useful duration would allow several biomes to be reached and both unfamiliar and recurring color variants to be encountered across substantial intervals, while leaving the actual encounters dependent on the agent's behavior.
Because no such duration has yet been calibrated, the proposed configuration leaves it as a prospective evaluation parameter.

Multiple procedural seeds can expose agents to different worlds while preserving the same rules and initialization procedure.
When seeds are used to estimate variability, each run is most naturally treated as an independent lifetime with the learner initialized anew.
Carrying learned state from one seed to the next instead creates an evaluator-imposed sequence of world resets and constitutes a different protocol.
For reproducible agent comparisons, the generator, configuration, and evaluation duration should be fixed and published before evaluation, with evaluation seeds held out during development and released with the results.

\subsection{Optional Fence Diagnostic}
\label{sec:fence-diagnostic}

The compact core already places the targeted behavioral desiderata under interacting pressure, but an optional fence can deepen its memory and planning probes without changing that coverage.
This variant adds an item that blocks movement without occluding vision, a mechanic that can be realized directly in JBW \cite{platanios2020}.
Every fence has the same fixed color in the first two channels, distinct from all bean colors, while its blue channel continues to show the local sunlight intensity.
A bean can therefore remain visible through a fence even when the agent cannot move directly toward it.

Fence orientation and length determine the required detour.
In diagnostic layouts, reaching a visible bean can require initially moving away from it, losing it from the field of view, reaching an end of the fence, and then returning toward its remembered location.
Longer fences increase the interval over which that information must guide behavior, while layouts in which greedy visual pursuit or distance reduction fails make direct attraction insufficient.
Circumnavigation consequently provides additional evidence relevant to memory under \Dref{6} and action-conditioned route prediction under \Dref{8}, while also supplying an intermediate-timescale subplan for the trade-offs of \Dref{9} and coordination under \Dref{10}.

\subsection{Behavioral Readouts}

Because the mechanics interact, interpreting situations and readouts across the lifetime is more informative than assigning one score to each desideratum or collapsing the complete lifetime into one return.
Table~\ref{tab:companion-diagnostics} therefore relates each behavioral desideratum to situations that may arise within the continuing interaction and to associated readouts, without claiming a one-to-one mapping.
Complete online curves can reveal acquisition, transient loss, and recovery that endpoint summaries would hide \cite{delange2023}, while tests of acquisition at later lifetime stages expose whether plasticity is maintained \cite{dohare2024}.

Useful traces include total reward and its separate blue-light, resource, and bitter-bean components; satiety and hydration; resource-deficit crossings and recovery times; bean consumption; inventory use; delayed hydration losses; and biome crossings and returns.
Prepared and unprepared salty-bean encounters expose whether behavior changes appropriately with the availability of recovery, while the optional fence adds detour length and success after a target leaves view.
Relating these quantities to cumulative interactions makes acquisition efficiency visible.
Elapsed time and cumulative computation provide complementary axes because actions differ in consequence and interaction counts omit internal processing.
Storage and computation trajectories can further contextualize behavioral results \cite{harun2023} without turning the scenario itself into an assessment of \Dref{3}.

The first endogenous resource deficit deserves a separate analysis because it links early reward-neutral exposure to later use.
Stratifying recovery by prior exposure, elapsed interval, and whether the relevant color is familiar or novel can distinguish direct recall from transfer to a new variant.
Runs that end before either resource becomes negative are right-censored for the first-recovery analysis and may be reported separately as resource maintenance.
Because the agent has not yet observed the resource penalty, such behavior alone does not establish learned anticipation.
More appropriate or faster recovery after prior reward-neutral exposure would provide evidence that earlier resource-effect experience influenced later behavior.
Without an otherwise matched comparison lacking that exposure, the proposed stratification provides only partial evidence for \Dref{7} and does not cleanly separate the benefit of earlier acquisition from learning after resource penalties appear.

Independent seeded lifetimes expose variation induced by different worlds.
Interval estimates and distributional summaries would convey this variability more clearly than point estimates alone \cite{agarwal2021,patterson2024}.
A profile of these readouts can support comparisons among agents while retaining the trade-offs that a single aggregate leaderboard score would hide.

\subsection{What This Benchmark Can and Cannot Show}

We intend the proposed benchmark to create diagnostic pressure, not to certify that an agent is a lifelong learner.
Evidence from the proposed benchmark is asymmetric: failure may expose an integration problem in the evaluated agent, whereas success supports only the narrower claims associated with its individual readouts.
Any evaluation observes only a finite part of the lifetime we intend to model and therefore cannot prove that learning will continue indefinitely.
Learning, adaptation, and learning to adapt are conceptually distinct, but a finite behavioral evaluation cannot cleanly separate them; the classification also depends on what changing state is included in the agent description \cite{abel2023}.
Nor should lifelong competence require parameters to change forever: an agent may learn a sufficiently general adaptation procedure in finite time and subsequently apply it with fixed learned parameters.
This is compatible with the behavioral target if incoming evidence can still change persistent internal state and future behavior when relevant.
Conversely, once every permitted variation can be handled without incorporating further experience, the benchmark has reached a ceiling and no longer provides pressure for continual learning.
One reason is sample efficiency.
In a continuing lifetime, experience is not freely repeatable: actions alter subsequent states, opportunities may disappear, and the world or the agent's resources may drift.
Episodic learners are also limited by time and compute, but resets make repeated sampling of essentially the same situation available in principle.
A useful CRL agent must therefore be ``sufficiently'' sample efficient, even though no task-independent threshold for sufficiently exists \cite{naik2019,abel2023}.
Chollet frames intelligence as skill-acquisition efficiency relative to priors, experience, and generalization difficulty \cite{chollet2019}.
We do not seek to define intelligence, and efficient skill acquisition is only one part of our CRL target; the relevant lesson here is that final skill alone cannot establish continual learning.
This does not mean that every encounter must be learned in isolation: a world model that captures reusable regularities can support generalization to unseen situations and thereby reduce the need for repeated experience \cite{ha2018,hafner2025}.

Our protocol can require one held-out evaluation lifetime, but it cannot erase an agent's development history.
One could train episodically on many benchmark instances, tune against the public generator, and deploy the resulting agent only once.
Such training may encode a benchmark-specific solution, but it may also build exactly this kind of reusable world model.
Held-out seeds reduce direct memorization, but they cannot cleanly separate benchmark-specific pretraining from legitimate generalization.
Indeed, inferring that a new green bean has the same flavor as a previously encountered red bean is exactly the kind of transfer and world knowledge targeted by \Dref{7} and \Dref{8}.
At the behavioral interface, the boundary remains fuzzy.
Results should therefore disclose all pretraining data and simulator interactions, tuning and model-selection budgets, and whether parameters or persistent state change during the evaluated lifetime; sample efficiency within that lifetime should be reported separately from development cost.

The integrated setting has a related ambiguity because its behavioral situations can exercise several desiderata at once.
For example, salty-bean opportunities require earlier effect information to remain available under \Dref{6}, to be composed into action-conditional predictions under \Dref{8}, and to be evaluated as a trade-off under \Dref{9}.
The corresponding behavioral readouts do not isolate these contributions, and the core scenario neither changes the underlying effect relations nor queries predictions directly, so it does not establish counterfactual reasoning, causal correctness, or continual world-model revision.
We deliberately combine diagnostics so that a single fixed agent and configuration must face them all; using a separately tuned architecture for each diagnostic would miss the point.
Integration also exposes cross-capability interference that isolated diagnostics can miss.
An agent may retain competence within one problem family yet lose it while adapting representations, memory, or control for another behavioral capability.
\Dref{2} is therefore cross-cutting: capability-specific performance should be tracked throughout the same lifetime so that passing each diagnostic in isolation is not mistaken for preserving the combined competence portfolio.
Yet behavior alone cannot reveal whether that system is monolithic or a router over specialists.
Nor is modularity automatically a failure: autonomous routing, reuse, and composition may themselves help satisfy the desiderata.
The operational requirement is one executable system, with no external task label or manual model switch, a fixed selection procedure, and one shared resource budget across the lifetime.
We therefore require implementation disclosures and ablations to determine whether the components were trained independently and whether their coordination is learned; the benchmark score alone cannot establish either point.

The controlled abstraction that makes the benchmark diagnostically useful also limits what success can establish.
Moravec's paradox observes that artificial systems may perform well on tasks humans regard as intellectually demanding while struggling with perceptual and sensorimotor abilities that humans perform with little conscious effort \cite{moravec1988}.
Our compact two-dimensional world uses simplified visual observations and discrete actions, leaving much of this difficulty outside the benchmark.
Success would therefore provide evidence about the targeted continual-learning capabilities without establishing general embodied competence.

The benchmark's simplicity creates a second interpretive limitation: a sufficiently large pretrained system may solve it largely through scale without demonstrating continual learning during the evaluated lifetime.
Generalist transformer policies already act across diverse control domains, embodied language models transfer broad prior knowledge into planning, and language-model-based agents explore open-ended simulated worlds while accumulating executable skills \cite{reed2022,driess2023,wang2024b}.
We therefore conjecture that a sufficiently large language-model-based agent could solve the proposed benchmark, with or without an explicit reasoning scaffold.
Such a result would be behaviorally valid, but it would provide limited evidence for CRL if a frozen model arrived with most of the solution or brute-force computation substituted for continued learning.
Because the environment and its interaction rules are deliberately compact, a comparatively compact system that acquires the necessary behavior during its evaluated lifetime would provide stronger evidence.
Parameter count, pretraining data and compute, persistent storage, inference cost, and comparisons with simpler baselines should therefore accompany behavioral results.
This benchmark serves primarily as a concrete companion to our desiderata and, at best, as a proposal to the community for testing several of them together; it is not a definitive test of CRL.

\section{Conclusion}

The central lesson of this paper is that CRL is an integration problem.
Evidence for stability, memory, world modeling, or planning in isolation does not establish that these capabilities can coexist and remain usable throughout one continuing lifetime.
Our ten desiderata offer an opinionated map of this problem across protocol conditions, system-level properties, and behavioral capabilities.
They are neither a definition of CRL nor ten independent pass/fail tests.
The representative literature shows substantial progress on individual parts, but we are not aware of a published evaluation that integrates all ten within our task-free, non-episodic online target setting and a shared resource budget.

Existing benchmarks provide both focused diagnostics and settings in which several capabilities matter together, but results obtained under separate protocols do not establish their integration in one agent throughout a continuing interaction.
Our playable companion makes this integration problem tangible by using simple persistent-world rules to place several behavioral desiderata under joint pressure.
With its suggested configuration, protocol, and readouts, it can serve as a proposed benchmark, but not as a definitive test of CRL; system-level claims still require implementation-specific evidence.
The next step is to build and evaluate bounded agents whose capabilities support one another instead of succeeding only under separate assumptions.
We hope the desiderata provide a useful map for that work.

\ifCLASSOPTIONcompsoc
    \section*{Acknowledgments}
\else
    \section*{Acknowledgment}
\fi

We are grateful to everyone who has contributed, directly or indirectly, to the ideas developed in this paper.
Comments on the manuscript and broader discussions of continual reinforcement learning, both during its preparation and over a longer period, have helped shape our perspective.
In particular, we are grateful to Karl Tuyls, Gregory Makoff, Gerald Tesauro, Murray Campbell, Thore Graepel, Joel Z.~Leibo, Edgar Du\'{e}\~{n}ez-Guzm\'{a}n, Alexander Lavin, Dion H\"{a}fner, Mateusz Malinowski, Ulrich Paquet, Tyler Jackson and Demis Hassabis.

\bibliographystyle{IEEEtran}
\bibliography{crl_desiderata}

@inproceedings{abel2023,
    author    = {Abel, David and Barreto, Andr{\'e} and Van Roy, Benjamin and Precup, Doina and van Hasselt, Hado P. and Singh, Satinder P.},
    title     = {A Definition of Continual Reinforcement Learning},
    booktitle = {Advances in Neural Information Processing Systems},
    volume    = {36},
    pages     = {50377--50407},
    doi       = {10.52202/075280-2192},
    year      = {2023}
}

@inproceedings{abels2019,
    author    = {Abels, Axel and Roijers, Diederik M. and Lenaerts, Tom and Now{\'e}, Ann and Steckelmacher, Denis},
    title     = {Dynamic Weights in Multi-Objective Deep Reinforcement Learning},
    booktitle = {Proceedings of the 36th International Conference on Machine Learning},
    series    = {Proceedings of Machine Learning Research},
    volume    = {97},
    pages     = {11--20},
    publisher = {PMLR},
    year      = {2019}
}

@inproceedings{agarwal2021,
    author    = {Agarwal, Rishabh and Schwarzer, Max and Castro, Pablo Samuel and Courville, Aaron and Bellemare, Marc G.},
    title     = {Deep Reinforcement Learning at the Edge of the Statistical Precipice},
    booktitle = {Advances in Neural Information Processing Systems},
    volume    = {34},
    pages     = {29304--29320},
    year      = {2021}
}

@inproceedings{ahmed2021,
    author    = {Ahmed, Ossama and Tr{\"a}uble, Frederik and Goyal, Anirudh and Neitz, Alexander and Bengio, Yoshua and Sch{\"o}lkopf, Bernhard and Bauer, Stefan and W{\"u}thrich, Manuel},
    title     = {{CausalWorld}: A Robotic Manipulation Benchmark for Causal Structure and Transfer Learning},
    booktitle = {International Conference on Learning Representations},
    year      = {2021}
}

@inproceedings{aljundi2018,
    author    = {Aljundi, Rahaf and Babiloni, Francesca and Elhoseiny, Mohamed and Rohrbach, Marcus and Tuytelaars, Tinne},
    title     = {Memory Aware Synapses: Learning What (Not) to Forget},
    booktitle = {Proceedings of the European Conference on Computer Vision},
    pages     = {144--161},
    doi       = {10.1007/978-3-030-01219-9_9},
    year      = {2018}
}

@inproceedings{aljundi2019,
    author    = {Aljundi, Rahaf and Kelchtermans, Klaas and Tuytelaars, Tinne},
    title     = {Task-Free Continual Learning},
    booktitle = {Proceedings of the IEEE/CVF Conference on Computer Vision and Pattern Recognition},
    pages     = {11246--11255},
    doi       = {10.1109/cvpr.2019.01151},
    year      = {2019}
}

@article{aloimonos1988,
    author  = {Aloimonos, Yiannis and Weiss, Isaac and Bandyopadhyay, Amit},
    title   = {Active Vision},
    journal = {International Journal of Computer Vision},
    volume  = {1},
    number  = {4},
    pages   = {333--356},
    year    = {1988},
    doi     = {10.1007/bf00133571}
}

@article{alyahya2026,
    author  = {Alyahya, Abdulaziz and Al Siyabi, Abdallah and Ernst, Markus R. and Yang, Luke and Kuhlmann, Levin and Kowadlo, Gideon},
    title   = {{ARROW}: Augmented Replay for {RObust} World Models},
    journal = {Transactions on Machine Learning Research},
    year    = {2026}
}

@inproceedings{anand2023,
    author    = {Anand, Nishanth and Precup, Doina},
    title     = {Prediction and Control in Continual Reinforcement Learning},
    booktitle = {Advances in Neural Information Processing Systems},
    volume    = {36},
    pages     = {63779--63817},
    doi       = {10.52202/075280-2786},
    year      = {2023}
}

@inproceedings{andreas2016,
    author    = {Andreas, Jacob and Rohrbach, Marcus and Darrell, Trevor and Klein, Dan},
    title     = {Neural Module Networks},
    booktitle = {Proceedings of the IEEE Conference on Computer Vision and Pattern Recognition},
    pages     = {39--48},
    doi       = {10.1109/cvpr.2016.12},
    year      = {2016}
}

@article{arcprize2026,
    author  = {{ARC Prize Foundation}},
    title   = {{ARC-AGI-3}: A New Challenge for Frontier Agentic Intelligence},
    journal = {arXiv preprint arXiv:2603.24621},
    doi     = {10.48550/arxiv.2603.24621},
    year    = {2026}
}

@inproceedings{arjona2019,
    author    = {Arjona-Medina, Jose A. and Gillhofer, Michael and Widrich, Michael and Unterthiner, Thomas and Brandstetter, Johannes and Hochreiter, Sepp},
    title     = {{RUDDER}: Return Decomposition for Delayed Rewards},
    booktitle = {Advances in Neural Information Processing Systems},
    volume    = {32},
    year      = {2019}
}

@article{assran2025,
    author  = {Assran, Mahmoud and Bardes, Adrien and Fan, David and Garrido, Quentin and Howes, Russell and Komeili, Mojtaba and Muckley, Matthew and Rizvi, Ammar and Roberts, Claire and Sinha, Koustuv and Zholus, Artem and Arnaud, Sergio and Gejji, Abha and Martin, Ada and Robert Hogan, Francois and Dugas, Daniel and Bojanowski, Piotr and Khalidov, Vasil and Labatut, Patrick and Massa, Francisco and Szafraniec, Marc and Krishnakumar, Kapil and Li, Yong and Ma, Xiaodong and Chandar, Sarath and Meier, Franziska and LeCun, Yann and Rabbat, Michael and Ballas, Nicolas},
    title   = {{V-JEPA 2}: Self-Supervised Video Models Enable Understanding, Prediction and Planning},
    journal = {arXiv preprint arXiv:2506.09985},
    year    = {2025},
    doi     = {10.48550/arxiv.2506.09985}
}

@inproceedings{bacon2017,
    author    = {Bacon, Pierre-Luc and Harb, Jean and Precup, Doina},
    title     = {The Option-Critic Architecture},
    booktitle = {Proceedings of the AAAI Conference on Artificial Intelligence},
    volume    = {31},
    number    = {1},
    pages     = {1726--1734},
    doi       = {10.1609/aaai.v31i1.10916},
    year      = {2017}
}

@inproceedings{badia2020,
    author    = {Puigdom{\`e}nech Badia, Adri{\`a} and Sprechmann, Pablo and Vitvitskyi, Alex and Guo, Daniel and Piot, Bilal and Kapturowski, Steven and Tieleman, Olivier and Arjovsky, Mart{\'i}n and Pritzel, Alexander and Bolt, Andrew and Blundell, Charles},
    title     = {Never Give Up: Learning Directed Exploration Strategies},
    booktitle = {International Conference on Learning Representations},
    year      = {2020}
}

@article{badre2007,
    author  = {Badre, David and D'Esposito, Mark},
    title   = {Functional Magnetic Resonance Imaging Evidence for a Hierarchical Organization of the Prefrontal Cortex},
    journal = {Journal of Cognitive Neuroscience},
    volume  = {19},
    number  = {12},
    pages   = {2082--2099},
    year    = {2007},
    doi     = {10.1162/jocn.2007.19.12.2082}
}

@inproceedings{bagaria2021,
    author    = {Bagaria, Akhil and Senthil, Jason and Slivinski, Matthew and Konidaris, George},
    title     = {Robustly Learning Composable Options in Deep Reinforcement Learning},
    booktitle = {Proceedings of the Thirtieth International Joint Conference on Artificial Intelligence},
    pages     = {2161--2169},
    year      = {2021},
    doi       = {10.24963/ijcai.2021/298}
}

@article{bajcsy1988,
    author  = {Bajcsy, Ruzena},
    title   = {Active Perception},
    journal = {Proceedings of the IEEE},
    volume  = {76},
    number  = {8},
    pages   = {996--1005},
    year    = {1988},
    doi     = {10.1109/5.5968}
}

@article{bajcsy1992,
    author  = {Bajcsy, Ruzena and Campos, Mario},
    title   = {Active and Exploratory Perception},
    journal = {{CVGIP}: Image Understanding},
    volume  = {56},
    number  = {1},
    pages   = {31--40},
    year    = {1992},
    doi     = {10.1016/1049-9660(92)90083-f}
}

@article{bajcsy2018,
    author  = {Bajcsy, Ruzena and Aloimonos, Yiannis and Tsotsos, John K.},
    title   = {Revisiting Active Perception},
    journal = {Autonomous Robots},
    volume  = {42},
    number  = {2},
    pages   = {177--196},
    year    = {2018},
    doi     = {10.1007/s10514-017-9615-3}
}

@article{baker2023,
    author  = {Baker, Megan M. and New, Alexander and Aguilar-Simon, Mario and Al-Halah, Ziad and Arnold, S{\'e}bastien M. R. and Ben-Iwhiwhu, Ese and Brna, Andrew P. and Brooks, Ethan and Brown, Ryan C. and Daniels, Zachary and Daram, Anurag and Delattre, Fabien and Dellana, Ryan and Eaton, Eric and Fu, Haotian and Grauman, Kristen and Hostetler, Jesse and Iqbal, Shariq and Kent, Cassandra and Ketz, Nicholas and Kolouri, Soheil and Konidaris, George and Kudithipudi, Dhireesha and Learned-Miller, Erik and Lee, Seungwon and Littman, Michael L. and Madireddy, Sandeep and Mendez, Jorge A. and Nguyen, Eric Q. and Piatko, Christine D. and Pilly, Praveen K. and Raghavan, Aswin and Rahman, Abrar and Ramakrishnan, Santhosh Kumar and Ratzlaff, Neale and Soltoggio, Andrea and Stone, Peter and Sur, Indranil and Tang, Zhipeng and Tiwari, Saket and Vedder, Kyle and Wang, Felix and Xu, Zifan and Yanguas-Gil, Angel and Yedidsion, Harel and Yu, Shangqun and Vallabha, Gautam K.},
    title   = {A Domain-Agnostic Approach for Characterization of Lifelong Learning Systems},
    journal = {Neural Networks},
    volume  = {160},
    pages   = {274--296},
    doi     = {10.1016/j.neunet.2023.01.007},
    year    = {2023}
}

@article{ballard1991,
    author  = {Ballard, Dana H.},
    title   = {Animate Vision},
    journal = {Artificial Intelligence},
    volume  = {48},
    number  = {1},
    pages   = {57--86},
    year    = {1991},
    doi     = {10.1016/0004-3702(91)90080-4}
}

@article{ballon2026,
    author  = {Ballon, Marthe and Algaba, Andres and Ginis, Vincent},
    title   = {The Relationship between Reasoning and Performance in Large Language Models---{o3} (Mini) Thinks Harder, Not Longer},
    journal = {Scientific Reports},
    volume  = {16},
    number  = {1},
    pages   = {20896},
    year    = {2026},
    doi     = {10.1038/s41598-026-50923-2}
}

@inproceedings{barreto2017,
    author    = {Barreto, Andr{\'e} and Dabney, Will and Munos, R{\'e}mi and Hunt, Jonathan J. and Schaul, Tom and van Hasselt, Hado and Silver, David},
    title     = {Successor Features for Transfer in Reinforcement Learning},
    booktitle = {Advances in Neural Information Processing Systems},
    volume    = {30},
    year      = {2017}
}

@inproceedings{barreto2019,
    author    = {Barreto, Andr{\'e} and Borsa, Diana and Hou, Shaobo and Comanici, Gheorghe and Ayg{\"u}n, Eser and Hamel, Philippe and Toyama, Daniel and Hunt, Jonathan and Mourad, Shibl and Silver, David and Precup, Doina},
    title     = {The Option Keyboard: Combining Skills in Reinforcement Learning},
    booktitle = {Advances in Neural Information Processing Systems},
    volume    = {32},
    year      = {2019}
}

@article{barto2003,
    author  = {Barto, Andrew G. and Mahadevan, Sridhar},
    title   = {Recent Advances in Hierarchical Reinforcement Learning},
    journal = {Discrete Event Dynamic Systems},
    volume  = {13},
    number  = {4},
    pages   = {341--379},
    year    = {2003},
    doi     = {10.1023/a:1025696116075}
}

@inproceedings{behrouz2025,
    author    = {Behrouz, Ali and Zhong, Peilin and Mirrokni, Vahab},
    title     = {{Titans}: Learning to Memorize at Test Time},
    booktitle = {Advances in Neural Information Processing Systems},
    volume    = {38},
    year      = {2025},
    doi       = {10.52202/085713-3786}
}

@article{bellec2020,
    author  = {Bellec, Guillaume and Scherr, Franz and Subramoney, Anand and Hajek, Elias and Salaj, Darjan and Legenstein, Robert and Maass, Wolfgang},
    title   = {A Solution to the Learning Dilemma for Recurrent Networks of Spiking Neurons},
    journal = {Nature Communications},
    volume  = {11},
    pages   = {3625},
    doi     = {10.1038/s41467-020-17236-y},
    year    = {2020}
}

@book{bible1611,
    author    = {{King James Bible Translators}},
    editor    = {Bancroft, Richard},
    title     = {The Holy Bible, King James Version},
    publisher = {Robert Barker},
    address   = {London},
    year      = {1611},
    note      = {Hebrews 9:27}
}

@IEEEtranBSTCTL{bstctl:etal,
    CTLuse_forced_etal       = {yes},
    CTLmax_names_forced_etal = {6},
    CTLnames_show_etal       = {1}
}

@inproceedings{buesing2019,
    author    = {Buesing, Lars and Weber, Th{\'e}ophane and Zwols, Yori and Racani{\`e}re, S{\'e}bastien and Guez, Arthur and Lespiau, Jean-Baptiste and Heess, Nicolas},
    title     = {Woulda, Coulda, Shoulda: Counterfactually-Guided Policy Search},
    booktitle = {International Conference on Learning Representations},
    year      = {2019}
}

@inproceedings{buetgolfouse2023,
    author    = {Buet-Golfouse, Fran{\c{c}}ois and Pahwa, Parth},
    title     = {Robust Multi-Objective Reinforcement Learning with Dynamic Preferences},
    booktitle = {Proceedings of the 14th Asian Conference on Machine Learning},
    series    = {Proceedings of Machine Learning Research},
    volume    = {189},
    pages     = {96--111},
    publisher = {PMLR},
    year      = {2023}
}

@inproceedings{burda2019,
    author    = {Burda, Yuri and Edwards, Harri and Pathak, Deepak and Storkey, Amos and Darrell, Trevor and Efros, Alexei A.},
    title     = {Large-Scale Study of Curiosity-Driven Learning},
    booktitle = {International Conference on Learning Representations},
    year      = {2019}
}

@article{burnell2026,
    author  = {Burnell, Ryan and Yamamori, Yumeya and Firat, Orhan and Olszewska, Kate and Hughes-Fitt, Steph and Kelly, Oran and Galatzer-Levy, Isaac R. and Morris, Meredith Ringel and Dafoe, Allan and Snyder, Alison M. and Goodman, Noah D. and Botvinick, Matthew and Legg, Shane},
    title   = {Measuring Progress Toward {AGI}: A Cognitive Framework},
    journal = {arXiv preprint arXiv:2605.28405},
    doi     = {10.48550/arxiv.2605.28405},
    year    = {2026}
}

@inproceedings{caccia2023,
    author    = {Caccia, Massimo and Mueller, Jonas and Kim, Taesup and Charlin, Laurent and Fakoor, Rasool},
    title     = {Task-Agnostic Continual Reinforcement Learning: Gaining Insights and Overcoming Challenges},
    booktitle = {Proceedings of the 2nd Conference on Lifelong Learning Agents},
    series    = {Proceedings of Machine Learning Research},
    volume    = {232},
    pages     = {89--119},
    publisher = {PMLR},
    year      = {2023}
}

@article{carpenter1987,
    author  = {Carpenter, Gail A. and Grossberg, Stephen},
    title   = {A Massively Parallel Architecture for a Self-Organizing Neural Pattern Recognition Machine},
    journal = {Computer Vision, Graphics, and Image Processing},
    volume  = {37},
    number  = {1},
    pages   = {54--115},
    year    = {1987},
    doi     = {10.1016/s0734-189x(87)80014-2}
}

@inproceedings{chen2022,
    author    = {Chen, Annie S. and Sharma, Archit and Levine, Sergey and Finn, Chelsea},
    title     = {You Only Live Once: Single-Life Reinforcement Learning},
    booktitle = {Advances in Neural Information Processing Systems},
    volume    = {35},
    pages     = {14784--14797},
    doi       = {10.52202/068431-1075},
    year      = {2022}
}

@article{chollet2019,
    author  = {Chollet, Fran{\c{c}}ois},
    title   = {On the Measure of Intelligence},
    journal = {arXiv preprint arXiv:1911.01547},
    doi     = {10.48550/arxiv.1911.01547},
    year    = {2019}
}

@article{chow2018,
    author  = {Chow, Yinlam and Ghavamzadeh, Mohammad and Janson, Lucas and Pavone, Marco},
    title   = {Risk-Constrained Reinforcement Learning with Percentile Risk Criteria},
    journal = {Journal of Machine Learning Research},
    volume  = {18},
    number  = {167},
    pages   = {1--51},
    year    = {2018}
}

@article{dacosta2022,
    author  = {Da Costa, Lancelot and Lanillos, Pablo and Sajid, Noor and Friston, Karl and Khan, Shujhat},
    title   = {How Active Inference Could Help Revolutionise Robotics},
    journal = {Entropy},
    volume  = {24},
    number  = {3},
    pages   = {361},
    year    = {2022},
    doi     = {10.3390/e24030361}
}

@inproceedings{darcet2024,
    author    = {Darcet, Timoth{\'e}e and Oquab, Maxime and Mairal, Julien and Bojanowski, Piotr},
    title     = {Vision Transformers Need Registers},
    booktitle = {International Conference on Learning Representations},
    volume    = {2024},
    pages     = {2632--2652},
    year      = {2024}
}

@inproceedings{dayan1993a,
    author    = {Dayan, Peter and Hinton, Geoffrey E.},
    title     = {Feudal Reinforcement Learning},
    booktitle = {Advances in Neural Information Processing Systems},
    year      = {1993}
}

@article{dayan1993b,
    author  = {Dayan, Peter},
    title   = {Improving Generalization for Temporal Difference Learning: The Successor Representation},
    journal = {Neural Computation},
    volume  = {5},
    number  = {4},
    pages   = {613--624},
    doi     = {10.1162/neco.1993.5.4.613},
    year    = {1993}
}

@inproceedings{delange2023,
    author    = {De Lange, Matthias and van de Ven, Gido M. and Tuytelaars, Tinne},
    title     = {Continual Evaluation for Lifelong Learning: Identifying the Stability Gap},
    booktitle = {International Conference on Learning Representations},
    year      = {2023}
}

@inproceedings{deng2022,
    author    = {Deng, Fei and Jang, Ingook and Ahn, Sungjin},
    title     = {{DreamerPro}: Reconstruction-Free Model-Based Reinforcement Learning with Prototypical Representations},
    booktitle = {Proceedings of the 39th International Conference on Machine Learning},
    series    = {Proceedings of Machine Learning Research},
    volume    = {162},
    pages     = {4956--4975},
    publisher = {PMLR},
    year      = {2022}
}

@inproceedings{deschamps2026,
    author    = {Deschamps, Timon and Chaput, R{\'e}my and Guillermin, Mathieu and Matignon, La{\"e}titia},
    title     = {Utility-Based Soft Masking for Continual Multi-Objective Reinforcement Learning},
    booktitle = {Proceedings of the 25th International Conference on Autonomous Agents and Multiagent Systems},
    pages     = {2133--2141},
    publisher = {IFAAMAS},
    year      = {2026},
    doi       = {10.65109/fxdn3629}
}

@inproceedings{dewanto2022,
    author    = {Dewanto, Vektor and Gallagher, Marcus},
    title     = {Examining Average and Discounted Reward Optimality Criteria in Reinforcement Learning},
    booktitle = {{AI} 2022: Advances in Artificial Intelligence},
    series    = {Lecture Notes in Computer Science},
    volume    = {13728},
    pages     = {800--813},
    publisher = {Springer},
    doi       = {10.1007/978-3-031-22695-3_56},
    year      = {2022}
}

@article{dietterich2000,
    author  = {Dietterich, Thomas G.},
    title   = {Hierarchical Reinforcement Learning with the {MAXQ} Value Function Decomposition},
    journal = {Journal of Artificial Intelligence Research},
    volume  = {13},
    pages   = {227--303},
    year    = {2000},
    doi     = {10.1613/jair.639}
}

@article{dohare2024,
    author  = {Dohare, Shibhansh and Hernandez-Garcia, J. Fernando and Lan, Qingfeng and Rahman, Parash and Mahmood, A. Rupam and Sutton, Richard S.},
    title   = {Loss of Plasticity in Deep Continual Learning},
    journal = {Nature},
    volume  = {632},
    pages   = {768--774},
    doi     = {10.1038/s41586-024-07711-7},
    year    = {2024}
}

@book{drescher1991,
    author    = {Drescher, Gary L.},
    title     = {Made-Up Minds: A Constructivist Approach to Artificial Intelligence},
    publisher = {MIT Press},
    doi       = {10.7551/mitpress/4378.001.0001},
    year      = {1991}
}

@inproceedings{driess2023,
    author    = {Driess, Danny and Xia, Fei and Sajjadi, Mehdi S. M. and Lynch, Corey and Chowdhery, Aakanksha and Ichter, Brian and Wahid, Ayzaan and Tompson, Jonathan and Vuong, Quan and Yu, Tianhe and Huang, Wenlong and Chebotar, Yevgen and Sermanet, Pierre and Duckworth, Daniel and Levine, Sergey and Vanhoucke, Vincent and Hausman, Karol and Toussaint, Marc and Greff, Klaus and Zeng, Andy and Mordatch, Igor and Florence, Pete},
    title     = {{PaLM-E}: An Embodied Multimodal Language Model},
    booktitle = {Proceedings of the 40th International Conference on Machine Learning},
    series    = {Proceedings of Machine Learning Research},
    volume    = {202},
    pages     = {8469--8488},
    publisher = {PMLR},
    year      = {2023}
}

@inproceedings{elelimy2024,
    author    = {Elelimy, Esraa and White, Adam and Bowling, Michael and White, Martha},
    title     = {Real-Time Recurrent Learning Using Trace Units in Reinforcement Learning},
    booktitle = {Advances in Neural Information Processing Systems},
    volume    = {37},
    pages     = {17006--17043},
    doi       = {10.52202/079017-0541},
    year      = {2024}
}

@inproceedings{eysenbach2018,
    author    = {Eysenbach, Benjamin and Gu, Shixiang and Ibarz, Julian and Levine, Sergey},
    title     = {Leave No Trace: Learning to Reset for Safe and Autonomous Reinforcement Learning},
    booktitle = {International Conference on Learning Representations},
    year      = {2018}
}

@inproceedings{eysenbach2019,
    author    = {Eysenbach, Benjamin and Gupta, Abhishek and Ibarz, Julian and Levine, Sergey},
    title     = {Diversity Is All You Need: Learning Skills without a Reward Function},
    booktitle = {International Conference on Learning Representations},
    year      = {2019}
}

@inproceedings{fahlman1990,
    author    = {Fahlman, Scott E. and Lebiere, Christian},
    title     = {The Cascade-Correlation Learning Architecture},
    booktitle = {Advances in Neural Information Processing Systems},
    volume    = {2},
    year      = {1990}
}

@inproceedings{fan2022,
    author    = {Fan, Linxi and Wang, Guanzhi and Jiang, Yunfan and Mandlekar, Ajay and Yang, Yuncong and Zhu, Haoyi and Tang, Andrew and Huang, De-An and Zhu, Yuke and Anandkumar, Anima},
    title     = {{MineDojo}: Building Open-Ended Embodied Agents with Internet-Scale Knowledge},
    booktitle = {Advances in Neural Information Processing Systems},
    volume    = {35},
    pages     = {18343--18362},
    doi       = {10.52202/068431-1333},
    year      = {2022}
}

@article{felleman1991,
    author  = {Felleman, Daniel J. and Van Essen, David C.},
    title   = {Distributed Hierarchical Processing in the Primate Cerebral Cortex},
    journal = {Cerebral Cortex},
    volume  = {1},
    number  = {1},
    pages   = {1--47},
    year    = {1991},
    doi     = {10.1093/cercor/1.1.1}
}

@article{fernando2017,
    author    = {Fernando, Chrisantha and Banarse, Dylan and Blundell, Charles and Zwols, Yori and Ha, David and Rusu, Andrei A. and Pritzel, Alexander and Wierstra, Daan},
    title     = {{PathNet}: Evolution Channels Gradient Descent in Super Neural Networks},
    journal   = {arXiv preprint arXiv:1701.08734},
    doi       = {10.48550/arxiv.1701.08734},
    year      = {2017}
}

@inproceedings{frankle2019,
    author    = {Frankle, Jonathan and Carbin, Michael},
    title     = {The Lottery Ticket Hypothesis: Finding Sparse, Trainable Neural Networks},
    booktitle = {International Conference on Learning Representations},
    year      = {2019}
}

@article{french1999,
    author  = {French, Robert M.},
    title   = {Catastrophic Forgetting in Connectionist Networks},
    journal = {Trends in Cognitive Sciences},
    volume  = {3},
    number  = {4},
    pages   = {128--135},
    doi     = {10.1016/s1364-6613(99)01294-2},
    year    = {1999}
}

@article{friston2017,
    author  = {Friston, Karl J. and FitzGerald, Thomas and Rigoli, Francesco and Schwartenbeck, Philipp and Pezzulo, Giovanni},
    title   = {Active Inference: A Process Theory},
    journal = {Neural Computation},
    volume  = {29},
    number  = {1},
    pages   = {1--49},
    doi     = {10.1162/neco_a_00912},
    year    = {2017}
}

@inproceedings{fu2025,
    author    = {Fu, Haotian and Sun, Yixiang and Littman, Michael and Konidaris, George},
    title     = {Knowledge Retention in Continual Model-Based Reinforcement Learning},
    booktitle = {Proceedings of the 42nd International Conference on Machine Learning},
    series    = {Proceedings of Machine Learning Research},
    volume    = {267},
    pages     = {17832--17851},
    publisher = {PMLR},
    year      = {2025}
}

@inproceedings{goodfellow2015,
    author    = {Goodfellow, Ian J. and Shlens, Jonathon and Szegedy, Christian},
    title     = {Explaining and Harnessing Adversarial Examples},
    booktitle = {International Conference on Learning Representations},
    year      = {2015}
}

@article{graves2014,
    author  = {Graves, Alex and Wayne, Greg and Danihelka, Ivo},
    title   = {Neural {Turing} Machines},
    journal = {arXiv preprint arXiv:1410.5401},
    doi     = {10.48550/arxiv.1410.5401},
    year    = {2014}
}

@article{graves2016a,
    author  = {Graves, Alex},
    title   = {Adaptive Computation Time for Recurrent Neural Networks},
    journal = {arXiv preprint arXiv:1603.08983},
    doi     = {10.48550/arxiv.1603.08983},
    year    = {2016}
}

@article{graves2016b,
    author  = {Graves, Alex and Wayne, Greg and Reynolds, Malcolm and Harley, Tim and Danihelka, Ivo and Grabska-Barwi{\'n}ska, Agnieszka and G{\'o}mez Colmenarejo, Sergio and Grefenstette, Edward and Ramalho, Tiago and Agapiou, John and Puigdom{\`e}nech Badia, Adri{\`a} and Hermann, Karl Moritz and Zwols, Yori and Ostrovski, Georg and Cain, Adam and King, Helen and Summerfield, Christopher and Blunsom, Phil and Kavukcuoglu, Koray and Hassabis, Demis},
    title   = {Hybrid Computing Using a Neural Network with Dynamic External Memory},
    journal = {Nature},
    volume  = {538},
    pages   = {471--476},
    doi     = {10.1038/nature20101},
    year    = {2016}
}

@article{guo2025,
    author  = {Guo, Daya and Yang, Dejian and Zhang, Haowei and Song, Junxiao and Wang, Peiyi and Zhu, Qihao and Xu, Runxin and Zhang, Ruoyu and Ma, Shirong and Bi, Xiao and Zhang, Xiaokang and Yu, Xingkai and Wu, Yu and Wu, Z. F. and Gou, Zhibin and Shao, Zhihong and Li, Zhuoshu and Gao, Ziyi and Liu, Aixin and Xue, Bing and Wang, Bingxuan and Wu, Bochao and Feng, Bei and Lu, Chengda and Zhao, Chenggang and Deng, Chengqi and Ruan, Chong and Dai, Damai and Chen, Deli and Ji, Dongjie and Li, Erhang and Lin, Fangyun and Dai, Fucong and Luo, Fuli and Hao, Guangbo and Chen, Guanting and Li, Guowei and Zhang, H. and Xu, Hanwei and Ding, Honghui and Gao, Huazuo and Qu, Hui and Li, Hui and Guo, Jianzhong and Li, Jiashi and Chen, Jingchang and Yuan, Jingyang and Tu, Jinhao and Qiu, Junjie and Li, Junlong and Cai, J. L. and Ni, Jiaqi and Liang, Jian and Chen, Jin and Dong, Kai and Hu, Kai and You, Kaichao and Gao, Kaige and Guan, Kang and Huang, Kexin and Yu, Kuai and Wang, Lean and Zhang, Lecong and Zhao, Liang and Wang, Litong and Zhang, Liyue and Xu, Lei and Xia, Leyi and Zhang, Mingchuan and Zhang, Minghua and Tang, Minghui and Zhou, Mingxu and Li, Meng and Wang, Miaojun and Li, Mingming and Tian, Ning and Huang, Panpan and Zhang, Peng and Wang, Qiancheng and Chen, Qinyu and Du, Qiushi and Ge, Ruiqi and Zhang, Ruisong and Pan, Ruizhe and Wang, Runji and Chen, R. J. and Jin, R. L. and Chen, Ruyi and Lu, Shanghao and Zhou, Shangyan and Chen, Shanhuang and Ye, Shengfeng and Wang, Shiyu and Yu, Shuiping and Zhou, Shunfeng and Pan, Shuting and Li, S. S. and Zhou, Shuang and Wu, Shaoqing and Yun, Tao and Pei, Tian and Sun, Tianyu and Wang, T. and Zeng, Wangding and Liu, Wen and Liang, Wenfeng and Gao, Wenjun and Yu, Wenqin and Zhang, Wentao and Xiao, W. L. and An, Wei and Liu, Xiaodong and Wang, Xiaohan and Chen, Xiaokang and Nie, Xiaotao and Cheng, Xin and Liu, Xin and Xie, Xin and Liu, Xingchao and Yang, Xinyu and Li, Xinyuan and Su, Xuecheng and Lin, Xuheng and Li, X. Q. and Jin, Xiangyue and Shen, Xiaojin and Chen, Xiaosha and Sun, Xiaowen and Wang, Xiaoxiang and Song, Xinnan and Zhou, Xinyi and Wang, Xianzu and Shan, Xinxia and Li, Y. K. and Wang, Y. Q. and Wei, Y. X. and Zhang, Yang and Xu, Yanhong and Li, Yao and Zhao, Yao and Sun, Yaofeng and Wang, Yaohui and Yu, Yi and Zhang, Yichao and Shi, Yifan and Xiong, Yiliang and He, Ying and Piao, Yishi and Wang, Yisong and Tan, Yixuan and Ma, Yiyang and Liu, Yiyuan and Guo, Yongqiang and Ou, Yuan and Wang, Yuduan and Gong, Yue and Zou, Yuheng and He, Yujia and Xiong, Yunfan and Luo, Yuxiang and You, Yuxiang and Liu, Yuxuan and Zhou, Yuyang and Zhu, Y. X. and Huang, Yanping and Li, Yaohui and Zheng, Yi and Zhu, Yuchen and Ma, Yunxian and Tang, Ying and Zha, Yukun and Yan, Yuting and Ren, Z. Z. and Ren, Zehui and Sha, Zhangli and Fu, Zhe and Xu, Zhean and Xie, Zhenda and Zhang, Zhengyan and Hao, Zhewen and Ma, Zhicheng and Yan, Zhigang and Wu, Zhiyu and Gu, Zihui and Zhu, Zijia and Liu, Zijun and Li, Zilin and Xie, Ziwei and Song, Ziyang and Pan, Zizheng and Huang, Zhen and Xu, Zhipeng and Zhang, Zhongyu and Zhang, Zhen},
    title   = {{DeepSeek-R1} Incentivizes Reasoning in {LLM}s through Reinforcement Learning},
    journal = {Nature},
    volume  = {645},
    pages   = {633--638},
    doi     = {10.1038/s41586-025-09422-z},
    year    = {2025}
}

@inproceedings{gurnee2024,
    author    = {Gurnee, Wes and Tegmark, Max},
    title     = {Language Models Represent Space and Time},
    booktitle = {International Conference on Learning Representations},
    volume    = {2024},
    pages     = {2483--2503},
    year      = {2024}
}

@inproceedings{guss2019,
    author    = {Guss, William H. and Houghton, Brandon and Topin, Nicholay and Wang, Phillip and Codel, Cayden and Veloso, Manuela and Salakhutdinov, Ruslan},
    title     = {{MineRL}: A Large-Scale Dataset of {Minecraft} Demonstrations},
    booktitle = {Proceedings of the Twenty-Eighth International Joint Conference on Artificial Intelligence},
    pages     = {2442--2448},
    doi       = {10.24963/ijcai.2019/339},
    year      = {2019}
}

@inproceedings{ha2018,
    author    = {Ha, David and Schmidhuber, J{\"u}rgen},
    title     = {Recurrent World Models Facilitate Policy Evolution},
    booktitle = {Advances in Neural Information Processing Systems},
    volume    = {31},
    pages     = {2450--2462},
    year      = {2018}
}

@inproceedings{haarnoja2018,
    author    = {Haarnoja, Tuomas and Zhou, Aurick and Abbeel, Pieter and Levine, Sergey},
    title     = {Soft Actor-Critic: Off-Policy Maximum Entropy Deep Reinforcement Learning with a Stochastic Actor},
    booktitle = {Proceedings of the 35th International Conference on Machine Learning},
    series    = {Proceedings of Machine Learning Research},
    volume    = {80},
    pages     = {1861--1870},
    publisher = {PMLR},
    year      = {2018}
}

@inproceedings{hafner2020,
    author    = {Hafner, Danijar and Lillicrap, Timothy and Ba, Jimmy and Norouzi, Mohammad},
    title     = {Dream to Control: Learning Behaviors by Latent Imagination},
    booktitle = {International Conference on Learning Representations},
    year      = {2020}
}

@inproceedings{hafner2021,
    author    = {Hafner, Danijar and Lillicrap, Timothy and Norouzi, Mohammad and Ba, Jimmy},
    title     = {Mastering Atari with Discrete World Models},
    booktitle = {International Conference on Learning Representations},
    year      = {2021}
}

@inproceedings{hafner2022,
    author    = {Hafner, Danijar and Lee, Kuang-Huei and Fischer, Ian and Abbeel, Pieter},
    title     = {Deep Hierarchical Planning from Pixels},
    booktitle = {Advances in Neural Information Processing Systems},
    volume    = {35},
    pages     = {26091--26104},
    year      = {2022},
    doi       = {10.52202/068431-1892}
}

@article{hafner2025,
    author  = {Hafner, Danijar and Pa{\v{s}}ukonis, Jurgis and Ba, Jimmy and Lillicrap, Timothy},
    title   = {Mastering Diverse Control Tasks through World Models},
    journal = {Nature},
    volume  = {640},
    pages   = {647--653},
    year    = {2025},
    doi     = {10.1038/s41586-025-08744-2}
}

@inproceedings{han2015,
    author    = {Han, Song and Pool, Jeff and Tran, John and Dally, William J.},
    title     = {Learning Both Weights and Connections for Efficient Neural Network},
    booktitle = {Advances in Neural Information Processing Systems},
    volume    = {28},
    year      = {2015}
}

@inproceedings{han2023,
    author    = {Han, Bing and Zhao, Feifei and Zeng, Yi and Pan, Wenxuan and Shen, Guobin},
    title     = {Enhancing Efficient Continual Learning with Dynamic Structure Development of Spiking Neural Networks},
    booktitle = {Proceedings of the Thirty-Second International Joint Conference on Artificial Intelligence},
    pages     = {2993--3001},
    year      = {2023},
    doi       = {10.24963/ijcai.2023/334}
}

@inproceedings{hansen2020,
    author    = {Hansen, Steven and Dabney, Will and Barreto, Andr{\'e} and Van de Wiele, Tom and Warde-Farley, David and Mnih, Volodymyr},
    title     = {Fast Task Inference with Variational Intrinsic Successor Features},
    booktitle = {International Conference on Learning Representations},
    year      = {2020}
}

@inproceedings{hao2023,
    author    = {Hao, Shibo and Gu, Yi and Ma, Haodi and Hong, Joshua Jiahua and Wang, Zhen and Wang, Daisy Zhe and Hu, Zhiting},
    title     = {Reasoning with Language Model Is Planning with World Model},
    booktitle = {Proceedings of the 2023 Conference on Empirical Methods in Natural Language Processing},
    pages     = {8154--8173},
    publisher = {Association for Computational Linguistics},
    doi       = {10.18653/v1/2023.emnlp-main.507},
    year      = {2023}
}

@inproceedings{harun2023,
    author    = {Harun, Md Yousuf and Gallardo, Jhair and Hayes, Tyler L. and Kanan, Christopher},
    title     = {How Efficient Are Today's Continual Learning Algorithms?},
    booktitle = {Proceedings of the IEEE/CVF Conference on Computer Vision and Pattern Recognition Workshops},
    pages     = {2431--2436},
    doi       = {10.1109/cvprw59228.2023.00241},
    year      = {2023}
}

@inproceedings{harutyunyan2019,
    author    = {Harutyunyan, Anna and Dabney, Will and M{\'e}snard, Thomas and Gheshlaghi Azar, Mohammad and Piot, Bilal and Heess, Nicolas and van Hasselt, Hado and Wayne, Greg and Singh, Satinder and Precup, Doina and Munos, R{\'e}mi},
    title     = {Hindsight Credit Assignment},
    booktitle = {Advances in Neural Information Processing Systems},
    volume    = {32},
    year      = {2019}
}

@article{hasson2008,
    author  = {Hasson, Uri and Yang, Eunice and Vallines, Ignacio and Heeger, David J. and Rubin, Nava},
    title   = {A Hierarchy of Temporal Receptive Windows in Human Cortex},
    journal = {The Journal of Neuroscience},
    volume  = {28},
    number  = {10},
    pages   = {2539--2550},
    year    = {2008},
    doi     = {10.1523/jneurosci.5487-07.2008}
}

@article{hayes2022,
    author  = {Hayes, Conor F. and R{\u{a}}dulescu, Roxana and Bargiacchi, Eugenio and K{\"a}llstr{\"o}m, Johan and Macfarlane, Matthew and Reymond, Mathieu and Verstraeten, Timothy and Zintgraf, Luisa M. and Dazeley, Richard and Heintz, Fredrik and Howley, Enda and Irissappane, Athirai A. and Mannion, Patrick and Now{\'e}, Ann and de Oliveira Ramos, Gabriel and Restelli, Marcello and Vamplew, Peter and Roijers, Diederik M.},
    title   = {A Practical Guide to Multi-Objective Reinforcement Learning and Planning},
    journal = {Autonomous Agents and Multi-Agent Systems},
    volume  = {36},
    pages   = {26},
    doi     = {10.1007/s10458-022-09552-y},
    year    = {2022}
}

@article{hayes2023,
    author  = {Hayes, Conor F. and Reymond, Mathieu and Roijers, Diederik M. and Howley, Enda and Mannion, Patrick},
    title   = {Monte Carlo Tree Search Algorithms for Risk-Aware and Multi-Objective Reinforcement Learning},
    journal = {Autonomous Agents and Multi-Agent Systems},
    volume  = {37},
    pages   = {26},
    year    = {2023},
    doi     = {10.1007/s10458-022-09596-0}
}

@article{hayhoe2003,
    author  = {Hayhoe, Mary M. and Shrivastava, Anurag and Mruczek, Ryan and Pelz, Jeff B.},
    title   = {Visual Memory and Motor Planning in a Natural Task},
    journal = {Journal of Vision},
    volume  = {3},
    number  = {1},
    pages   = {49--63},
    year    = {2003},
    doi     = {10.1167/3.1.6}
}

@inproceedings{hendrycks2019,
    author    = {Hendrycks, Dan and Dietterich, Thomas},
    title     = {Benchmarking Neural Network Robustness to Common Corruptions and Perturbations},
    booktitle = {International Conference on Learning Representations},
    year      = {2019}
}

@article{hendrycks2025,
    author  = {Hendrycks, Dan and Song, Dawn and Szegedy, Christian and Lee, Honglak and Gal, Yarin and Brynjolfsson, Erik and Li, Sharon and Zou, Andy and Levine, Lionel and Han, Bo and Fu, Jie and Liu, Ziwei and Shin, Jinwoo and Lee, Kimin and Mazeika, Mantas and Phan, Long and Ingebretsen, George and Khoja, Adam and Xie, Cihang and Salaudeen, Olawale and Hein, Matthias and Zhao, Kevin and Pan, Alexander and Duvenaud, David and Li, Bo and Omohundro, Steve and Alfour, Gabriel and Tegmark, Max and McGrew, Kevin and Marcus, Gary and Tallinn, Jaan and Schmidt, Eric and Bengio, Yoshua},
    title   = {A Definition of {AGI}},
    journal = {arXiv preprint arXiv:2510.18212},
    doi     = {10.48550/arxiv.2510.18212},
    year    = {2025}
}

@article{hochreiter1997,
    author  = {Hochreiter, Sepp and Schmidhuber, J{\"u}rgen},
    title   = {Long Short-Term Memory},
    journal = {Neural Computation},
    volume  = {9},
    number  = {8},
    pages   = {1735--1780},
    doi     = {10.1162/neco.1997.9.8.1735},
    year    = {1997}
}

@article{hopfield1982,
    author  = {Hopfield, John J.},
    title   = {Neural Networks and Physical Systems with Emergent Collective Computational Abilities},
    journal = {Proceedings of the National Academy of Sciences},
    volume  = {79},
    number  = {8},
    pages   = {2554--2558},
    year    = {1982},
    doi     = {10.1073/pnas.79.8.2554}
}

@inproceedings{huang2017,
    author    = {Huang, Sandy and Papernot, Nicolas and Goodfellow, Ian and Duan, Yan and Abbeel, Pieter},
    title     = {Adversarial Attacks on Neural Network Policies},
    booktitle = {International Conference on Learning Representations, Workshop Track},
    year      = {2017}
}

@inproceedings{huang2024,
    author    = {Huang, Jianheng and Cui, Leyang and Wang, Ante and Yang, Chengyi and Liao, Xinting and Song, Linfeng and Yao, Junfeng and Su, Jinsong},
    title     = {Mitigating Catastrophic Forgetting in Large Language Models with Self-Synthesized Rehearsal},
    booktitle = {Proceedings of the 62nd Annual Meeting of the Association for Computational Linguistics (Volume 1: Long Papers)},
    pages     = {1416--1428},
    publisher = {Association for Computational Linguistics},
    year      = {2024},
    doi       = {10.18653/v1/2024.acl-long.77}
}

@article{hung2019,
    author    = {Hung, Chia-Chun and Lillicrap, Timothy and Abramson, Josh and Wu, Yan and Mirza, Mehdi and Carnevale, Federico and Ahuja, Arun and Wayne, Greg},
    title     = {Optimizing Agent Behavior over Long Time Scales by Transporting Value},
    journal   = {Nature Communications},
    volume    = {10},
    pages     = {5223},
    doi       = {10.1038/s41467-019-13073-w},
    year      = {2019}
}

@article{huszar2018,
    author  = {Husz{\'a}r, Ferenc},
    title   = {Note on the Quadratic Penalties in Elastic Weight Consolidation},
    journal = {Proceedings of the National Academy of Sciences},
    volume  = {115},
    number  = {11},
    pages   = {E2496--E2497},
    doi     = {10.1073/pnas.1717042115},
    year    = {2018}
}

@inproceedings{jaderberg2017,
    author    = {Jaderberg, Max and Mnih, Volodymyr and Czarnecki, Wojciech Marian and Schaul, Tom and Leibo, Joel Z. and Silver, David and Kavukcuoglu, Koray},
    title     = {Reinforcement Learning with Unsupervised Auxiliary Tasks},
    booktitle = {International Conference on Learning Representations},
    year      = {2017}
}

@inproceedings{jayaraman2018,
    author    = {Jayaraman, Dinesh and Grauman, Kristen},
    title     = {Learning to Look Around: Intelligently Exploring Unseen Environments for Unknown Tasks},
    booktitle = {Proceedings of the IEEE Conference on Computer Vision and Pattern Recognition},
    pages     = {1238--1247},
    doi       = {10.1109/cvpr.2018.00135},
    year      = {2018}
}

@inproceedings{jin2020,
    author    = {Jin, Chi and Krishnamurthy, Akshay and Simchowitz, Max and Yu, Tiancheng},
    title     = {Reward-Free Exploration for Reinforcement Learning},
    booktitle = {Proceedings of the 37th International Conference on Machine Learning},
    series    = {Proceedings of Machine Learning Research},
    volume    = {119},
    pages     = {4870--4879},
    publisher = {PMLR},
    year      = {2020}
}

@inproceedings{jin2023,
    author    = {Jin, Zhijing and Chen, Yuen and Leeb, Felix and Gresele, Luigi and Kamal, Ojasv and Lyu, Zhiheng and Blin, Kevin and Gonzalez Adauto, Fernando and Kleiman-Weiner, Max and Sachan, Mrinmaya and Sch{\"o}lkopf, Bernhard},
    title     = {{CLadder}: Assessing Causal Reasoning in Language Models},
    booktitle = {Advances in Neural Information Processing Systems},
    volume    = {36},
    pages     = {31038--31065},
    publisher = {Curran Associates, Inc.},
    doi       = {10.52202/075280-1353},
    year      = {2023}
}

@article{kaelbling1998,
    author  = {Kaelbling, Leslie Pack and Littman, Michael L. and Cassandra, Anthony R.},
    title   = {Planning and Acting in Partially Observable Stochastic Domains},
    journal = {Artificial Intelligence},
    volume  = {101},
    number  = {1--2},
    pages   = {99--134},
    doi     = {10.1016/S0004-3702(98)00023-X},
    year    = {1998}
}

@article{kanerva2009,
    author  = {Kanerva, Pentti},
    title   = {Hyperdimensional Computing: An Introduction to Computing in Distributed Representation with High-Dimensional Random Vectors},
    journal = {Cognitive Computation},
    volume  = {1},
    number  = {2},
    pages   = {139--159},
    doi     = {10.1007/s12559-009-9009-8},
    year    = {2009}
}

@inproceedings{kansky2017,
    author    = {Kansky, Ken and Silver, Tom and M{\'e}ly, David A. and Eldawy, Mohamed and L{\'a}zaro-Gredilla, Miguel and Lou, Xinghua and Dorfman, Nimrod and Sidor, Szymon and Phoenix, Scott and George, Dileep},
    title     = {Schema Networks: Zero-Shot Transfer with a Generative Causal Model of Intuitive Physics},
    booktitle = {Proceedings of the 34th International Conference on Machine Learning},
    series    = {Proceedings of Machine Learning Research},
    volume    = {70},
    pages     = {1809--1818},
    publisher = {PMLR},
    year      = {2017}
}

@inproceedings{ke2021,
    author    = {Ke, Nan Rosemary and Didolkar, Aniket and Mittal, Sarthak and Goyal, Anirudh and Lajoie, Guillaume and Bauer, Stefan and Rezende, Danilo Jimenez and Mozer, Michael and Bengio, Yoshua and Pal, Christopher},
    title     = {Systematic Evaluation of Causal Discovery in Visual Model Based Reinforcement Learning},
    booktitle = {Proceedings of the Neural Information Processing Systems Track on Datasets and Benchmarks},
    volume    = {1},
    year      = {2021}
}

@article{keramati2014,
    author  = {Keramati, Mehdi and Gutkin, Boris},
    title   = {Homeostatic Reinforcement Learning for Integrating Reward Collection and Physiological Stability},
    journal = {eLife},
    volume  = {3},
    pages   = {e04811},
    year    = {2014},
    doi     = {10.7554/elife.04811}
}

@inproceedings{kessler2022,
    author    = {Kessler, Samuel and Parker-Holder, Jack and Ball, Philip J. and Zohren, Stefan and Roberts, Stephen J.},
    title     = {Same State, Different Task: Continual Reinforcement Learning without Interference},
    booktitle = {Proceedings of the AAAI Conference on Artificial Intelligence},
    volume    = {36},
    number    = {7},
    pages     = {7143--7151},
    year      = {2022},
    doi       = {10.1609/aaai.v36i7.20674}
}

@inproceedings{kessler2023,
    author    = {Kessler, Samuel and Ostaszewski, Mateusz and Bortkiewicz, Micha{\l} Pawe{\l} and {\.Z}arski, Mateusz and Wo{\l}czyk, Maciej and Parker-Holder, Jack and Roberts, Stephen J. and Mi{\l}o{\'s}, Piotr},
    title     = {The Effectiveness of World Models for Continual Reinforcement Learning},
    booktitle = {Proceedings of the 2nd Conference on Lifelong Learning Agents},
    series    = {Proceedings of Machine Learning Research},
    volume    = {232},
    pages     = {184--204},
    publisher = {PMLR},
    year      = {2023}
}

@article{khetarpal2022,
    author  = {Khetarpal, Khimya and Riemer, Matthew and Rish, Irina and Precup, Doina},
    title   = {Towards Continual Reinforcement Learning: A Review and Perspectives},
    journal = {Journal of Artificial Intelligence Research},
    volume  = {75},
    pages   = {1401--1476},
    doi     = {10.1613/jair.1.13673},
    year    = {2022}
}

@inproceedings{kim2021,
    author    = {Kim, Dong Ki and Liu, Miao and Riemer, Matthew D. and Sun, Chuangchuang and Abdulhai, Marwa and Habibi, Golnaz and Lopez-Cot, Sebastian and Tesauro, Gerald and How, Jonathan P.},
    title     = {A Policy Gradient Algorithm for Learning to Learn in Multiagent Reinforcement Learning},
    booktitle = {Proceedings of the 38th International Conference on Machine Learning},
    series    = {Proceedings of Machine Learning Research},
    volume    = {139},
    pages     = {5541--5550},
    publisher = {PMLR},
    year      = {2021}
}

@article{kirkpatrick2017,
    author  = {Kirkpatrick, James and Pascanu, Razvan and Rabinowitz, Neil C. and Veness, Joel and Desjardins, Guillaume and Rusu, Andrei A. and Milan, Kieran and Quan, John and Ramalho, Tiago and Grabska-Barwi{\'n}ska, Agnieszka and Hassabis, Demis and Clopath, Claudia and Kumaran, Dharshan and Hadsell, Raia},
    title   = {Overcoming Catastrophic Forgetting in Neural Networks},
    journal = {Proceedings of the National Academy of Sciences},
    volume  = {114},
    number  = {13},
    pages   = {3521--3526},
    doi     = {10.1073/pnas.1611835114},
    year    = {2017}
}

@article{koopmans1960,
    author  = {Koopmans, Tjalling C.},
    title   = {Stationary Ordinal Utility and Impatience},
    journal = {Econometrica},
    volume  = {28},
    number  = {2},
    pages   = {287--309},
    doi     = {10.2307/1907722},
    year    = {1960}
}

@article{korbak2025,
    author  = {Korbak, Tomek and Balesni, Mikita and Barnes, Elizabeth and Bengio, Yoshua and Benton, Joe and Bloom, Joseph and Chen, Mark and Cooney, Alan and Dafoe, Allan and Dragan, Anca and Emmons, Scott and Evans, Owain and Farhi, David and Greenblatt, Ryan and Hendrycks, Dan and Hobbhahn, Marius and Hubinger, Evan and Irving, Geoffrey and Jenner, Erik and Kokotajlo, Daniel and Krakovna, Victoria and Legg, Shane and Lindner, David and Luan, David and M{\k{a}}dry, Aleksander and Michael, Julian and Nanda, Neel and Orr, Dave and Pachocki, Jakub and Perez, Ethan and Phuong, Mary and Roger, Fabien and Saxe, Joshua and Shlegeris, Buck and Soto, Mart{\'i}n and Steinberger, Eric and Wang, Jasmine and Zaremba, Wojciech and Baker, Bowen and Shah, Rohin and Mikulik, Vlad},
    title   = {Chain of Thought Monitorability: A New and Fragile Opportunity for {AI} Safety},
    journal = {arXiv preprint arXiv:2507.11473},
    doi     = {10.48550/arxiv.2507.11473},
    year    = {2025}
}

@inproceedings{krotov2016,
    author    = {Krotov, Dmitry and Hopfield, John J.},
    title     = {{Dense Associative Memory} for Pattern Recognition},
    booktitle = {Advances in Neural Information Processing Systems},
    volume    = {29},
    year      = {2016}
}

@inproceedings{laskin2021,
    author    = {Laskin, Michael and Yarats, Denis and Liu, Hao and Lee, Kimin and Zhan, Albert and Lu, Kevin and Cang, Catherine and Pinto, Lerrel and Abbeel, Pieter},
    title     = {{URLB}: Unsupervised Reinforcement Learning Benchmark},
    booktitle = {Advances in Neural Information Processing Systems Datasets and Benchmarks Track},
    volume    = {1},
    year      = {2021}
}

@article{lecun2022,
    author  = {LeCun, Yann},
    title   = {A Path Towards Autonomous Machine Intelligence},
    journal = {OpenReview},
    note    = {Version 0.9.2},
    year    = {2022}
}

@inproceedings{leibo2021,
    author    = {Leibo, Joel Z. and Due{\~n}ez-Guzm{\'a}n, Edgar A. and Vezhnevets, Alexander and Agapiou, John P. and Sunehag, Peter and Koster, Raphael and Matyas, Jayd and Beattie, Charlie and Mordatch, Igor and Graepel, Thore},
    title     = {Scalable Evaluation of Multi-Agent Reinforcement Learning with {Melting Pot}},
    booktitle = {Proceedings of the 38th International Conference on Machine Learning},
    series    = {Proceedings of Machine Learning Research},
    volume    = {139},
    pages     = {6187--6199},
    publisher = {PMLR},
    year      = {2021}
}

@article{li2018,
    author  = {Li, Zhizhong and Hoiem, Derek},
    title   = {Learning without Forgetting},
    journal = {IEEE Transactions on Pattern Analysis and Machine Intelligence},
    volume  = {40},
    number  = {12},
    pages   = {2935--2947},
    year    = {2018},
    doi     = {10.1109/tpami.2017.2773081}
}

@inproceedings{li2023,
    author    = {Li, Kenneth and Hopkins, Aspen K. and Bau, David and Vi{\'e}gas, Fernanda and Pfister, Hanspeter and Wattenberg, Martin},
    title     = {Emergent World Representations: Exploring a Sequence Model Trained on a Synthetic Task},
    booktitle = {International Conference on Learning Representations},
    year      = {2023}
}

@inproceedings{li2024a,
    author    = {Li, Lihe and Chen, Ruotong and Zhang, Ziqian and Wu, Zhichao and Li, Yi-Chen and Guan, Cong and Yu, Yang and Yuan, Lei},
    title     = {Continual Multi-Objective Reinforcement Learning via Reward Model Rehearsal},
    booktitle = {Proceedings of the Thirty-Third International Joint Conference on Artificial Intelligence},
    pages     = {4434--4442},
    year      = {2024},
    doi       = {10.24963/ijcai.2024/490}
}

@inproceedings{li2024b,
    author    = {Li, Zhiyuan and Liu, Hong and Zhou, Denny and Ma, Tengyu},
    title     = {Chain of Thought Empowers Transformers to Solve Inherently Serial Problems},
    booktitle = {International Conference on Learning Representations},
    volume    = {2024},
    pages     = {11911--11943},
    year      = {2024}
}

@inproceedings{liang2022,
    author    = {Liang, Jacky and Sharma, Mohit and LaGrassa, Alex and Vats, Shivam and Saxena, Saumya and Kroemer, Oliver},
    title     = {Search-Based Task Planning with Learned Skill Effect Models for Lifelong Robotic Manipulation},
    booktitle = {2022 IEEE International Conference on Robotics and Automation},
    pages     = {6351--6357},
    year      = {2022},
    doi       = {10.1109/icra46639.2022.9811575}
}

@inproceedings{lin2021,
    author    = {Lin, Zhiqiu and Shi, Jia and Pathak, Deepak and Ramanan, Deva},
    title     = {The {CLEAR} Benchmark: Continual {LEArning} on Real-World Imagery},
    booktitle = {Advances in Neural Information Processing Systems Datasets and Benchmarks Track},
    volume    = {1},
    year      = {2021}
}

@inproceedings{liu2021,
    author    = {Liu, Hao and Abbeel, Pieter},
    title     = {{APS}: Active Pretraining with Successor Features},
    booktitle = {Proceedings of the 38th International Conference on Machine Learning},
    series    = {Proceedings of Machine Learning Research},
    volume    = {139},
    pages     = {6736--6747},
    publisher = {PMLR},
    year      = {2021}
}

@article{liu2024,
    author  = {Liu, Nelson F. and Lin, Kevin and Hewitt, John and Paranjape, Ashwin and Bevilacqua, Michele and Petroni, Fabio and Liang, Percy},
    title   = {Lost in the Middle: How Language Models Use Long Contexts},
    journal = {Transactions of the Association for Computational Linguistics},
    volume  = {12},
    pages   = {157--173},
    year    = {2024},
    doi     = {10.1162/tacl_a_00638}
}

@inproceedings{liu2025,
    author    = {Liu, Zichen and Fu, Guoji and Du, Chao and Lee, Wee Sun and Lin, Min},
    title     = {Continual Reinforcement Learning by Planning with Online World Models},
    booktitle = {Proceedings of the 42nd International Conference on Machine Learning},
    series    = {Proceedings of Machine Learning Research},
    volume    = {267},
    pages     = {38397--38423},
    publisher = {PMLR},
    year      = {2025}
}

@inproceedings{lopezpaz2017,
    author    = {Lopez-Paz, David and Ranzato, Marc'Aurelio},
    title     = {Gradient Episodic Memory for Continual Learning},
    booktitle = {Advances in Neural Information Processing Systems},
    volume    = {30},
    year      = {2017}
}

@inproceedings{lu2021,
    author    = {Lu, Kevin and Grover, Aditya and Abbeel, Pieter and Mordatch, Igor},
    title     = {Reset-Free Lifelong Learning with Skill-Space Planning},
    booktitle = {International Conference on Learning Representations},
    year      = {2021}
}

@inproceedings{luketina2022,
    author    = {Luketina, Jelena and Flennerhag, Sebastian and Schroecker, Yannick and Abel, David and Zahavy, Tom and Singh, Satinder},
    title     = {Meta-Gradients in Non-Stationary Environments},
    booktitle = {Proceedings of the 1st Conference on Lifelong Learning Agents},
    series    = {Proceedings of Machine Learning Research},
    volume    = {199},
    pages     = {886--901},
    publisher = {PMLR},
    year      = {2022}
}

@article{mahadevan1996,
    author  = {Mahadevan, Sridhar},
    title   = {Average Reward Reinforcement Learning: Foundations, Algorithms, and Empirical Results},
    journal = {Machine Learning},
    volume  = {22},
    pages   = {159--195},
    doi     = {10.1007/bf00114727},
    year    = {1996}
}

@inproceedings{malagon2024,
    author    = {Malag{\'o}n, Mikel and Ceberio, Josu and Lozano, Jose A.},
    title     = {Self-Composing Policies for Scalable Continual Reinforcement Learning},
    booktitle = {Proceedings of the 41st International Conference on Machine Learning},
    series    = {Proceedings of Machine Learning Research},
    volume    = {235},
    pages     = {34432--34460},
    publisher = {PMLR},
    year      = {2024}
}

@inproceedings{malinowski2018,
    author    = {Malinowski, Mateusz and Doersch, Carl and Santoro, Adam and Battaglia, Peter},
    title     = {Learning Visual Question Answering by Bootstrapping Hard Attention},
    booktitle = {Proceedings of the European Conference on Computer Vision},
    pages     = {3--20},
    doi       = {10.1007/978-3-030-01231-1_1},
    year      = {2018}
}

@inproceedings{mallya2018,
    author    = {Mallya, Arun and Lazebnik, Svetlana},
    title     = {{PackNet}: Adding Multiple Tasks to a Single Network by Iterative Pruning},
    booktitle = {Proceedings of the IEEE Conference on Computer Vision and Pattern Recognition},
    pages     = {7765--7773},
    doi       = {10.1109/cvpr.2018.00810},
    year      = {2018}
}

@article{massiani2023,
    author  = {Massiani, Pierre-Fran{\c{c}}ois and Heim, Steve and Solowjow, Friedrich and Trimpe, Sebastian},
    title   = {Safe Value Functions},
    journal = {IEEE Transactions on Automatic Control},
    volume  = {68},
    number  = {5},
    pages   = {2743--2757},
    doi     = {10.1109/tac.2022.3200948},
    year    = {2023}
}

@article{mausam2008,
    author  = {Mausam and Weld, Daniel S.},
    title   = {Planning with Durative Actions in Stochastic Domains},
    journal = {Journal of Artificial Intelligence Research},
    volume  = {31},
    pages   = {33--82},
    doi     = {10.1613/jair.2269},
    year    = {2008}
}

@article{mcalister2025,
    author  = {McAlister, Hayden and Robins, Anthony and Szymanski, Lech},
    title   = {Sequential Learning in the Dense Associative Memory},
    journal = {Neural Computation},
    volume  = {37},
    number  = {10},
    pages   = {1877--1924},
    year    = {2025},
    doi     = {10.1162/neco.a.20}
}

@article{mccann2011,
    author  = {McCann, Brian C. and Hayhoe, Mary M. and Geisler, Wilson S.},
    title   = {Decoding Natural Signals from the Peripheral Retina},
    journal = {Journal of Vision},
    volume  = {11},
    number  = {10},
    pages   = {19},
    year    = {2011},
    doi     = {10.1167/11.10.19}
}

@incollection{mccloskey1989,
    author    = {McCloskey, Michael and Cohen, Neal J.},
    title     = {Catastrophic Interference in Connectionist Networks: The Sequential Learning Problem},
    booktitle = {Psychology of Learning and Motivation},
    editor    = {Bower, Gordon H.},
    volume    = {24},
    pages     = {109--165},
    publisher = {Academic Press},
    doi       = {10.1016/s0079-7421(08)60536-8},
    year      = {1989}
}

@inproceedings{mendez2022a,
    author    = {Mendez, Jorge A. and Hussing, Marcel and Gummadi, Meghna and Eaton, Eric},
    title     = {{CompoSuite}: A Compositional Reinforcement Learning Benchmark},
    booktitle = {Proceedings of the 1st Conference on Lifelong Learning Agents},
    series    = {Proceedings of Machine Learning Research},
    volume    = {199},
    pages     = {982--1003},
    publisher = {PMLR},
    year      = {2022}
}

@inproceedings{mendez2022b,
    author    = {Mendez, Jorge A. and van Seijen, Harm and Eaton, Eric},
    title     = {Modular Lifelong Reinforcement Learning via Neural Composition},
    booktitle = {International Conference on Learning Representations},
    year      = {2022}
}

@article{merel2019,
    author  = {Merel, Josh and Botvinick, Matthew and Wayne, Greg},
    title   = {Hierarchical Motor Control in Mammals and Machines},
    journal = {Nature Communications},
    volume  = {10},
    pages   = {5489},
    year    = {2019},
    doi     = {10.1038/s41467-019-13239-6}
}

@inproceedings{mesbahi2025,
    author    = {Mesbahi, Golnaz and Mohammad Panahi, Parham and Mastikhina, Olya and Tang, Steven and White, Martha and White, Adam},
    title     = {Position: Lifetime Tuning Is Incompatible with Continual Reinforcement Learning},
    booktitle = {Proceedings of the 42nd International Conference on Machine Learning},
    series    = {Proceedings of Machine Learning Research},
    volume    = {267},
    pages     = {81786--81796},
    publisher = {PMLR},
    year      = {2025}
}

@article{miller2001,
    author  = {Miller, Earl K. and Cohen, Jonathan D.},
    title   = {An Integrative Theory of Prefrontal Cortex Function},
    journal = {Annual Review of Neuroscience},
    volume  = {24},
    pages   = {167--202},
    year    = {2001},
    doi     = {10.1146/annurev.neuro.24.1.167}
}

@article{millidge2022,
    author  = {Millidge, Beren and Tschantz, Alexander and Buckley, Christopher L.},
    title   = {Predictive Coding Approximates Backprop Along Arbitrary Computation Graphs},
    journal = {Neural Computation},
    volume  = {34},
    number  = {6},
    pages   = {1329--1368},
    year    = {2022},
    doi     = {10.1162/neco_a_01497}
}

@inproceedings{mnih2014,
    author    = {Mnih, Volodymyr and Heess, Nicolas and Graves, Alex and Kavukcuoglu, Koray},
    title     = {Recurrent Models of Visual Attention},
    booktitle = {Advances in Neural Information Processing Systems},
    volume    = {27},
    year      = {2014}
}

@article{mohamed2025,
    author  = {Mohamed, Mohamed A. and Nekhomiazh, Kateryna and Vyas, Vedant and Jose, Marcos M. and Patterson, Andrew and Machado, Marlos C.},
    title   = {The Cell Must Go On: {Agar.io} for Continual Reinforcement Learning},
    journal = {arXiv preprint arXiv:2505.18347},
    doi     = {10.48550/arxiv.2505.18347},
    year    = {2025}
}

@book{moravec1988,
    author    = {Moravec, Hans P.},
    title     = {Mind Children: The Future of Robot and Human Intelligence},
    publisher = {Harvard University Press},
    year      = {1988}
}

@inproceedings{morris2024,
    author    = {Morris, Meredith Ringel and Sohl-Dickstein, Jascha and Fiedel, Noah and Warkentin, Tris and Dafoe, Allan and Faust, Aleksandra and Farabet, Cl{\'e}ment and Legg, Shane},
    title     = {Position: Levels of {AGI} for Operationalizing Progress on the Path to {AGI}},
    booktitle = {Proceedings of the 41st International Conference on Machine Learning},
    series    = {Proceedings of Machine Learning Research},
    volume    = {235},
    pages     = {36308--36321},
    publisher = {PMLR},
    year      = {2024}
}

@article{mukherjee2025,
    author  = {Mukherjee, Amitangshu and Ibrayev, Timur and Roy, Kaushik},
    title   = {On Inherent Adversarial Robustness of Active Vision Systems},
    journal = {Transactions on Machine Learning Research},
    year    = {2025}
}

@article{mumford1992,
    author  = {Mumford, David},
    title   = {On the Computational Architecture of the Neocortex. {II}. The Role of Cortico-Cortical Loops},
    journal = {Biological Cybernetics},
    volume  = {66},
    number  = {3},
    pages   = {241--251},
    doi     = {10.1007/bf00198477},
    year    = {1992}
}

@inproceedings{nachum2018,
    author    = {Nachum, Ofir and Gu, Shixiang and Lee, Honglak and Levine, Sergey},
    title     = {Data-Efficient Hierarchical Reinforcement Learning},
    booktitle = {Advances in Neural Information Processing Systems},
    volume    = {31},
    year      = {2018}
}

@inproceedings{naik2019,
    author    = {Naik, Abhishek and Shariff, Roshan and Yasui, Niko and Yao, Hengshuai and Sutton, Richard S.},
    title     = {Discounted Reinforcement Learning Is Not an Optimization Problem},
    booktitle = {Optimization Foundations of Reinforcement Learning Workshop at NeurIPS 2019},
    year      = {2019}
}

@article{najemnik2005,
    author  = {Najemnik, Jiri and Geisler, Wilson S.},
    title   = {Optimal Eye Movement Strategies in Visual Search},
    journal = {Nature},
    volume  = {434},
    number  = {7031},
    pages   = {387--391},
    year    = {2005},
    doi     = {10.1038/nature03390}
}

@inproceedings{natarajan2005,
    author    = {Natarajan, Sriraam and Tadepalli, Prasad},
    title     = {Dynamic Preferences in Multi-Criteria Reinforcement Learning},
    booktitle = {Proceedings of the 22nd International Conference on Machine Learning},
    pages     = {601--608},
    publisher = {ACM},
    year      = {2005},
    doi       = {10.1145/1102351.1102427}
}

@inproceedings{nekoei2021,
    author    = {Nekoei, Hadi and Badrinaaraayanan, Akilesh and Courville, Aaron and Chandar, Sarath},
    title     = {Continuous Coordination as a Realistic Scenario for Lifelong Learning},
    booktitle = {Proceedings of the 38th International Conference on Machine Learning},
    series    = {Proceedings of Machine Learning Research},
    volume    = {139},
    pages     = {8016--8024},
    publisher = {PMLR},
    year      = {2021}
}

@inproceedings{ni2023,
    author    = {Ni, Tianwei and Ma, Michel and Eysenbach, Benjamin and Bacon, Pierre-Luc},
    title     = {When Do Transformers Shine in {RL}? Decoupling Memory from Credit Assignment},
    booktitle = {Advances in Neural Information Processing Systems},
    volume    = {36},
    doi       = {10.52202/075280-2194},
    year      = {2023}
}

@inproceedings{okada2021,
    author    = {Okada, Masashi and Taniguchi, Tadahiro},
    title     = {Dreaming: Model-Based Reinforcement Learning by Latent Imagination without Reconstruction},
    booktitle = {2021 IEEE International Conference on Robotics and Automation},
    pages     = {4209--4215},
    year      = {2021},
    doi       = {10.1109/icra48506.2021.9560734}
}

@inproceedings{ororbia2022,
    author    = {Ororbia, Alexander G. and Mali, Ankur and Giles, C. Lee and Kifer, Daniel},
    title     = {Lifelong Neural Predictive Coding: Learning Cumulatively Online without Forgetting},
    booktitle = {Advances in Neural Information Processing Systems},
    volume    = {35},
    pages     = {5867--5881},
    doi       = {10.52202/068431-0425},
    year      = {2022}
}

@inproceedings{ororbia2023,
    author    = {Ororbia, Alexander and Mali, Ankur},
    title     = {Active Predictive Coding: Brain-Inspired Reinforcement Learning for Sparse Reward Robotic Control Problems},
    booktitle = {2023 IEEE International Conference on Robotics and Automation ({ICRA})},
    pages     = {3015--3021},
    doi       = {10.1109/ICRA48891.2023.10160530},
    year      = {2023}
}

@article{oudeyer2007,
    author  = {Oudeyer, Pierre-Yves and Kaplan, Fr{\'e}d{\'e}ric and Hafner, Verena V.},
    title   = {Intrinsic Motivation Systems for Autonomous Mental Development},
    journal = {IEEE Transactions on Evolutionary Computation},
    volume  = {11},
    number  = {2},
    pages   = {265--286},
    doi     = {10.1109/tevc.2006.890271},
    year    = {2007}
}

@article{parisi2019,
    author  = {Parisi, German I. and Kemker, Ronald and Part, Jose L. and Kanan, Christopher and Wermter, Stefan},
    title   = {Continual Lifelong Learning with Neural Networks: A Review},
    journal = {Neural Networks},
    volume  = {113},
    pages   = {54--71},
    doi     = {10.1016/j.neunet.2019.01.012},
    year    = {2019}
}

@inproceedings{park2024,
    author    = {Park, Seohong and Rybkin, Oleh and Levine, Sergey},
    title     = {{METRA}: Scalable Unsupervised {RL} with Metric-Aware Abstraction},
    booktitle = {International Conference on Learning Representations},
    volume    = {2024},
    pages     = {18579--18603},
    year      = {2024}
}

@article{pateria2021,
    author  = {Pateria, Shubham and Subagdja, Budhitama and Tan, Ah-Hwee and Quek, Chai},
    title   = {Hierarchical Reinforcement Learning: A Comprehensive Survey},
    journal = {ACM Computing Surveys},
    volume  = {54},
    number  = {5},
    pages   = {1--35},
    year    = {2021},
    doi     = {10.1145/3453160}
}

@inproceedings{pathak2017,
    author    = {Pathak, Deepak and Agrawal, Pulkit and Efros, Alexei A. and Darrell, Trevor},
    title     = {Curiosity-Driven Exploration by Self-Supervised Prediction},
    booktitle = {Proceedings of the 34th International Conference on Machine Learning},
    series    = {Proceedings of Machine Learning Research},
    volume    = {70},
    pages     = {2778--2787},
    publisher = {PMLR},
    year      = {2017}
}

@article{patterson2024,
    author  = {Patterson, Andrew and Neumann, Samuel and White, Martha and White, Adam},
    title   = {Empirical Design in Reinforcement Learning},
    journal = {Journal of Machine Learning Research},
    volume  = {25},
    number  = {318},
    pages   = {1--63},
    year    = {2024}
}

@article{payeur2021,
    author  = {Payeur, Alexandre and Guerguiev, Jordan and Zenke, Friedemann and Richards, Blake A. and Naud, Richard},
    title   = {Burst-Dependent Synaptic Plasticity Can Coordinate Learning in Hierarchical Circuits},
    journal = {Nature Neuroscience},
    volume  = {24},
    pages   = {1010--1019},
    doi     = {10.1038/s41593-021-00857-x},
    year    = {2021}
}

@article{pearl2010,
    author  = {Pearl, Judea},
    title   = {An Introduction to Causal Inference},
    journal = {The International Journal of Biostatistics},
    volume  = {6},
    number  = {2},
    doi     = {10.2202/1557-4679.1203},
    year    = {2010}
}

@book{pearl2018,
    author    = {Pearl, Judea and Mackenzie, Dana},
    title     = {The Book of Why: The New Science of Cause and Effect},
    publisher = {Basic Books},
    year      = {2018}
}

@article{pignatelli2024,
    author  = {Pignatelli, Eduardo and Ferret, Johan and van Hasselt, Hado and Geist, Matthieu and Mesnard, Thomas and Pietquin, Olivier and Toni, Laura},
    title   = {A Survey of Temporal Credit Assignment in Deep Reinforcement Learning},
    journal = {Transactions on Machine Learning Research},
    year    = {2024}
}

@inproceedings{pitis2019,
    author    = {Pitis, Silviu},
    title     = {Rethinking the Discount Factor in Reinforcement Learning: A Decision Theoretic Approach},
    booktitle = {Proceedings of the AAAI Conference on Artificial Intelligence},
    volume    = {33},
    number    = {1},
    pages     = {7949--7956},
    doi       = {10.1609/aaai.v33i01.33017949},
    year      = {2019}
}

@inproceedings{platanios2020,
    author    = {Platanios, Emmanouil Antonios and Saparov, Abulhair and Mitchell, Tom M.},
    title     = {Jelly Bean World: A Testbed for Never-Ending Learning},
    booktitle = {International Conference on Learning Representations},
    year      = {2020}
}

@inproceedings{powers2022,
    author    = {Powers, Sam and Xing, Eliot and Kolve, Eric and Mottaghi, Roozbeh and Gupta, Abhinav},
    title     = {{CORA}: Benchmarks, Baselines, and Metrics as a Platform for Continual Reinforcement Learning Agents},
    booktitle = {Proceedings of the 1st Conference on Lifelong Learning Agents},
    series    = {Proceedings of Machine Learning Research},
    volume    = {199},
    pages     = {705--743},
    publisher = {PMLR},
    year      = {2022}
}

@book{puterman1994,
    author    = {Puterman, Martin L.},
    title     = {Markov Decision Processes: Discrete Stochastic Dynamic Programming},
    publisher = {John Wiley \& Sons},
    year      = {1994},
    doi       = {10.1002/9780470316887}
}

@article{rafiee2023,
    author  = {Rafiee, Banafsheh and Abbas, Zaheer and Ghiassian, Sina and Kumaraswamy, Raksha and Sutton, Richard S. and Ludvig, Elliot A. and White, Adam},
    title   = {From Eye-Blinks to State Construction: Diagnostic Benchmarks for Online Representation Learning},
    journal = {Adaptive Behavior},
    volume  = {31},
    number  = {1},
    pages   = {3--19},
    doi     = {10.1177/10597123221085039},
    year    = {2023}
}

@inproceedings{ramsauer2021,
    author    = {Ramsauer, Hubert and Sch{\"a}fl, Bernhard and Lehner, Johannes and Seidl, Philipp and Widrich, Michael and Adler, Thomas and Gruber, Lukas and Holzleitner, Markus and Kreil, David and Kopp, Michael and Klambauer, G{\"u}nter and Brandstetter, Johannes and Hochreiter, Sepp},
    title     = {Hopfield Networks Is All You Need},
    booktitle = {International Conference on Learning Representations},
    year      = {2021}
}

@inproceedings{rannen2017,
    author    = {Rannen, Amal and Aljundi, Rahaf and Blaschko, Matthew B. and Tuytelaars, Tinne},
    title     = {Encoder Based Lifelong Learning},
    booktitle = {Proceedings of the IEEE International Conference on Computer Vision},
    pages     = {1329--1337},
    year      = {2017},
    doi       = {10.1109/iccv.2017.148}
}

@article{rao1999,
    author  = {Rao, Rajesh P. N. and Ballard, Dana H.},
    title   = {Predictive Coding in the Visual Cortex: A Functional Interpretation of Some Extra-Classical Receptive-Field Effects},
    journal = {Nature Neuroscience},
    volume  = {2},
    pages   = {79--87},
    doi     = {10.1038/4580},
    year    = {1999}
}

@article{reed2022,
    author  = {Reed, Scott and {\.Z}o{\l}na, Konrad and Parisotto, Emilio and G{\'o}mez Colmenarejo, Sergio and Novikov, Alexander and Barth-Maron, Gabriel and Gim{\'e}nez, Mai and Sulsky, Yury and Kay, Jackie and Springenberg, Jost Tobias and Eccles, Tom and Bruce, Jake and Razavi, Ali and Edwards, Ashley and Heess, Nicolas and Chen, Yutian and Hadsell, Raia and Vinyals, Oriol and Bordbar, Mahyar and de Freitas, Nando},
    title   = {A Generalist Agent},
    journal = {Transactions on Machine Learning Research},
    year    = {2022}
}

@article{rezende2020,
    author  = {Rezende, Danilo J. and Danihelka, Ivo and Papamakarios, George and Ke, Nan Rosemary and Jiang, Ray and Weber, Th{\'e}ophane and Gregor, Karol and Merzic, Hamza and Viola, Fabio and Wang, Jane and Mitrovic, Jovana and Besse, Frederic and Antonoglou, Ioannis and Buesing, Lars},
    title   = {Causally Correct Partial Models for Reinforcement Learning},
    journal = {arXiv preprint arXiv:2002.02836},
    doi     = {10.48550/arxiv.2002.02836},
    year    = {2020}
}

@article{ring1997,
    author  = {Ring, Mark B.},
    title   = {{CHILD}: A First Step Towards Continual Learning},
    journal = {Machine Learning},
    volume  = {28},
    pages   = {77--104},
    year    = {1997},
    doi     = {10.1023/a:1007331723572}
}

@inproceedings{rohanimanesh2002,
    author    = {Rohanimanesh, Khashayar and Mahadevan, Sridhar},
    title     = {Learning to Take Concurrent Actions},
    booktitle = {Advances in Neural Information Processing Systems},
    volume    = {15},
    pages     = {1651--1658},
    year      = {2002}
}

@article{roijers2013,
    author  = {Roijers, Diederik M. and Vamplew, Peter and Whiteson, Shimon and Dazeley, Richard},
    title   = {A Survey of Multi-Objective Sequential Decision-Making},
    journal = {Journal of Artificial Intelligence Research},
    volume  = {48},
    pages   = {67--113},
    year    = {2013},
    doi     = {10.1613/jair.3987}
}

@inproceedings{rolnick2019,
    author    = {Rolnick, David and Ahuja, Arun and Schwarz, Jonathan and Lillicrap, Timothy and Wayne, Gregory},
    title     = {Experience Replay for Continual Learning},
    booktitle = {Advances in Neural Information Processing Systems},
    volume    = {32},
    year      = {2019}
}

@inproceedings{rosenbaum2018,
    author    = {Rosenbaum, Clemens and Klinger, Tim and Riemer, Matthew},
    title     = {Routing Networks: Adaptive Selection of Non-Linear Functions for Multi-Task Learning},
    booktitle = {International Conference on Learning Representations},
    year      = {2018}
}

@article{rusu2016,
    author  = {Rusu, Andrei A. and Rabinowitz, Neil C. and Desjardins, Guillaume and Soyer, Hubert and Kirkpatrick, James and Kavukcuoglu, Koray and Pascanu, Razvan and Hadsell, Raia},
    title   = {Progressive Neural Networks},
    journal = {arXiv preprint arXiv:1606.04671},
    doi     = {10.48550/arxiv.1606.04671},
    year    = {2016}
}

@article{salvatori2026,
    author  = {Salvatori, Tommaso and Mali, Ankur and Buckley, Christopher L. and Lukasiewicz, Thomas and Rao, Rajesh P. N. and Friston, Karl J. and Ororbia, Alexander G.},
    title   = {A Survey on Neuro-Mimetic Deep Learning via Predictive Coding},
    journal = {Neural Networks},
    volume  = {195},
    pages   = {108161},
    year    = {2026},
    doi     = {10.1016/j.neunet.2025.108161}
}

@inproceedings{saxe2017,
    author    = {Saxe, Andrew M. and Earle, Adam C. and Rosman, Benjamin},
    title     = {Hierarchy Through Composition with Multitask {LMDP}s},
    booktitle = {Proceedings of the 34th International Conference on Machine Learning},
    series    = {Proceedings of Machine Learning Research},
    volume    = {70},
    pages     = {3017--3026},
    publisher = {PMLR},
    year      = {2017}
}

@inproceedings{scialom2022,
    author    = {Scialom, Thomas and Chakrabarty, Tuhin and Muresan, Smaranda},
    title     = {Fine-tuned Language Models Are Continual Learners},
    booktitle = {Proceedings of the 2022 Conference on Empirical Methods in Natural Language Processing},
    pages     = {6107--6122},
    publisher = {Association for Computational Linguistics},
    year      = {2022},
    doi       = {10.18653/v1/2022.emnlp-main.410}
}

@article{scellier2017,
    author  = {Scellier, Benjamin and Bengio, Yoshua},
    title   = {Equilibrium Propagation: Bridging the Gap Between Energy-Based Models and Backpropagation},
    journal = {Frontiers in Computational Neuroscience},
    volume  = {11},
    pages   = {24},
    doi     = {10.3389/fncom.2017.00024},
    year    = {2017}
}

@inproceedings{schaul2018,
    author    = {Schaul, Tom and van Hasselt, Hado and Modayil, Joseph and White, Martha and White, Adam and Bacon, Pierre-Luc and Harb, Jean and Mourad, Shibl and Bellemare, Marc G. and Precup, Doina},
    title     = {The {Barbados} 2018 List of Open Issues in Continual Learning},
    booktitle = {NeurIPS Workshop on Continual Learning},
    year      = {2018}
}

@article{schrittwieser2020,
    author  = {Schrittwieser, Julian and Antonoglou, Ioannis and Hubert, Thomas and Simonyan, Karen and Sifre, Laurent and Schmitt, Simon and Guez, Arthur and Lockhart, Edward and Hassabis, Demis and Graepel, Thore and Lillicrap, Timothy and Silver, David},
    title   = {Mastering Atari, Go, Chess and Shogi by Planning with a Learned Model},
    journal = {Nature},
    volume  = {588},
    pages   = {604--609},
    doi     = {10.1038/s41586-020-03051-4},
    year    = {2020}
}

@inproceedings{schwarz2018,
    author    = {Schwarz, Jonathan and Czarnecki, Wojciech M. and Luketina, Jelena and Grabska-Barwi{\'n}ska, Agnieszka and Teh, Yee Whye and Pascanu, Razvan and Hadsell, Raia},
    title     = {Progress \& Compress: A Scalable Framework for Continual Learning},
    booktitle = {Proceedings of the 35th International Conference on Machine Learning},
    series    = {Proceedings of Machine Learning Research},
    volume    = {80},
    pages     = {4528--4537},
    publisher = {PMLR},
    year      = {2018}
}

@inproceedings{sekar2020,
    author    = {Sekar, Ramanan and Rybkin, Oleh and Daniilidis, Kostas and Abbeel, Pieter and Hafner, Danijar and Pathak, Deepak},
    title     = {Planning to Explore via Self-Supervised World Models},
    booktitle = {Proceedings of the 37th International Conference on Machine Learning},
    series    = {Proceedings of Machine Learning Research},
    volume    = {119},
    pages     = {8583--8592},
    publisher = {PMLR},
    year      = {2020}
}

@inproceedings{serra2018,
    author    = {Serra, Joan and Suris, Didac and Miron, Marius and Karatzoglou, Alexandros},
    title     = {Overcoming Catastrophic Forgetting with Hard Attention to the Task},
    booktitle = {Proceedings of the 35th International Conference on Machine Learning},
    series    = {Proceedings of Machine Learning Research},
    volume    = {80},
    pages     = {4548--4557},
    publisher = {PMLR},
    year      = {2018}
}

@inproceedings{sezener2020,
    author    = {Sezener, Eren and Hutter, Marcus and Budden, David and Wang, Jianan and Veness, Joel},
    title     = {Online Learning in Contextual Bandits Using {Gated Linear Networks}},
    booktitle = {Advances in Neural Information Processing Systems},
    volume    = {33},
    pages     = {19467--19477},
    year      = {2020}
}

@inproceedings{shah2023,
    author    = {Shah, Muhammad A. and Kashaf, Aqsa and Raj, Bhiksha},
    title     = {Training on Foveated Images Improves Robustness to Adversarial Attacks},
    booktitle = {Advances in Neural Information Processing Systems},
    volume    = {36},
    pages     = {71008--71028},
    year      = {2023},
    doi       = {10.52202/075280-3110}
}

@article{shah2025,
    author  = {Shah, Darsh J. and Rushton, Peter and Singla, Somanshu and Parmar, Mohit and Smith, Kurt and Vanjani, Yash and Vaswani, Ashish and Chaluvaraju, Adarsh and Hojel, Andrew and Ma, Andrew and Thomas, Anil and Polloreno, Anthony and Tanwer, Ashish and Drak Sibai, Burhan and Mansingka, Divya S. and Shivaprasad, Divya and Shah, Ishaan and Stratos, Karl and Nguyen, Khoi and Callahan, Michael and Pust, Michael and Iyer, Mrinal and Monk, Philip and Mazarakis, Platon and Kapila, Ritvik and Srivastava, Saurabh and Romanski, Tim},
    title   = {Rethinking Reflection in Pre-Training},
    journal = {arXiv preprint arXiv:2504.04022},
    year    = {2025},
    doi     = {10.48550/arxiv.2504.04022}
}

@inproceedings{shang2023,
    author    = {Shang, Jinghuan and Ryoo, Michael S.},
    title     = {Active Vision Reinforcement Learning under Limited Visual Observability},
    booktitle = {Advances in Neural Information Processing Systems},
    volume    = {36},
    pages     = {10316--10338},
    year      = {2023},
    doi       = {10.52202/075280-0453}
}

@inproceedings{sharma2022,
    author    = {Sharma, Archit and Xu, Kelvin and Sardana, Nikhil and Gupta, Abhishek and Hausman, Karol and Levine, Sergey and Finn, Chelsea},
    title     = {Autonomous Reinforcement Learning: Formalism and Benchmarking},
    booktitle = {International Conference on Learning Representations},
    year      = {2022}
}

@inproceedings{sharma2020,
    author    = {Sharma, Archit and Gu, Shixiang and Levine, Sergey and Kumar, Vikash and Hausman, Karol},
    title     = {Dynamics-Aware Unsupervised Discovery of Skills},
    booktitle = {International Conference on Learning Representations},
    year      = {2020}
}

@inproceedings{shazeer2017,
    author    = {Shazeer, Noam and Mirhoseini, Azalia and Maziarz, Krzysztof and Davis, Andy and Le, Quoc V. and Hinton, Geoffrey E. and Dean, Jeff},
    title     = {Outrageously Large Neural Networks: The Sparsely-Gated Mixture-of-Experts Layer},
    booktitle = {International Conference on Learning Representations},
    year      = {2017}
}

@inproceedings{shin2017,
    author    = {Shin, Hanul and Lee, Jung Kwon and Kim, Jaehong and Kim, Jiwon},
    title     = {Continual Learning with Deep Generative Replay},
    booktitle = {Advances in Neural Information Processing Systems},
    volume    = {30},
    year      = {2017}
}

@inproceedings{shinn2023,
    author    = {Shinn, Noah and Cassano, Federico and Gopinath, Ashwin and Narasimhan, Karthik and Yao, Shunyu},
    title     = {Reflexion: Language Agents with Verbal Reinforcement Learning},
    booktitle = {Advances in Neural Information Processing Systems},
    volume    = {36},
    pages     = {8634--8652},
    year      = {2023},
    doi       = {10.52202/075280-0377}
}

@inproceedings{silver2013,
    author    = {Silver, Daniel L. and Yang, Qiang and Li, Lianghao},
    title     = {Lifelong Machine Learning Systems: Beyond Learning Algorithms},
    booktitle = {Lifelong Machine Learning---Papers from the AAAI Spring Symposium},
    pages     = {49--55},
    note      = {AAAI Technical Report SS-13-05},
    year      = {2013}
}

@inproceedings{skalse2023,
    author    = {Skalse, Joar and Abate, Alessandro},
    title     = {On the Limitations of {M}arkovian Rewards to Express Multi-Objective, Risk-Sensitive, and Modal Tasks},
    booktitle = {Proceedings of the Thirty-Ninth Conference on Uncertainty in Artificial Intelligence},
    series    = {Proceedings of Machine Learning Research},
    volume    = {216},
    pages     = {1974--1984},
    publisher = {PMLR},
    year      = {2023}
}

@article{stocco2010,
    author  = {Stocco, Andrea and Lebiere, Christian and Anderson, John R.},
    title   = {Conditional Routing of Information to the Cortex: A Model of the Basal Ganglia's Role in Cognitive Coordination},
    journal = {Psychological Review},
    volume  = {117},
    number  = {2},
    pages   = {541--574},
    year    = {2010},
    doi     = {10.1037/a0019077}
}

@inproceedings{sutton1990,
    author    = {Sutton, Richard S.},
    title     = {Integrated Architectures for Learning, Planning, and Reacting Based on Approximating Dynamic Programming},
    booktitle = {Machine Learning Proceedings 1990},
    pages     = {216--224},
    publisher = {Morgan Kaufmann},
    doi       = {10.1016/b978-1-55860-141-3.50030-4},
    year      = {1990}
}

@article{sutton1999,
    author  = {Sutton, Richard S. and Precup, Doina and Singh, Satinder},
    title   = {Between {MDPs} and Semi-{MDPs}: A Framework for Temporal Abstraction in Reinforcement Learning},
    journal = {Artificial Intelligence},
    volume  = {112},
    number  = {1--2},
    pages   = {181--211},
    doi     = {10.1016/s0004-3702(99)00052-1},
    year    = {1999}
}

@inproceedings{sutton2011,
    author    = {Sutton, Richard S. and Modayil, Joseph and Delp, Michael and Degris, Thomas and Pilarski, Patrick M. and White, Adam and Precup, Doina},
    title     = {Horde: A Scalable Real-Time Architecture for Learning Knowledge from Unsupervised Sensorimotor Interaction},
    booktitle = {Proceedings of the 10th International Conference on Autonomous Agents and Multiagent Systems},
    pages     = {761--768},
    publisher = {IFAAMAS},
    year      = {2011}
}

@book{sutton2018,
    author    = {Sutton, Richard S. and Barto, Andrew G.},
    title     = {Reinforcement Learning: An Introduction},
    edition   = {2},
    publisher = {MIT Press},
    year      = {2018}
}

@inproceedings{tamar2015,
    author    = {Tamar, Aviv and Glassner, Yonatan and Mannor, Shie},
    title     = {Optimizing the {CVaR} via Sampling},
    booktitle = {Proceedings of the AAAI Conference on Artificial Intelligence},
    volume    = {29},
    number    = {1},
    pages     = {2993--3000},
    doi       = {10.1609/aaai.v29i1.9561},
    year      = {2015}
}

@article{tang2026,
    author  = {Tang, Steven and Xiong, Xinze and Hakhverdyan, Anna and Patterson, Andrew and Adkins, Jacob and He, Jiamin and Elelimy, Esraa and Panahi, Parham Mohammad and White, Martha and White, Adam},
    title   = {Forager: A Lightweight Testbed for Continual Learning with Partial Observability in {RL}},
    journal = {arXiv preprint arXiv:2605.01131},
    doi     = {10.48550/arxiv.2605.01131},
    year    = {2026}
}

@inproceedings{tasse2022,
    author    = {Tasse, Geraud Nangue and James, Steven and Rosman, Benjamin},
    title     = {Generalisation in Lifelong Reinforcement Learning through Logical Composition},
    booktitle = {International Conference on Learning Representations},
    year      = {2022}
}

@article{teeters2023,
    author  = {Teeters, Jeffrey L. and Kleyko, Denis and Kanerva, Pentti and Olshausen, Bruno A.},
    title   = {On Separating Long- and Short-Term Memories in Hyperdimensional Computing},
    journal = {Frontiers in Neuroscience},
    volume  = {16},
    pages   = {867568},
    doi     = {10.3389/fnins.2022.867568},
    year    = {2023}
}

@article{thrun1995,
    author  = {Thrun, Sebastian and Mitchell, Tom M.},
    title   = {Lifelong Robot Learning},
    journal = {Robotics and Autonomous Systems},
    volume  = {15},
    number  = {1--2},
    pages   = {25--46},
    year    = {1995},
    doi     = {10.1016/0921-8890(95)00004-y}
}

@article{tolman1948,
    author  = {Tolman, Edward C.},
    title   = {Cognitive Maps in Rats and Men},
    journal = {Psychological Review},
    volume  = {55},
    number  = {4},
    pages   = {189--208},
    doi     = {10.1037/h0061626},
    year    = {1948}
}

@inproceedings{tomilin2023,
    author    = {Tomilin, Tristan and Fang, Meng and Zhang, Yudi and Pechenizkiy, Mykola},
    title     = {{COOM}: A Game Benchmark for Continual Reinforcement Learning},
    booktitle = {Advances in Neural Information Processing Systems},
    volume    = {36},
    pages     = {67794--67832},
    doi       = {10.52202/075280-2965},
    year      = {2023}
}

@inproceedings{touati2021,
    author    = {Touati, Ahmed and Ollivier, Yann},
    title     = {Learning One Representation to Optimize All Rewards},
    booktitle = {Advances in Neural Information Processing Systems},
    volume    = {34},
    year      = {2021}
}

@article{tsividis2026,
    author  = {Tsividis, Pedro and Loula, Jo{\~a}o and Burga, Jake and Rodriguez, Juan Pablo and Arnaud, Sergio and Foss, Nate and Campero, Andres and Subramanian, Ajay and Pouncy, Thomas and Gershman, Samuel J. and Tenenbaum, Joshua B.},
    title   = {Human-Level Learning of Complex Novel Tasks as Theory-Based Modelling, Exploration and Planning},
    journal = {Philosophical Transactions of the Royal Society A: Mathematical, Physical and Engineering Sciences},
    volume  = {384},
    number  = {2320},
    pages   = {20240529},
    year    = {2026},
    doi     = {10.1098/rsta.2024.0529}
}

@article{tsotsos1992,
    author  = {Tsotsos, John K.},
    title   = {On the Relative Complexity of Active vs. Passive Visual Search},
    journal = {International Journal of Computer Vision},
    volume  = {7},
    number  = {2},
    pages   = {127--141},
    year    = {1992},
    doi     = {10.1007/bf00128132}
}

@inproceedings{turpin2023,
    author    = {Turpin, Miles and Michael, Julian and Perez, Ethan and Bowman, Samuel R.},
    title     = {Language Models Don't Always Say What They Think: Unfaithful Explanations in Chain-of-Thought Prompting},
    booktitle = {Advances in Neural Information Processing Systems},
    volume    = {36},
    pages     = {74952--74965},
    doi       = {10.52202/075280-3275},
    year      = {2023}
}

@inproceedings{vamplew2024,
    author    = {Vamplew, Peter and Foale, Cameron and Hayes, Conor F. and Mannion, Patrick and Howley, Enda and Dazeley, Richard and Johnson, Scott and K{\"a}llstr{\"o}m, Johan and Ramos, Gabriel and R{\u{a}}dulescu, Roxana and R{\"o}pke, Willem and Roijers, Diederik M.},
    title     = {Utility-Based Reinforcement Learning: Unifying Single-Objective and Multi-Objective Reinforcement Learning},
    booktitle = {Proceedings of the 23rd International Conference on Autonomous Agents and Multiagent Systems},
    pages     = {2717--2721},
    publisher = {IFAAMAS},
    doi       = {10.65109/gqkw2339},
    year      = {2024}
}

@article{vanmoffaert2014,
    author  = {Van Moffaert, Kristof and Now{\'e}, Ann},
    title   = {Multi-Objective Reinforcement Learning Using Sets of Pareto Dominating Policies},
    journal = {Journal of Machine Learning Research},
    volume  = {15},
    pages   = {3663--3692},
    year    = {2014}
}

@inproceedings{vaswani2017,
    author    = {Vaswani, Ashish and Shazeer, Noam and Parmar, Niki and Uszkoreit, Jakob and Jones, Llion and Gomez, Aidan N. and Kaiser, {\L}ukasz and Polosukhin, Illia},
    title     = {Attention Is All You Need},
    booktitle = {Advances in Neural Information Processing Systems},
    volume    = {30},
    year      = {2017}
}

@inproceedings{veeriah2019,
    author    = {Veeriah, Vivek and Hessel, Matteo and Xu, Zhongwen and Lewis, Richard and Rajendran, Janarthanan and Oh, Junhyuk and van Hasselt, Hado and Silver, David and Singh, Satinder},
    title     = {Discovery of Useful Questions as Auxiliary Tasks},
    booktitle = {Advances in Neural Information Processing Systems},
    volume    = {32},
    year      = {2019}
}

@inproceedings{veit2016,
    author    = {Veit, Andreas and Wilber, Michael J. and Belongie, Serge J.},
    title     = {Residual Networks Behave Like Ensembles of Relatively Shallow Networks},
    booktitle = {Advances in Neural Information Processing Systems},
    volume    = {29},
    pages     = {550--558},
    year      = {2016}
}

@inproceedings{veness2021,
    author    = {Veness, Joel and Lattimore, Tor and Budden, David and Bhoopchand, Avishkar and Mattern, Christopher and Grabska-Barwi{\'n}ska, Agnieszka and Sezener, Eren and Wang, Jianan and Toth, Peter and Schmitt, Simon and Hutter, Marcus},
    title     = {Gated Linear Networks},
    booktitle = {Proceedings of the AAAI Conference on Artificial Intelligence},
    volume    = {35},
    number    = {11},
    pages     = {10015--10023},
    year      = {2021},
    doi       = {10.1609/aaai.v35i11.17202}
}

@inproceedings{veniat2021,
    author    = {Veniat, Tom and Denoyer, Ludovic and Ranzato, Marc'Aurelio},
    title     = {Efficient Continual Learning with Modular Networks and Task-Driven Priors},
    booktitle = {International Conference on Learning Representations},
    year      = {2021}
}

@article{verwimp2024,
    author  = {Verwimp, Eli and Aljundi, Rahaf and Ben-David, Shai and Bethge, Matthias and Cossu, Andrea and Gepperth, Alexander and Hayes, Tyler L. and H{\"u}llermeier, Eyke and Kanan, Christopher and Kudithipudi, Dhireesha and Lampert, Christoph H. and Mundt, Martin and Pascanu, Razvan and Popescu, Adrian and Tolias, Andreas S. and van de Weijer, Joost and Liu, Bing and Lomonaco, Vincenzo and Tuytelaars, Tinne and van de Ven, Gido M.},
    title   = {Continual Learning: Applications and the Road Forward},
    journal = {Transactions on Machine Learning Research},
    year    = {2024},
    issn    = {2835-8856}
}

@inproceedings{vezhnevets2017,
    author    = {Vezhnevets, Alexander Sasha and Osindero, Simon and Schaul, Tom and Heess, Nicolas and Jaderberg, Max and Silver, David and Kavukcuoglu, Koray},
    title     = {{FeUdal} Networks for Hierarchical Reinforcement Learning},
    booktitle = {Proceedings of the 34th International Conference on Machine Learning},
    series    = {Proceedings of Machine Learning Research},
    volume    = {70},
    pages     = {3540--3549},
    publisher = {PMLR},
    year      = {2017}
}

@inproceedings{vezhnevets2020,
    author    = {Vezhnevets, Alexander and Wu, Yuhuai and Eckstein, Maria and Leblond, R{\'e}mi and Leibo, Joel Z.},
    title     = {{OP}tions as {RE}sponses: Grounding Behavioural Hierarchies in Multi-Agent Reinforcement Learning},
    booktitle = {Proceedings of the 37th International Conference on Machine Learning},
    series    = {Proceedings of Machine Learning Research},
    volume    = {119},
    pages     = {9733--9742},
    publisher = {PMLR},
    year      = {2020}
}

@inproceedings{wan2021,
    author    = {Wan, Yi and Naik, Abhishek and Sutton, Richard S.},
    title     = {Learning and Planning in Average-Reward Markov Decision Processes},
    booktitle = {Proceedings of the 38th International Conference on Machine Learning},
    series    = {Proceedings of Machine Learning Research},
    volume    = {139},
    pages     = {10653--10662},
    publisher = {PMLR},
    year      = {2021}
}

@inproceedings{wan2022,
    author    = {Wan, Yi and Rahimi-Kalahroudi, Ali and Rajendran, Janarthanan and Momennejad, Ida and Chandar, Sarath and van Seijen, Harm H.},
    title     = {Towards Evaluating Adaptivity of Model-Based Reinforcement Learning Methods},
    booktitle = {Proceedings of the 39th International Conference on Machine Learning},
    series    = {Proceedings of Machine Learning Research},
    volume    = {162},
    pages     = {22536--22561},
    publisher = {PMLR},
    year      = {2022}
}

@inproceedings{wang2020,
    author    = {Wang, Yulin and Lv, Kangchen and Huang, Rui and Song, Shiji and Yang, Le and Huang, Gao},
    title     = {Glance and Focus: A Dynamic Approach to Reducing Spatial Redundancy in Image Classification},
    booktitle = {Advances in Neural Information Processing Systems},
    volume    = {33},
    pages     = {2432--2444},
    year      = {2020}
}

@article{wang2024b,
    author  = {Wang, Guanzhi and Xie, Yuqi and Jiang, Yunfan and Mandlekar, Ajay and Xiao, Chaowei and Zhu, Yuke and Fan, Linxi and Anandkumar, Anima},
    title   = {Voyager: An Open-Ended Embodied Agent with Large Language Models},
    journal = {Transactions on Machine Learning Research},
    year    = {2024}
}

@article{wayne2018,
    author  = {Wayne, Greg and Hung, Chia-Chun and Amos, David and Mirza, Mehdi and Ahuja, Arun and Grabska-Barwi{\'n}ska, Agnieszka and Rae, Jack and Mirowski, Piotr and Leibo, Joel Z. and Santoro, Adam and Gemici, Mevlana and Reynolds, Malcolm and Harley, Tim and Abramson, Josh and Mohamed, Shakir and Rezende, Danilo and Saxton, David and Cain, Adam and Hillier, Chloe and Silver, David and Kavukcuoglu, Koray and Botvinick, Matt and Hassabis, Demis and Lillicrap, Timothy},
    title   = {Unsupervised Predictive Memory in a Goal-Directed Agent},
    journal = {arXiv preprint arXiv:1803.10760},
    doi     = {10.48550/arxiv.1803.10760},
    year    = {2018}
}

@inproceedings{wei2022,
    author    = {Wei, Jason and Wang, Xuezhi and Schuurmans, Dale and Bosma, Maarten and Ichter, Brian and Xia, Fei and Chi, Ed H. and Le, Quoc V. and Zhou, Denny},
    title     = {Chain-of-Thought Prompting Elicits Reasoning in Large Language Models},
    booktitle = {Advances in Neural Information Processing Systems},
    volume    = {35},
    pages     = {24824--24837},
    doi       = {10.52202/068431-1800},
    year      = {2022}
}

@inproceedings{whitehead1990,
    author    = {Whitehead, Steven D. and Ballard, Dana H.},
    title     = {Active Perception and Reinforcement Learning},
    booktitle = {Machine Learning Proceedings 1990},
    pages     = {179--188},
    year      = {1990},
    doi       = {10.1016/b978-1-55860-141-3.50025-0}
}

@article{williams2017,
    author  = {Williams, Grady and Aldrich, Andrew and Theodorou, Evangelos A.},
    title   = {Model Predictive Path Integral Control: From Theory to Parallel Computation},
    journal = {Journal of Guidance, Control, and Dynamics},
    volume  = {40},
    number  = {2},
    pages   = {344--357},
    year    = {2017},
    doi     = {10.2514/1.g001921}
}

@inproceedings{wolczyk2021,
    author    = {Wo{\l}czyk, Maciej and Zaj{\k{a}}c, Micha{\l} and Pascanu, Razvan and Kuci{\'n}ski, {\L}ukasz and Mi{\l}o{\'s}, Piotr},
    title     = {Continual World: A Robotic Benchmark for Continual Reinforcement Learning},
    booktitle = {Advances in Neural Information Processing Systems},
    volume    = {34},
    pages     = {28496--28510},
    year      = {2021}
}

@inproceedings{wolczyk2022,
    author    = {Wo{\l}czyk, Maciej and Zaj{\k{a}}c, Micha{\l} and Pascanu, Razvan and Kuci{\'n}ski, {\L}ukasz and Mi{\l}o{\'s}, Piotr},
    title     = {Disentangling Transfer in Continual Reinforcement Learning},
    booktitle = {Advances in Neural Information Processing Systems},
    volume    = {35},
    pages     = {6304--6317},
    doi       = {10.52202/068431-0456},
    year      = {2022}
}

@article{yang1992,
    author  = {Yang, Qiang and Nau, Dana S. and Hendler, James},
    title   = {Merging Separately Generated Plans with Restricted Interactions},
    journal = {Computational Intelligence},
    volume  = {8},
    number  = {4},
    pages   = {648--676},
    doi     = {10.1111/j.1467-8640.1992.tb00383.x},
    year    = {1992}
}

@inproceedings{yang2019,
    author    = {Yang, Runzhe and Sun, Xingyuan and Narasimhan, Karthik},
    title     = {A Generalized Algorithm for Multi-Objective Reinforcement Learning and Policy Adaptation},
    booktitle = {Advances in Neural Information Processing Systems},
    volume    = {32},
    year      = {2019}
}

@article{ye2026,
    author  = {Ye, Seonghyeon and Ge, Yunhao and Zheng, Kaiyuan and Gao, Shenyuan and Yu, Sihyun and Kurian, George and Indupuru, Suneel and Tan, You Liang and Zhu, Chuning and Xiang, Jiannan and Malik, Ayaan and Lee, Kyungmin and Liang, William and Ranawaka, Nadun and Gu, Jiasheng and Xu, Yinzhen and Wang, Guanzhi and Hu, Fengyuan and Narayan, Avnish and Bjorck, Johan and Wang, Jing and Kim, Gwanghyun and Niu, Dantong and Zheng, Ruijie and Xie, Yuqi and Wu, Jimmy and Wang, Qi and Julian, Ryan and Xu, Danfei and Du, Yilun and Chebotar, Yevgen and Reed, Scott and Kautz, Jan and Zhu, Yuke and Fan, Linxi {``Jim''} and Jang, Joel},
    title   = {World Action Models are Zero-shot Policies},
    journal = {arXiv preprint arXiv:2602.15922},
    doi     = {10.48550/arxiv.2602.15922},
    year    = {2026}
}

@inproceedings{yoon2018,
    author    = {Yoon, Jaehong and Yang, Eunho and Lee, Jeongtae and Hwang, Sung Ju},
    title     = {Lifelong Learning with Dynamically Expandable Networks},
    booktitle = {International Conference on Learning Representations},
    year      = {2018}
}

@inproceedings{yosinski2014,
    author    = {Yosinski, Jason and Clune, Jeff and Bengio, Yoshua and Lipson, Hod},
    title     = {How Transferable Are Features in Deep Neural Networks?},
    booktitle = {Advances in Neural Information Processing Systems},
    volume    = {27},
    pages     = {3320--3328},
    year      = {2014}
}

@inproceedings{young2023,
    author    = {Young, Kenny John and Ramesh, Aditya and Kirsch, Louis and Schmidhuber, J{\"u}rgen},
    title     = {The Benefits of Model-Based Generalization in Reinforcement Learning},
    booktitle = {Proceedings of the 40th International Conference on Machine Learning},
    series    = {Proceedings of Machine Learning Research},
    volume    = {202},
    pages     = {40254--40276},
    publisher = {PMLR},
    year      = {2023}
}

@article{zador2023,
    author  = {Zador, Anthony and Escola, Sean and Richards, Blake and {\"O}lveczky, Bence and Bengio, Yoshua and Boahen, Kwabena and Botvinick, Matthew and Chklovskii, Dmitri and Churchland, Anne and Clopath, Claudia and DiCarlo, James and Ganguli, Surya and Hawkins, Jeff and K{\"o}rding, Konrad and Koulakov, Alexei and LeCun, Yann and Lillicrap, Timothy and Marblestone, Adam and Olshausen, Bruno and Pouget, Alexandre and Savin, Cristina and Sejnowski, Terrence and Simoncelli, Eero and Solla, Sara and Sussillo, David and Tolias, Andreas S. and Tsao, Doris},
    title   = {Catalyzing Next-Generation Artificial Intelligence through {NeuroAI}},
    journal = {Nature Communications},
    volume  = {14},
    pages   = {1597},
    year    = {2023},
    doi     = {10.1038/s41467-023-37180-x}
}

@inproceedings{zeiler2014,
    author    = {Zeiler, Matthew D. and Fergus, Rob},
    title     = {Visualizing and Understanding Convolutional Networks},
    booktitle = {European Conference on Computer Vision},
    pages     = {818--833},
    year      = {2014},
    doi       = {10.1007/978-3-319-10590-1_53}
}

@inproceedings{zenke2017,
    author    = {Zenke, Friedemann and Poole, Ben and Ganguli, Surya},
    title     = {Continual Learning Through Synaptic Intelligence},
    booktitle = {Proceedings of the 34th International Conference on Machine Learning},
    series    = {Proceedings of Machine Learning Research},
    volume    = {70},
    pages     = {3987--3995},
    publisher = {PMLR},
    year      = {2017}
}

@article{zeno2021,
    author  = {Zeno, Chen and Golan, Itay and Hoffer, Elad and Soudry, Daniel},
    title   = {Task-Agnostic Continual Learning Using Online Variational Bayes with Fixed-Point Updates},
    journal = {Neural Computation},
    volume  = {33},
    number  = {11},
    pages   = {3139--3177},
    year    = {2021},
    doi     = {10.1162/neco_a_01430}
}

\end{document}